\documentclass{svproc}
\usepackage{url}

\usepackage{epsfig}
\usepackage{graphicx}
\usepackage{subfig}

\usepackage{amsmath}
\usepackage{amssymb}
\usepackage{mathtools}
\usepackage{xcolor}
\usepackage{multirow}
\usepackage{adjustbox}
\usepackage{booktabs}
\usepackage{makecell}

\usepackage{tikz}

\usepackage{pgfplots}
\usepackage{pgfplotstable}
\usepackage{filecontents}

\usetikzlibrary{calc}
\usetikzlibrary[pgfplots.groupplots]
\usepgfplotslibrary{fillbetween}
\definecolor{Pal_1}{HTML}{4472C4}
\definecolor{Pal_2}{HTML}{ED7D31}
\definecolor{Pal_3}{HTML}{A5A5A5}
\definecolor{Pal_4}{HTML}{FFC000}
\definecolor{Pal_5}{HTML}{5B9BD5}
\definecolor{Pal_6}{HTML}{70AD47}
\definecolor{Pal_7}{HTML}{264478}
\definecolor{Pal_8}{HTML}{122366}

\pgfplotscreateplotcyclelist{mycolorlist}{%
Pal_1 ,mark=x,mark options={scale=2}, line width=0.7pt\\%
Pal_2,mark=square, line width=0.7pt\\%
Pal_7,mark=square, line width=0.7pt\\%
Pal_6,mark=square, line width=0.7pt\\%
Pal_4, mark=diamond, line width=0.7pt\\%
Pal_5, mark=triangle, line width=0.7pt\\%
Pal_6,mark=triangle, line width=0.7pt\\%
Pal_8,mark=triangle, line width=0.7pt\\%
Pal_2,mark=o, line width=0.7pt\\%
}

\pgfplotscreateplotcyclelist{pgdcolorlist}{%
Pal_1 ,mark=x,mark options={scale=2}, line width=1pt\\%
Pal_4, mark=diamond, line width=1pt\\%
Pal_5, mark=triangle, line width=1pt\\%
Pal_6,mark=triangle, line width=1pt\\%
Pal_8,mark=triangle, line width=1pt\\%
Pal_2,mark=o, line width=1pt\\%
Pal_7,mark=square, line width=1pt\\%
Pal_6,mark=square, line width=1pt\\%
Pal_2,mark=square, line width=1pt\\%
}

\pgfplotscreateplotcyclelist{pgddashcolorlist}{%
Pal_1 ,mark=x, line width=1pt, opacity= 0.7, dotted,mark options={solid,scale=2}\\%
Pal_4, mark=diamond, line width=1pt, opacity= 0.7, dotted, mark options={solid}\\%
Pal_5, mark=triangle, line width=1pt, opacity= 0.7, dotted, mark options={solid}\\%
Pal_6,mark=triangle, line width=1pt, opacity= 0.7, dotted, mark options={solid}\\%
Pal_8,mark=triangle, line width=1pt, opacity= 0.7, dotted, mark options={solid}\\%
Pal_2,mark=o, line width=1pt, opacity= 0.7, dotted, mark options={solid}\\%
Pal_7,mark=square, line width=1pt, opacity= 0.7, dotted, mark options={solid}\\%
Pal_6,mark=square, line width=1pt, opacity= 0.7, dotted, mark options={solid}\\%
Pal_2,mark=square, line width=1pt, opacity= 0.7, dotted, mark options={solid}\\%
}

\pgfplotscreateplotcyclelist{mycolorlist3}{%
Pal_1 \\%
Pal_2,densely dashed \\%
Pal_7,densely dashed \\%
Pal_6,densely dashed \\%
Pal_4 \\%
Pal_5 \\%
Pal_6 \\%
Pal_8 \\%
Pal_2 \\%
}

\pgfplotscreateplotcyclelist{mycolorlist2}{%
Pal_1 ,mark=x,mark options={scale=1.5}, line width=0.7pt, only marks \\%
Pal_2,mark=square,mark options={scale=0.5}, line width=0.7pt, only marks \\%
Pal_7,mark=square,mark options={scale=0.5}, line width=0.7pt, only marks \\%
Pal_6,mark=square,mark options={scale=0.5}, line width=0.7pt, only marks \\%
Pal_3,mark=square,mark options={scale=0.5}, line width=0.7pt, only marks \\%
Pal_4, mark=diamond,mark options={scale=0.5}, line width=0.7pt, only marks \\%
Pal_5, mark=triangle,mark options={scale=0.5}, line width=0.7pt, only marks \\%
Pal_6,mark=triangle,mark options={scale=0.5}, line width=0.7pt, only marks \\%
Pal_8,mark=triangle,mark options={scale=0.5}, line width=0.7pt, only marks \\%
Pal_1 ,mark=o,mark options={scale=0.5}, line width=0.7pt, only marks \\%
Pal_2,mark=o,mark options={scale=0.5}, line width=0.7pt, only marks \\%
}

\pgfplotscreateplotcyclelist{ymodcolorlist}{%
fill=Pal_1 \\%
fill=Pal_2 \\%
fill=Pal_7 \\%
fill=Pal_4 \\%
}

\makeatletter
\pgfplotsset{
    /pgfplots/flexible xticklabels from table/.code n args={3}{%
        \pgfplotstableread[#3]{#1}\coordinate@table
        \pgfplotstablegetcolumn{#2}\of{\coordinate@table}\to\pgfplots@xticklabels
        \let\pgfplots@xticklabel=\pgfplots@user@ticklabel@list@x
    }
}
\makeatother

\usepgfplotslibrary{statistics}
\makeatletter
\pgfplotsset{
    boxplot prepared from table/.code={
        \def\tikz@plot@handler{\pgfplotsplothandlerboxplotprepared}%
        \pgfplotsset{
            /pgfplots/boxplot prepared from table/.cd,
            #1,
        }
    },
    /pgfplots/boxplot prepared from table/.cd,
        table/.code={\pgfplotstablecopy{#1}\to\boxplot@datatable},
        row/.initial=0,
        make style readable from table/.style={
            #1/.code={
                \pgfplotstablegetelem{\pgfkeysvalueof{/pgfplots/boxplot prepared from table/row}}{##1}\of\boxplot@datatable
                \pgfplotsset{boxplot/#1/.expand once={\pgfplotsretval}}
            }
        },
        make style readable from table=lower whisker,
        make style readable from table=upper whisker,
        make style readable from table=lower quartile,
        make style readable from table=upper quartile,
        make style readable from table=median,
        make style readable from table=lower notch,
        make style readable from table=upper notch
}
\makeatother

\pgfplotsset{
legend image code/.code={
\draw[mark repeat=2,mark phase=2]
plot coordinates {
(0cm,0cm)
(0.15cm,0cm)        %% default is (0.3cm,0cm)
(0.3cm,0cm)         %% default is (0.6cm,0cm)
};%
}
}
\usepackage{enumitem,amssymb}
\usepackage{pifont}
\newcommand{\etal}{et al.~}

\begin{document}
\mainmatter
%%%%%%%%% TITLE
\title{BreakingBED - Breaking Binary and Efficient Deep Neural Networks by Adversarial Attacks}
\author{Manoj-Rohit Vemparala*\inst{1},
Alexander Frickenstein*\inst{1}, Nael Fasfous*\inst{2},\\
Lukas Frickenstein*\inst{1}, Qi Zhao\inst{1}, Sabine Kuhn\inst{1}, Daniel Ehrhardt\inst{1},\\ Yuankai Wu\inst{1}, Christian Unger \inst{1}, 
 Naveen-Shankar Nagaraja \inst{1}, Walter Stechele\inst{2}\\
* indicates equal contributions}
\authorrunning{Vemparala, A. Frickenstein, Fasfous, L. Frickenstrin et al.}
\institute{BMW Autonomous Driving \and Technical University of Munich\\
\textsuperscript{1}$\mathtt{firstname.lastname@bmw.de}$, \textsuperscript{2}$\mathtt{firstname.lastname@tum.de}$}
\titlerunning{Breaking Binary and Efficient Deep Neural Networks}

\maketitle
%       __   __  ___  __        __  ___ 
%  /\  |__) /__`  |  |__)  /\  /  `  |  
% /~~\ |__) .__/  |  |  \ /~~\ \__,  |
\begin{abstract}
%Exploring Convolutional Neural Networks (CNNs) for embedded applications is challenging because the neural network need to be resource-efficient and also accurate with respect to the given task.
%This are the two most discussed aspects of CNN compression. But when it comes to an real-world application, also the robustness of the CNN has to be taken into consideration. In this paper, we thoroughly study the \emph{robustness} of a wide variety of \emph{distilled}, \emph{pruned} and \emph{binarized} neural networks against blackbox and whitebox adversarial attacks, \ie Local Search, GenAttack, FGSM, PGD,C\&W and DeepFool. Experimental results are shown for knowledge distilled CNNs, pruned models, using SotA agent based compression, and binarized neural notworks, i.e. XNOR and ABC, trained on Cifar10 and ImageNet datasets. The analysis comes with CNN heat maps with class activation mapping (CAM).
Deploying convolutional neural networks (CNNs) for embedded applications presents many challenges in balancing resource-efficiency and task-related accuracy.
These two aspects have been well-researched in the field of CNN compression. In real-world applications, a third important aspect comes into play, namely the robustness of the CNN. In this paper, we thoroughly study the robustness of  uncompressed, distilled, pruned and binarized neural networks against white-box and black-box adversarial attacks (FGSM, PGD, C\&W, DeepFool, LocalSearch and GenAttack). These new insights facilitate defensive training schemes or reactive filtering methods, where the attack is detected and the input is discarded and/or cleaned. Experimental results are shown for distilled CNNs, agent-based state-of-the-art pruned models, and binarized neural networks (BNNs) such as XNOR-Net and ABC-Net, trained on CIFAR-10 and ImageNet datasets. We present evaluation methods to simplify the comparison between CNNs under different attack schemes using loss/accuracy levels, stress-strain graphs, box-plots and class activation mapping (CAM). 
%trailer result 
Our analysis reveals susceptible behavior of uncompressed and pruned CNNs against all kinds of attacks. The distilled models exhibit their strength against all white box attacks with an exception of C\&W. Furthermore, binary neural networks exhibit resilient behavior compared to their baselines and other compressed variants.  
% We analyze CNNs with different baseline accuracies, compression schemes under different attack schemes, allowing for classification of robustness properties. The analysis is supported by attack loss niveaus, stress-strain graphs, box-plots, and class activation mapping (CAM).
% We propose multiple robustness evaluation methods to intuitively 
% \lukas{intuitively sounds subjective. Alternativ: ".. evaluation methods to ease/simplify the comparison between CNNs .."} make comparisons between CNNs with different baseline accuracies under different attack schemes, allowing for a classification of robustness properties. The analysis is supported by \manoj{PGD loss niveaus}, stress-strain graphs, box-plots, principle component analysis (PCA) and class activation mapping (CAM). \manoj{Some conclusion would be great.}
% %\alex{strength surface (lukas)}
%\alex{reformulate Abstract - more clear motivation: breakingBED and its analysis is precondition for firewall and other defensive training schemes}
\end{abstract}

% %%%%%%%%% BODY TEXT

%        ___  __   __   __        __  ___    __       
% | |\ |  |  |__) /  \ |  \ |  | /  `  |  | /  \ |\ | 
% | | \|  |  |  \ \__/ |__/ \__/ \__,  |  | \__/ | \| 
\section{Introduction}
    %Neural Network Compression is an extensively studied topic reducing the computational complexity~\cite{}, the memory demand~\cite{} and/or the energy consumption~\cite{} of a Deep Neural Network (DNN) deployed on an embedded system. 
Neural network compression is an extensively studied topic for reducing the computational complexity~\cite{rastegari2016,lin2017,courbariaux2016}, the memory demand~\cite{LeCun1990,AMC2019,Han2015} and/or the energy consumption~\cite{EnergyAwarePruning} of deep neural networks (DNN) deployed on embedded systems. 
%Therefore, compression widens the potential for DNNs being applied in real world scenarios. Especially, in the field of autonomous driving, increasingly deeper and larger Convolutional Neural Networks (CNNs) enabling computer vision based applications, \ie pedestrian detection or free-space detection been deployed on resource constrained HW-platforms. But, systems in autonomous cars are safety critical to which a threat can cause major issues which are not negotiable.
These aspects widen the potential for DNN applications in real-world scenarios. Particularly in the field of robotics and autonomous driving, increasingly deeper and larger convolutional neural networks (CNNs) are deployed on resource-constrained hardware platforms, enabling computer vision-based applications, such as pedestrian detection or free-space detection.
Systems in autonomous vehicles are safety critical, maintaining zero-tolerance for potential threats to functional safety.
%One option to attack (break) neural networks is by injecting small permutations called adversarial attacks. Under the assumption of varying degree of information of the CNN, several \emph{blackbox} (\eg GenAttack\cite{}, LocalSearch\cite{}) and \emph{whitebox} (FGSM\cite{}, DeepFool\cite{}, JSMA\cite{}) are potential threats.
Attacking (breaking) neural networks can be done by injecting small perturbations to their inputs, referred to as adversarial attacks~\cite{szegedy2013intriguing}. Under the assumption of varying degrees of information on the CNN and the accessibility of its internal parameters, several \emph{black-box} (GenAttack~\cite{GenAttack}, LocalSearch~\cite{LocalSearch}) and \emph{white-box} (FGSM~\cite{goodfellow2014explaining}, DeepFool~\cite{Deepfool} and Carlini \& Wagner~\cite{CW}) attacks are potential threats. Understanding these threats helps to develop pro-active~\cite{Goldblum2020AdversariallyRD} and re-active~\cite{Papernot2017ExtendingDD} methods to defend against adversarial examples and thereby improve CNN robustness.

% \begin{table}[ht]
%     \centering
%     %\caption{BreakingBED experimental setup.}
%     \resizebox{0.95\textwidth}{!}{
%     \begin{tabular}{ll|ccccc}
%     \toprule
%         \multicolumn{2}{l|}{\multirow{2}{*}{\textbf{Adversarial Attack}}}&\textbf{Vanilla} & \textbf{KD} & \textbf{AMC} & \multicolumn{2}{c}{\textbf{BNN}}\\
%         &&\cite{He2015DeepRL} &\cite{contrastive_kd} & \cite{AMC2019} & XNOR\cite{rastegari2016} & ABC\cite{lin2017}\\
%         \midrule
%         \midrule
%         \multirow{3}{*}{\rotatebox{90}{\makecell{\textbf{White} \\ \textbf{Box}}}}&FGSM/PGD\cite{goodfellow2014explaining} &\cmark &\cmark &\cmark&\cmark &\cmark\\
%         &C$\&$W\cite{CW} &\cmark &\cmark &\cmark&\cmark&\cmark\\
%         &DeepFool\cite{Deepfool} &\cmark &\cmark &\cmark&\cmark&\cmark\\
%         \midrule
%         \multirow{2}{*}{\rotatebox{90}{\makecell{\textbf{Black} \\ \textbf{Box}}}}&Local Search\cite{LocalSearch} &\cmark &\cmark &\cmark&\cmark&\cmark\\
%         &GenAttack\cite{GenAttack} &\cmark &\cmark &\cmark&\cmark&\cmark\\
%         \bottomrule
%     \end{tabular}}
%     \label{tab:summary_experiments}
%     %\vspace{-3ex}
% \end{table}

\begin{figure*}[h]
    \centering
    \includegraphics[width=1\textwidth]{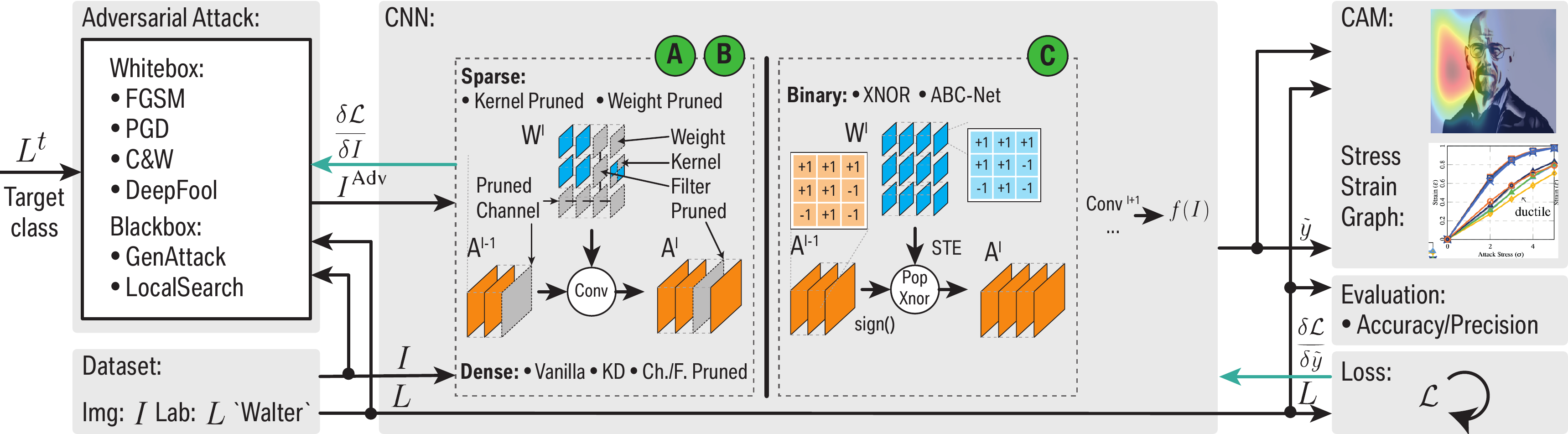}
    \caption{Experimental setup of \emph{BreakingBED} for breaking binary (\textcolor{green}{C}) and efficient (\textcolor{green}{A}) and (\textcolor{green}{B}) DNNs attacked with white-box (FGSM, PGD and C\&W) and black-box (LocalSearch and GenAttack) adversarial attacks. Evaluated by using  loss/accuracy levels, stress-strain graphs, box-plots and class activation mapping (CAM).}
    \label{fig:breaking_bed}
    \vspace{-3.5ex}
\end{figure*}

%\nael{most points addressed below, let me know if any corrections/additional refs are needed}
%\alex{fine!}
% \manoj{Incorporate attack detection methods as one solution.}
Recent works investigated the mitigation of such threats through robust training of neural networks~\cite{coteachNIPS2018_8072} and robust neural architecture search (NAS) techniques~\cite{guo2019nas}. In~\cite{lin2018defensive}, the authors compress neural networks through robust quantization, lowering the computational complexity while maintaining good performance against potential attacks. Further investigations on the robustness of binary neural networks (BNNs) were carried out in~\cite{galloway2018attacking}, where BNNs were attacked with white-box (FGSM~\cite{goodfellow2014explaining} and C\&W~\cite{CW}) and black-box~\cite{Papernot} techniques. The robustness of BNNs was concluded, albeit on basic adverserially trained networks from~\cite{Papernot} and a small set of attacks.
%\alex{does this sentence belong to our contribution -> put to next block in line 53}
%\alex{In this context, our?}The analysis helps to develop pro-active~\cite{Goldblum2020AdversariallyRD} and re-active~\cite{Papernot2017ExtendingDD} methods to defend against adversarial examples and improve CNN robustness.

%In order to gain a deeper knowledge about the effectiveness of adversarial attacks applied to binary and efficient deep neural networks we performed a set of thorough experiments. In detail we infiltrate adversarial attacks to vanilla full-precision, knowledge distilled, pruned and binary DNNs, summarized in Tab.~\ref{tab:summary_experiments}.

%\alex{may help to explain the structure the paper, @Nael: what is your intuition to remove it?}\nael{Just to save some space, but lets keep it then}
%Before going into more detail on the analysis presented in Sec.~\ref{sec:experiments},
%the subsequent two sections give an introduction to CNN compression \ref{sec:compression} %and an
%overview of state-of-the-art methods for applying adversarial attacks CNNs %\ref{sec:adversarial_attacks}. 

In order to get a deeper understanding of the effectiveness of adversarial attacks (Sec.~\ref{sec:adversarial_attacks}), applied to binary and efficient DNNs (Sec.~\ref{sec:compression}), we perform an extensive set of 
% \lukas{robustness evaluation} 
robustness evaluation experiments. In detail, we expose vanilla full-precision, distilled, pruned and binary DNNs to a variety of adversarial attacks in Sec.~\ref{sec:experiments}.
%In the following sections, an overview of CNN compression techniques is presented in Sec.~\ref{sec:compression}, followed by an overview of state-of-the-art methods for applying adversarial attacks on CNNs in Sec.~\ref{sec:adversarial_attacks}. Subsequently, a detailed analysis of the experimentation results is presented in Sec.~\ref{sec:experiments}.
%The experimental results are presented in Chapter 5, followed by a conclusion to summarize the work and a short outlook on future work in Chapter (6).

    %\vspace{-2.5ex}
    
%  __   __         __   __   ___  __   __     __       
% /  ` /  \  |\/| |__) |__) |__  /__` /__` | /  \ |\ | 
% \__, \__/  |  | |    |  \ |___ .__/ .__/ | \__/ | \|    
\section{Compression of Deep Neural Networks}
\label{sec:compression}
    %\nael{Lets not \textit{explicitly} say this as most of our experiments are on CIFAR-10: Training Convolutional Neural Networks on large datasets like ImageNet is a computationally intensive process that generally requires an array of hardware accelerators. After training the network it usually needs to be deployed on smaller appliances or even embedded devices. On these devices, the amount of computations is rather limited.}

Many works in literature have focused on reducing the redundancy emerging from training deeper and wider neural networks, aiming to mitigate the challenges of their deployment on edge devices. Compression techniques such as knowledge distillation, pruning, and binarization can potentially make CNNs more efficient in embedded settings.%. The following subsections discuss the three techniques in detail.

%In order to still be able to perform inference for a Neural Network, the amount of calculations needs to be reduces. A great number of network compression techniques for Neural Networks has been proposed in literature. Prominent candidates are \emph{Knowledge distillation} ,\emph{Pruning} and \emph{Binarization}, but not limited to it.
    \subsection{Knowledge Distillation}
    %Knowledge distillation, the transfer of knowledge between a teacher and a student network, turn out to be a very promising compression technique, because neural networks perform better when trained on the softmax ouput of a powerful teacher than on hard class labels. However with regard to adversarial robustness a naturally trained teacher produces student models, which are not robust to adversarial attacks \cite{CaliniW16a}.

%This problem was addressed by Adversarially Robust Destillation \cite{}, by combining adversarial training with knowledge distillation to transfer adversarial robustness even on large scale datasets (Cifar-100) \alex{DS not relevant}, where KD fails.

Knowledge distillation (KD) is the transfer of knowledge from a teacher to a student network~\cite{contrastive_kd,hinton2015distilling}. The student can be a smaller DNN, which is trained on the soft labels of the larger teacher network, achieving an improvement in an accuracy-efficiency trade-off.
% \lukas{trade-off}. 
The student represents a compressed version of the teacher, condensing its knowledge. 
% \manoj{We can discard the KD-CE formulation. }
%As the soft labels of teacher 
%\lukas{switch from soft labels of teacher to softmax output of teacher gives room for miss understandings/confusion despite the fact that both are the same - see Proposal Sentence}  contains information on the other classes in contrast to the hard dataset labels, the student is ideally able to learn both the correct classification \textit{and} the distribution of closeness among other classes. 
% Proposal
This paper focuses on 
% two common methods for 
KD training, using Kullback-Leibler (KL) divergence between the teacher and the student output distribution formulated as the loss function in Eq.~\eqref{eq:kd}. Here, $\sigma(f_t(I))$ and $\sigma(f_s(I))$ represent the softmax output logits of the teacher and student network respectively, computed for a sample image $I$ in a mini batch of $N$ samples.
% \lukas{Here, $\sigma(f_t(i))$ and $\sigma(f_s(i))$ represent the softmax output logits of the teacher and student network respectively, computed for a sample image $i$ in a mini batch of $N$ samples.}
% The objective  for KD-CE is to minimize the cross-entropy (CE) between the teacher ($f_t(i)$) and the student softmax outputs ($f_s(i)$) for a given image $i$, 
% In the following, we refer it to $\mathcal{L}^{KL}_{\text{KD}}$.
\begin{equation} \label{eq:kd}\small
%  \mathcal{L}^{CE}_{\text{KD}}= - \sum_{i=1}^N f_t(i) \log(f_s(i))
 \hspace{0.5cm} \mathcal{L}^{KL}_{\text{KD}}(f_t, f_s, T)=\sum_{n=1}^N \sigma(f_t(I_{n})/T) \log\bigg( \frac{\sigma(f_t(I_{n})/T)}{\sigma(f_s(I_{n})/T)}\bigg)
\end{equation}
During the knowledge transfer using the teacher's logits, a softmax temperature $T\gg1$ is used. During the evaluation, we use $T=1$ to obtain softmax-cross entropy loss. 
% This loss results in saturated gradients and therby weakens attacks such as FGSM~\cite{goodfellow2014explaining}. However, distilled networks with high temperatures can also be vulnerable against the CW attack~\cite{CW}. \lukas{Is it a good idea to put conclusions/experiment findings in the background section? - Move the findings to the particular conclusion section where the conclusion can be drawn from some figure/experiment results.}
% This problem was addressed in Adversarially Robust Distillation~\cite{goldblum2019adversarially}, by combining adversarial training with knowledge distillation to transfer adversarial robustness. \manoj{We need to double check. }

%\vspace{-1ex}
%During training the student is updated with a combination of cross entropy loss, to improve natural accuracy and kulback-leibler divergence, which acts as a consistency regualrizer to train the neural network to produce identical output on a natural image and adversarial image.
% \yuankai{To occur together with better performance, knowledge distillation are extending to revise deeper and complex convolution neural network in the object detection. Chen~\etal~\cite{chen2017learning} demonstrate a novel framework for learning compact and fast CNN based object detectors with the knowledge distillation.  }
    %\vspace{-2.5ex}
    \subsection{Pruning}
    %\textit{Pruning} is one of the methods that aims to develop smaller and more efficient neural networks. It has been proven as a significant process in a wide range of compress CNN models for achieving higher compression rates while maintaining prediction accuracy. In previous work, \cite{LeCun1990} used it to shrink network complexity and avoid over-fitting. The main idea is to use second-derivative information to make a trade-off between network complexity and training set error.

Pruning aims to eliminate redundancies in DNNs and produce smaller, more efficient neural networks. Pruning has been investigated in many works, over a wide range of DNN models, achieving high compression rates while maintaining high prediction accuracy~\cite{Han2015,AMC2019,geometric_mean_filter}. % \nael{Too specific: In previous work, \cite{LeCun1990} used it to shrink network complexity and avoid over-fitting. The main idea is to use second-derivative information to make a trade-off between network complexity and training set error.}

%Recently, the regular pruning techniques which remove a larger part of the network, such as a channel or a filter, are dedicated on achieving significant acceleration and compression with negligible accuracy loss. In 2017 He et al.\cite{He2017} propose an iterative two-step algorithm to effectively prune network layers , by a LASSO regression based channel selection and leastsquare reconstruction.

Guo~\etal~\cite{Guo2016} present an irregular pruning method, which can significantly reduce the parameter redundancy by integrating connection pruning with the retraining process.
Recently, structured pruning techniques, which remove larger, regular parts of the network, achieve a tangible improvement in hardware acceleration with a negligible accuracy loss~\cite{Han2015,ADAM2018,Frickenstein2019,He2017}. 
%He~\etal~\cite{He2017} propose an iterative, two-step algorithm to effectively prune network layers, through LASSO regression-based channel selection and least-square reconstruction. 
More recently, He~\etal proposed a learning-based compression method in AMC-AutoML~\cite{AMC2019}. The authors leverage a reinforcement learning (RL) agent, which learns the possible sparsities in each layer and prunes them based on
% \lukas{a?} 
an $\ell_2$-norm heuristic. We adapt the RL-agent of AMC-AutoML to support different pruning regularities such as filter-wise (F. Prune), channel-wise (Ch. Prune), kernel-wise (K. Prune) and weight-wise (W. Prune) pruning (shown in Fig.~\ref{fig:breaking_bed}). Pruning input channels from a layer also discards corresponding output filters from previous layers. Thus, Ch./F. Prune result in a similar compression ratio and CNN structure. The pruning regularity has a direct impact on the hardware implementation complexity and throughput benefits. In this paper, the pruning rate is set at a constant value of $50\%$ over all experiments and pruning regularities.

    \subsection{Binarization}
    %\textit{Binarization} represents the most drastic form of \textit{Quantization} where the network weights $B$ are constrained to just 2 values.

Binarization represents the most aggressive form of quantization, where the network weights $W$ %$B = \mathrm{sign}(W) \in \{-1, +1\}$ 
and activations are constrained to $\pm1$ values.
%This greatly reduces the memory consumption of a Neural Network. In theory this technique could decrease a network's memory footprint by up to $32 \times$ compared to standard 4-byte floating point numbers used on modern accelerators like GPUs.
This greatly reduces the memory requirements of DNNs. In theory, binarizing a single-precision floating-point DNN, reduces its memory footprint by up to $\times 32$.
%Different schemes for Binarization of a CNN have been proposed. Courbariaux et al. introduced the concept of training neural networks with binary weights during the forward propagation \cite{courbariaux2015}. The authors later augmented this approach by using binary activations in addition to binary weights (BNN). \cite{courbariaux2016}
Different schemes for binarization of a DNN have been proposed. Courbariaux~\etal~\cite{courbariaux2015} introduced the concept of training neural networks with binary weights $B$ during inference and maintaining a latent representation during back-propagation. The authors later augmented this approach with binarized activations~\cite{courbariaux2016}.

Rastegari~\etal~\cite{rastegari2016} introduced XNOR-Net, where the convolution of an input feature map $A^{l-1}$ and weight tensor $W$ is approximated by a combination of XNOR operations and \textit{popcounts} $\oplus$, followed by a multiplication with a scaling factor $\alpha$, such that $\mathtt{Conv}(A^{l-1},W) \approx (\mathtt{sign}(A^{l-1}) \oplus B) \cdot \alpha$ (shown in Fig.~\ref{fig:breaking_bed}).

%Rastegari et al. introduced XNOR-Net for which the convolutions of input $I$ and weights $W$ are approximated by a combination of XNOR operations and bitcount followed by a multiplication with a scaling factor $\alpha$.

%\nael{Could be removed: Furthermore, the authors present a new structure for each CNN block. The new structure reduces the information loss due to binarization:
%\begin{align*}
%  \mathrm{Batch Normalization} \\
%    \Downarrow \\
%  \mathrm{BinaryActivation} \\
%    \Downarrow \\
%  \mathrm{BinaryConvolution} \\
%    \Downarrow \\
%  \mathrm{Pooling} \\
%\end{align*}
%Batch Normalization $\rightarrow$ Binary Activation $\rightarrow$ Binary Convolution $\rightarrow$ Pooling.
%Rastegari et al. applied BNN to large datasets like ImageNet for the first time and achieved 51.2\% accuracy using ResNet18 as a model. \cite{rastegari2016}}

%BinaryNets typically suffer from accuaracy degradation. Therefore, a lot of work has been done in order to mitigate this problem. Lin et al. created a novel scheme for the training of binary CNNs named Accurate Binary Convolutional Neural Network (ABC-Net). The authors improved the accuracy of Binary NN by using a linear combination of multiple binary bases for weights and activations. The ABC-Net achieves $65.0\%$ top-1 accuracy on ImageNet with a ResNet18 architecture if 5 binary bases are used for both weights and activations. This is very close to the $69.3\%$ accuracy of the full-precision counterpart. \cite{lin2017}

Binary neural networks (BNNs) typically suffer from accuracy degradation. To mitigate this problem, Lin~\etal~\cite{lin2017} proposed a scheme for Accurate Binary CNNs (ABC-Net). The authors approximated the convolution by using a linear combination of multiple binary bases for weights and activations, shrinking the accuracy gap to full-precision counterparts. In this paper, we implement ABC-Net and XNOR-Net binarization techniques, to evaluate the effect of adversarial attacks on accurate BNNs.
\section{Adversarial Attacks}
\label{sec:adversarial_attacks}
    %Adversarial examples are manipulated inputs that are created from natural data by adding adversarial biases. The purpose of adding such an adversarial noise is to secretly deceive the deep learning model, i.e. distort the input by applying small intentional worst-case szenarios to input examples, so that the model is forced to predict a wrong result with high confidence \cite{DBLP:journals/corr/abs-1807-00051}. 
One option to attack (break) neural networks is by injecting small perturbations (adversarial biases) called adversarial attacks. %The purpose of adding such biases is to deceive the DNN. 
%An adversarial attack is successful if the model makes a wrong prediction with high confidence~\cite{DBLP:journals/corr/abs-1807-00051}. 
%Finding a minimial, inperceptual adversarial pertubation $\delta$, that forces a given classifier to missclassify an image $I$ with true label $L$ can be discribed by:
An adversarial example $I^{Adv}$ that forces a given classifier with parameters $\theta$ to misclassify an image $I$ with true label $L$, renders a successful non-target attack: $\mathcal{A} = \{ I^{Adv} |  \theta(I^{Adv}) \neq L$ \}. Whereas, a successful target attack can be defined as: $\mathcal{A} = \{ I^{Adv} | \theta(I^{Adv}) = L^{t}$\} for some target class $t$.
%Finding an imperceptible adversarial perturbation $\delta$, that forces a given classifier with parameters $\theta$ to misclassify an image $I$ with true label $L$ renders a successful attack: $\mathcal{A} = \{ \mathcal{I}_{Adv} \hspace{0.1cm} | \hspace{0.1cm} \theta(\mathcal{I}_{Adv}) \neq L$ \}. 
% is detailed in Eq.~\ref{eq:AdvEx}.
The capability of the adversary can be described by a set of allowed perturbations $S: D(I, I^{Adv}) \leq \epsilon$, restricting the maximum possible perturbation distance $D(I, I^{Adv})$ to a given image $I$ by some adversarial manipulation budget $\epsilon$.
%$\epsilon$ is the adversarial manipulation budget and $\|.\|_p$ is an $\ell_p$ norm operator. 
Finding $I^{Adv}$ can be formulated as a maximization problem as defined in Eq.~\eqref{eq:AdvEx}, whereby various attacks are designed to be effective using different distance metrics ($\ell_1$, $\ell_2$, $\ell_\infty$)~\cite{EvaluatingAdvRobust2019}.
\begin{equation}\label{eq:AdvEx}
    \max\limits_{I^{Adv}\in \mathcal{S}} \mathcal{L}(I^{Adv}, L, \theta) %\hspace{0.5cm}  \text{ s.t.} \hspace{0.5cm} \|\delta\|_p \leq \epsilon
\end{equation}
% \cite{shafahi2019adversarial} \lukas{$\|l\|_p$ refers to a distance and $\|.\|_p$ refers just to a norm operator rewrite as ... "and $\|l\|_p$ is some distance metric. "}. 

%Finding adversarial examples can be formally discribed by solving the following minimization problem \cite{szegedy2013intriguing}. The objective is to find a pertubation ${\delta}$, which minimizes the classifier ${f}$'s prediction $x^{adv} = x + \delta$ for correct label $y$, by adding it to the input sample $x$:

%Attacks can be categorized regarding to the access to internal model parameters:

Attacks can be categorized regarding the degree of accessibility to a model's internal parameters $\theta$. White-box attacks~\cite{goodfellow2014explaining,MakhzaniSJG15,CW,MoosaviDezfooli16,KurakinGB16a,szegedy2013intriguing} assume complete model transparency, allowing full control and access to the target CNN. 
%In this case, the adversary has complete information on the model's structure, parameters and gradients.
%White Box Attacks \cite{goodfellow2014explaining, MakhzaniSJG15, CW, MoosaviDezfooli16, KurakinGB16a, szegedy2013intriguing} assume model transparency that allows full control and access to a target Cnn, that means the adversary has full information about structure, weight parameters and gradients. 
In most real-world scenarios, a model's fine internal details are not easily accessible, rendering white-box attacks less practical~\cite{ZOO}.
%\nael{Unfortunately?? Its good that most systems dont release their internal config :P} Unfortunately, most real systems do not release their internal configurations (including network structure and weighting), so that white box attacks cannot be used in practice \cite{Chen2017ZOOZO}.
On the other hand, black-box attacks~\cite{GenAttack,LocalSearch} assume no such information. The adversary can be a standard user, with access to only the inputs and the outputs of a targeted model. Such attacks are more practical and can have severe consequences in real-time critical applications.

%In Black-Box case the adversary acts as a standard user and can only access the input and the output of a targeted model but not the internal configurations /cite{advexample} to generate adversarial examples \cite{GenAttack, LocalSearch}.

%Further because different models learn similar functions when they are trained for the same task, adversarial perturbations being highly aligned with the weight vectors of a model. This results in generalization of adversarial examples over different models \cite{goodfellow2014explaining}. Therefore Black-Box adversarial examples are often White-Box attacks transfered from one model to another \cite{KurakinGB16a}.

Different models learn similar features when they are trained for the same task. Adversarial perturbations are highly aligned with the weight vectors of a model. This results in the generalization of adversarial examples over different models~\cite{goodfellow2014explaining}, making it possible to transfer a white-box attack from one model as a black-box attack to another~\cite{KurakinGB16a}.
    \subsection{White-box Attacks}\label{subsec:WhiteBox}
        %The most commonly used attack to verify the robustness of a neural network against input purtubations is Fast Gradient Sign Method (FGSM) \cite{goodfellow2014explaining}. FGSM is an efficient solution for equation \ref{eq:AdvEx} to generate adversarial examples $I^{Adv}$ by calculating the $\nabla$ gradient of the cost function $\mathcal{L}$ with respect to the input $I$ and correct label $L^{Tar}$. The input variation parameter $\epsilon$ controls the pertubation's amplitude \cite{KurakinGB16a}:
\noindent
\textbf{Fast Gradient Sign Method: }
The most commonly used attack to verify the robustness of neural networks against input perturbations is the fast gradient sign method (FGSM)~\cite{goodfellow2014explaining}. FGSM linearizes the loss function of a neural network around $\theta$ by calculating its gradient $\nabla \mathcal{L}(I, L, \theta)$ to generate adversarial examples $I^{Adv}$, resulting in an efficient solution to Eq.~\eqref{eq:AdvEx}.
%FGSM linearizes the loss function around $\theta$ and thus, is an efficient solution for Eq.~\ref{eq:AdvEx} to generate adversarial examples $I^{Adv}$ by calculating the $\nabla$ gradient of the loss function $\mathcal{L}$ with respect to the input $I$. 
The input variation parameter $\epsilon$ controls the perturbation's amplitude \cite{KurakinGB16a}, as expressed in Eq.~\eqref{eq:FGSM}. 
%\lukas{Note:  $I^{Adv}$ is computed by calculating the gradient of the loss function w.r.t. the input image i  (denoted by $\nabla_{I} \mathcal{L}$ ) - the loss function itself is dependent on the input, theta and the label - however the text describes that the gradient of the loss function is computed with respect to all three parameters. }
% \lukas{Gradient of Loss is missing the Model Parameter \(\theta\) since the gradient of the loss also depends on the model params. - see also original paper of FGSM. }

\begin{equation}\label{eq:FGSM}
   I^{Adv} = I + \epsilon \cdot \text{sign} \left(\nabla \mathcal{L} \left( I, L,  \theta \right) \right)
\end{equation}

The attack is strengthened when performed iteratively. This can be considered as an extension of FGSM, generating adversarial samples using a small step-size~\cite{KurakinGB16a}.
%
%   PGD-Attack
%

\noindent
\textbf{Projected Gradient Descent: }
An even more effective variant is iterative projected gradient descent (PGD) on the loss function with uniform random noise initialization~\cite{shafahi2019adversarial}, expressed in Eq.~\eqref{eq:PGD}.
\begin{equation}\label{eq:PGD}
   I^{Adv}_{i+1} = \pi_{\mathcal{S}} \left( I^{Adv}_{i} + \alpha \cdot \nabla \mathcal{L} \left( I_{i}^{Adv}, L,\theta \right) \right) 
\end{equation}
Here, adversary examples $I^{Adv}_{i+1}$ are generated by taking one step into the ascent direction of the loss gradient $\nabla \mathcal{L}(I_{i}^{Adv}, L, \theta)$ with respect to the previous image $I_{i}^{Adv}$ at iteration $i$, where the step-size is scaled by $\alpha$, followed by a potential projection $\pi$ onto the legal set $\mathcal{S}$. 
%\nael{reconsider use of $t$, it represents the target class in CW paragraph} \lukas{done - restructured/reformulated adversarial attacks}
Legal adversaries are ensured by a projection $\pi$ onto the legal set $\mathcal{I}+\mathcal{S}$ with $\mathcal{S} = \{ \delta : ||\delta||_{p} \leq \epsilon \}$.
%\nael{any reason to introduce $\mathcal{S}?$, why not use $\delta$ directly as before. Legality was also defined before, so we can reuse it. If you want to capture compounded perturbations due to iterative attacks, then a better legality definition can be formulated}. 
% \lukas{done}.
If not mentioned otherwise, PGD attacks focus on the $\ell_\infty$-norm as a distance metric for $D(I, I^{Adv})$, representing an $\ell_\infty$-ball around natural images $I$.
%A projection onto the legal set $\pi_{S}$ is performed by clipping $I_{t+1}^{Adv}$ to the interval $[-\epsilon, \epsilon]$.
%$I_t$ defines the image at iteration step $t$ and $\alpha$ defines the step-size of the optimization step in the ascent direction of $\nabla_{I} \mathcal{L}$. 

The iterative multi-step optimization method is able to converge to local maxima of the non-concave and constrained maximization problem, defined in Eq.~\ref{eq:AdvEx}, representing possible worst-case adversaries for the underlying model. By considering random uniform initialization, arbitrary starting points on the corresponding loss surface are ensured, thus resulting in an exploration of potentially varying local maxima and lastly giving rise to the structural behavior of the corresponding loss surface. This renders the PGD attack as the ``ultimate'' first-order adversary, as stated by Madry et al.~\cite{Madry2018PGD}. 
% NOTE: Might invoke following sentence agin! - Lukas
%Thus, investigating the resulting landscape of local maxima indicates the adversarial robustness of the model at hand~\cite{Madry2018PGD} with regard to the threat model. \manoj{We did not define the threat model. with regard to the selected hyperparamters of PGD ($\alpha, \mathcal{I}, \epsilon$)} \lukas{Would put the definition of threat model to sec 4.1. Same as e.g. naming the batch size of some experiment.}

%Random uniform initialization ensures random seeds on the corresponding loss surface $\mathcal{L}(I_t, \theta, L)$, thus ensures random exploration of potential varying local maxima and lastly gives rise to structural behavior of the loss surface, highlighting the necessity of random initialization.  
% The projection of the current solution $I_{t+1}^{Adv}$ onto the legal constrained set $\pi_{||\delta||_{p}}$ can be defined as $\mathcal{X}_{adv}^{t+1} \xleftarrow{} \pi_{\mathcal{S}}\left( \mathcal{X}_{adv}^{t+1}  \right)$

        %\input{data/30_aa_methods/JSMA.tex}
        % Carlini and Wagner Attacks
%Carlini and Wagner (C\&W) \cite{CW} proposed a target attack, which turned out as one of the strongest attacks \cite{ThreatOfAdversarialAttacks}, to refute the promising defensive approach of defensive distillation \cite{DefensiveDistillation}.

\noindent
\textbf{Carlini \& Wagner: }
Carlini and Wagner (C\&W)~\cite{CW} presented a targeted attack, to refute the promising defensive approach of defensive distillation~\cite{DefensiveDistillation}. The proposed C\&W attack emerged as one of the strongest attacks in literature~\cite{ThreatOfAdversarialAttacks}. CW finds perturbations $\delta$ with minimal distance $D(I, I+\delta)$ that will change the classification of image $I$ to the target class $t$. This is a challenging non-linear optimization problem and therefore the authors introduce a function $g$, such that $g(I + \delta) = 0$ when the classifier gets fooled towards the target class. The attack constructs adversarial examples which try to minimize the objective as mentioned in Eq.~\eqref{eq:CW}. 
\begin{equation}\label{eq:CW}
\begin{aligned}
        &\min (\| \delta \|_{p} + \epsilon \cdot g(I + \delta)),\\
        \text{where}~&g(I) = ((\max_{j \neq t} Z(I)_j) - Z(I)_t)^+
\end{aligned}
\end{equation}
%\lukas{move "," to the end of the formula since it is needed before the "where"}
%where $\min \| \delta \|_{p}$ denotes the distance of original and attacked image, $f(I+\delta)$ is the objective function and is regularized by a factor $c$. The authors evaluated the following objective function $f(I)$ as the most effective one to minimize the distance between the actual prediction logits and the prediction logits of the target class.
$Z(I)_j$ indicates the output of the CNN for class $j$ 
% \nael{reconsider use of $i$, it is used in other parts as the iterator} 
before the softmax layer. The minimum condition $g(I) = 0$ occurs when $Z(I)_t \leq Z(I)_j$ $\forall j \neq t$. The choice of $\epsilon$ maintains a trade-off between the attacked image similarity and the success rate of the target class. Using $\ell_2$ distance metric, the objective function is minimized through the gradient decent. 

\noindent
\textbf{DeepFool: }
With the DeepFool~\cite{Deepfool} attack, the authors propose a method to generate adversarial examples that fool classifiers on large-scale datasets by estimating the distance of an input instance $I$ to the closest decision boundary. The iterative method estimates the perturbation $\delta_{i}$ 
% \nael{$\delta_{t}$ to maintain consistent iterator?} 
at each iteration $i$ till the classifier $f(I_{i})$
% \nael{why are iteration definitions inconsistent here w.r.t. PGD. $I^{Adv}_{t+1}$  was used and $t$ was the iterator. Also ``i'' was used in CW to define the input class. We must maintain consistency among definitions} 
% \lukas{Fixed: PGD iteration is now i. Should be now consistent with DeepFool @Manoj: Could you verify consistency with C&W? checked and changed the class iterations from i to t}
changes its prediction ($f(I_{i}) \neq L$). In practice, once an adversarial perturbation $\delta$ is found, the adversarial example is pushed further beyond the decision boundary.
The algorithm is not guaranteed to converge to the optimal perturbation, nevertheless it generates adversarial examples with good approximations of the minimal perturbation. The size of the calculated perturbation can also be interpreted as a metric for the model's robustness against adversarial attacks~\cite{surveyaa}. % The authors also show that the perturbation of DeepFool is smaller than the perturbation generated by FGSM.

% In our experiment, we use multiple class DeepFool for testing. So I would like just to briefly explain the algorithm for multiple class perturbation:
% \begin{figure}
%     \centering
%     \includegraphics[scale=0.01]{data/Photos/DeepFool_multi_al.png
%     \caption{Deepfool Algorithm for Multiple Classes \cite{Deepfool}}
%     \label{fig:deepfool}
% \end{figure}

%In the algorithm, $\hat{k}(\boldsymbol{x_{i}})$ represents the current predicted class of perturbed input image $\boldsymbol{x_{i}}$, and $\hat{k}(\boldsymbol{x_{0}})$ denotes the real class of original input image. In the for loop, $\boldsymbol{\omega_{k}^{'}}$ accumulates all difference between the gradient of cost w.r.t. class $k$ to the perturbed image as well the gradient of cost w.r.t. real class $\hat{k}(\boldsymbol{x_{0}})$ to the perturbed image. At the same time, difference between costs w.r.t. class $k$ and real class $\hat{k}(\boldsymbol{x_{0}})$ is saved in $f_{k}^{'}$. $k$ represents all each class different from real class. In the line 10., the nearest class $\hat{l}$ will be found by computing the smallest norm of $\boldsymbol{x_{i}}$. Then we can get the smallest effective perturbation $\boldsymbol{r_{i}}$. After the whole while-loop, the sum of each step perturbation $\boldsymbol{r_{i}}$ is the final perturbation $\hat{\boldsymbol{r}}$.

    \subsection{Black-box Attacks}\label{subsec:BlackBox}
        % Local Search

%\textit{Local-Search}~\cite{LocalSearch} is a simple gradient-free adversarial black-box attack which is designed based on the idea of randomly perturbation and ``greedy search algorithm'' around perturbed pixels.
\noindent
\textbf{LocalSearch: }
LocalSearch~\cite{LocalSearch} is a simple gradient-free adversarial black-box attack, which is based on random perturbation and a \emph{greedy search algorithm} around the perturbed pixels.
%\textit{Local-Search} procedure works in iterations and each iteration consists of two steps. The first step is to select a small subset of points $Z = \{ \boldsymbol{\hat{z}}_{1}, ..., \boldsymbol{\hat{z}}_{n} \}$, which is called \textit{local neighborhood}, and is then evaluated with objective function $f(\boldsymbol{\hat{z}}_{j})$ for every $\boldsymbol{\hat{z}}_{j} \in Z$. The set $Z$ consists of all points that locates around the selected/perturbed pixels $\boldsymbol{\hat{z}}_{i-1}$ from last iteration and they are all domain specific. In the second step, a new group of pixels $\boldsymbol{\hat{z}}_{i}$ are selected by taking the previous solution $\boldsymbol{\hat{z}}_{i-1}$ and the points in Z into account. 
%The LocalSearch procedure works in iterations, where each iteration consists of two steps. The first step is to select a small subset of points $P_{i} = \{ \boldsymbol{\hat{p}}_{1}, ..., \boldsymbol{\hat{p}}_{n} \}$, which is referred to as the \textit{local neighborhood}. The local neighborhood is then evaluated with objective function $f(\boldsymbol{\hat{p}}_{j})$ for every $\boldsymbol{\hat{p}}_{j} \in P_i$. The set $P_i$ consists of points located around $\boldsymbol{p}_{i-1}$, which is the solution of the last iteration. In the second step, a new solution $\boldsymbol{p}_{i}$ is selected by taking the previous solution $\boldsymbol{p}_{i-1}$ and the points in $P_i$ into account. 
The LocalSearch procedure works in iterations, where each iteration consists of two steps. The first step is to select and evaluate a small subset of points $P_{i}$, referred to as the \textit{local neighborhood}. In the second step, a new solution $P_{i+1}$ is selected by taking the evaluation of the previous solution $P_i$ into account. 
LocalSearch is simple to implement, but is computationally expensive, similar to most greedy search algorithms. % The poor scalability of the attack with larger input images can make the algorithm prohibitively slow and inefficient.
        % \subsubsection{Rotation}
        % Gen-Attack

% For strict Black-Box Attack, the attacker is restricted solely to query access. Existing Black-Box approaches like Local-Search or ZOO\cite{ZOO} requires lots of query times or much larger computation memory than white-box to generate adversarial perturbation. For example for Local-Search that we have introduced above, if we want to attack one image with 16 pixels in each iteration and image batch is 128, input size of model is [16*128, h, w, d]. It requires much more GPU memory to finish. Therefore, Moustafa Alzantot et. al \cite{GenAttack} has develop one new Black-Box Attack for reducing the query times and memory use and being able to get almost the same result like ZOO method.

%\textit{Gen-Attack}\cite{GenAttack} is inspired by \textit{Genetic Algorithm} similar like natural evolution, and it belongs to the group of gradient-free optimization strategies. The initial populations of perturbed image examples are generated by adding uniform random noise. The best population will be then derived by iterative running \textit{Fitness} evaluation, \textit{Selection}, \textit{Crossover} and \textit{Mutation}.
\noindent
\textbf{GenAttack: }
GenAttack~\cite{GenAttack} is a gradient-free optimization strategy based on a genetic algorithm. The initial population of perturbed image examples is generated by adding uniform random noise. The best individuals survive the generation based on their fitness evaluation, the selection strategy and the crossover and mutation probabilities.
% a couple of random examples with uniform distribution are picked and applied on the given input image $x_{orig}$. This uniform distribution is defined over the sphere centered at the original example $x_{orig}$ whose radius = $\delta_{max}$ and the random noise will be generated in range $(- \delta_{max}, \delta_{max})$. After repeatedly adding the random noise on $x_{orig}$, the initial population $\mathcal{P}^{0}$ is born. Then we start to run multiple generations $G$ to finish the evolution progress of genetic algorithm. 
%\textit{Fitness Evaluation} essentially reflects the optimization objective. And \textit{Selection} gives the elite in all populations based on fitness evaluation scores and provides the selection probabilities generating new children perturbation in \textit{Crossover}. After that, \textit{Mutation operation} will be applied on the children adversarial populations from Crossover for exploring some better potential. The elite population and mutated children will be then saved as the start populations for next generation. 
%Fitness evaluation reflects the optimization objective, while the selection strategy allows elite individuals in the population to generate new children perturbations through a crossover mechanism. A mutation operation follows, and is applied on the children produced from the crossover operation. This allows the algorithm to explore potentially better solutions, outside of the elite population. The elite population and mutated children are then evaluated for the next generation. 
Fitness evaluation reflects the optimization objective, while the selection strategy allows elite individuals in the population to generate new children perturbations through crossover and mutation mechanisms.
GenAttack is a faster search algorithm when compared to LocalSearch~\cite{LocalSearch}, and generates perturbations which are imperceptible to the human eye.
% \begin{equation}
%     p = \frac{fitness(parent_{1})}{fitness(parent_1) + fitness{(parent_2)}}
% \end{equation}

%  ___      __   ___  __           ___      ___  __  
% |__  \_/ |__) |__  |__) |  |\/| |__  |\ |  |  /__` 
% |___ / \ |    |___ |  \ |  |  | |___ | \|  |  .__/ 
\section{Breaking Binary and Efficient DNNs}
    \label{sec:experiments}
    Although a successful attack could easily be carried out by adding large perturbations, the requirement of finding the minimum necessary perturbation in each case is typically desirable to perform the attack in an inconspicuous manner. 
This justifies CNNs to being particularly robust against adversarial attacks that are relevant or expected in practice. 
However, despite the requirement to keep the perturbation as small as possible, the target for training against an attack structure can be to maximize a corresponding loss function. 
A prior analysis on the robustness of real world compressed CNNs provides insights which facilitate the realization of strong adversarial defense methods. 
%improving the robustness is a wise step. 

We evaluate robustness of CNNs which are trained and evaluated on CIFAR-10~\cite{cifar10} or ImageNet~\cite{ILSVRC15} datasets.
% and CityScapes~\cite{CityScapes} for the semantic segmentation task. 
The 50K train and 10K test images (\(32\times32\) pixels) of CIFAR-10 are used to train and evaluate compressed variants of ResNet20/56. \cite{He2015DeepRL,contrastive_kd,AMC2019,rastegari2016,lin2017} respectively. % The images have a resolution of \(32\times32\) pixels. 
The ImageNet dataset consists of $\sim$1.28M train and 50K validation images ($256\times256$ pixels). % with a resolution of $256\times256$ pixels. 
Compressed variants of ResNet18/50 are trained and evaluated for ImageNet experiments. If not otherwise mentioned, all hyper-parameters specifying the training and the attacks were adopted from the reference implementation. 
%We report the prediction accuracy, normalized compute complexity and memory demand for each variant in the supplementary material (Tab.~\textcolor{red}{S1}).
The robustness evaluation covers various white-box (FGSM, PGD, C\&W, DeepFool) and black-box (LocalSearch, GenAttack) attacks on the CIFAR-10-trained ResNet20/56 compressed variants, as well as ImageNet-trained CNNs.   
% $\epsilon$ and $i$ represent the adversarial manipulation budget and iteration count.
%The tables summarizing all the experiments can be found in Tab.~\textcolor{red}{S2}-\textcolor{red}{S17}. 
%\nael{Check if correct after supp is done} in the supplementary material.
%For C\&W attacks (Tab.~\textcolor{red}{S8}, \textcolor{red}{S9}),
%\nael{Check if correct after supp is done}, 
%we use the target class $t=deer$ 
% \nael{why no use this in CW introduction paragraph instead of $t$} 
%from  CIFAR-10 dataset to fool the CNN. % and $c$ represents the regularization factor in Eq.~\ref{eq:CW}. 
%For GenAttacks on CIFAR-10 experiments, the mutation probability $\rho$ and step size $\alpha$ are set to $0.05$ and $1.0$ respectively as demonstrated in ~\cite{GenAttack}.
%Parent selection probability is based on sampling according to the softmax function applied to the fitness values of the population. 
%Elitism is applied for the highest fitness member. 
%On ImageNet experiments $\rho$ and $\alpha$ are adaptive based on the dataset configuration and the population size is 6, as in~\cite{GenAttack}. 
We perform all the experiments using the trained statistics for the batch normalization layers. 
% \manoj{Comment about the BN mode for all the experiments.}
    \subsection{CNN Compressed Variants}
    Tab.~\ref{tab:compressed_cnns} summarizes the compressed CNNs and their full-precision counterparts analyzed in this paper. It shows that the neural networks drastically vary in their memory demand and their compute complexity. Deep learning inference accelerators such as the NVIDIA-T4 GPU~\cite{nvidiat4} or Xilinx FPGAs with DSP48 blocks support SIMD-based bit-wise operations~\cite{orthruspe}. In particular, a single DSP48 block can perform two 16-bit fixed-point multiplications or 48 XNOR operations at once. 
% % Thus,24 XNOR operations can be normalized as one fixed point multiplication. 
The normalized compute complexity (NCC) is defined as the optimal utilization of MAC and XNOR operations in one compute unit. The DSP48 block serves as a reference implementation to compute NCC in Tab.~\ref{tab:compressed_cnns}.
%~\cite{nvidia_t4}
%~\cite{XilDSP48E1}
\begin{table*}[ht]
  \centering
  \resizebox{0.70\textwidth}{!}{
  \renewcommand{\arraystretch}{1.0}
  \begin{tabular}{ll|ccc} 
    \toprule
    \textbf{Dataset}&\textbf{Model} & \textbf{Acc. [\%]} & \textbf{Memory demand [MB]} & \textbf{NCC} [$10^6$]\\
    \midrule
    \midrule
    \multirow{23}{*}{\rotatebox{90}{\makecell{CIFAR-10}}} 
    &\textbf{ResNet20}~\cite{He2015DeepRL}           &92.46 \%   &1.07   &41\\
    %&\textbf{KD-CE}~\cite{contrastive_kd}           &92.70 \%   &1.07   &41\\
    &\textbf{KD-KL}~\cite{contrastive_kd}        &93.25 \%   &1.07   &41\\
    &\textbf{Ch.Prune}~\cite{AMC2019}           &89.76 \%   &0.70   &19\\
    %&\textbf{F.Prune}~\cite{AMC2019}            &89.91 \%   &0.82   &21\\
    &\textbf{K.Prune}~\cite{AMC2019}            &90.73 \%   &0.61   &20\\
    &\textbf{W.Prune}~\cite{AMC2019}            &91.98 \%   &0.59   &20\\
    &\textbf{XNOR}~\cite{rastegari2016}              &82.71 \%   &0.04   &1.3\\
    &\textbf{ABC(1$\times$1)}~\cite{lin2017}         &83.42 \%   &0.04   &1.3\\
    &\textbf{ABC(3$\times$3)}~\cite{lin2017}         &88.94 \%   &0.12   &8.0\\
    &\textbf{ABC(5$\times$5)}~\cite{lin2017}         &90.64 \%   &0.20   &21.3\\
    %&\textbf{ABC(7$\times$7)}~\cite{lin2017}         &91.34 \%   &0.28   &41.3\\
    \cmidrule{2-5}
    &\textbf{ResNet56}~\cite{He2015DeepRL}           &93.88 \%   &3.40   &125\\
    %&\textbf{KD-CE}~\cite{contrastive_kd}           &93.68 \%   &3.40   &125\\
    &\textbf{KD-KL}~\cite{contrastive_kd}        &94.24 \%   &3.40   &125\\
    &\textbf{Ch.Prune}~\cite{AMC2019}           &92.86 \%   &2.50   &62\\
    %&\textbf{F.Prune}~\cite{AMC2019}            &93.09 \%   &2.29   &64\\
    &\textbf{K.Prune}~\cite{AMC2019}            &93.04 \%   &2.19   &63\\
    &\textbf{W.Prune}~\cite{AMC2019}            &93.54 \%   &2.02   &62\\
    &\textbf{XNOR}~\cite{rastegari2016}              &83.24 \%   &0.11   &3.0\\
    &\textbf{ABC(1$\times$1)}~\cite{lin2017}         &86.29 \%   &0.11   &3.0\\
    &\textbf{ABC(3$\times$3)}~\cite{lin2017}         &92.48 \%   &0.33   &24\\
    &\textbf{ABC(5$\times$5)}~\cite{lin2017}         &92.82 \%   &0.55   &66\\
    \midrule
    \multirow{6}{*}{\rotatebox{90}{\makecell{ImageNet}}}
    &\textbf{ResNet50}~\cite{He2015DeepRL}           &75.43 \%   &102.01 &10216\\
    &\textbf{ResNet18}~\cite{He2015DeepRL}           &69.00 \%   &46.72  &1814\\
    &\textbf{ResNet18-Ch.Prune}~\cite{AMC2019}  &67.62 \%   &34.52  &884\\
    &\textbf{ResNet18-XNOR}~\cite{rastegari2016}     &49.10 \%   &4.14   &173\\
    &\textbf{ResNet18-ABC(1$\times$1)}~\cite{lin2017}&51.07 \%   &3.48   &153\\
    &\textbf{ResNet18-ABC(3$\times$3)}~\cite{lin2017}&59.83 \%   &6.28   &417\\
    \bottomrule 
    \end{tabular}}
    \vspace{1em}
    \caption{Accuracy Top1 [\%], Memory demand [MB] and the normalized compute complexity (NCC) of compressed CNNs and their full-precision counterparts.}
  \label{tab:compressed_cnns}
\end{table*}

    \subsection{Evaluation of Robustness}
    \iffalse
Propose a taxonomy for robustness properties, inspired by definitions of \textit{\textbf{breaking}} in material science:
\begin{itemize}
    \item Stiff - resists deformation
    \item Tough - resists failure, even after deforming
    \item Strong - resists both deformation and failure
    \item Ductile - deforms before it breaks
    \item Brittle - breaks before it deforms
    \item Hard - resists dents, scratches, and other permanent changes under compressive force
\end{itemize}
\fi
\noindent
\textbf{PGD-Evaluation:}
% Justification for Experiment
Considering PGD attack as the ``ultimate'' first-order attack, this section experimentally explores the structure of the loss surfaces and the corresponding accuracy deterioration of the proposed efficient DNNs, while exposing the models to PGD adversaries, similar to those proposed by Madry et al.~\cite{Madry2018PGD}.
Investigating the resulting structural behavior, especially the loss level to which the PGD attack is converging to and the speed of deterioration of accuracy, helps in understanding the adversarial robustness of the underlying models with respect to a defined PGD threat model $\tau_{PGD}$=$\{$ $\epsilon$, $\alpha$, $i$ $\}$. 
%\nael{we never introduced/defined loss niveau. Also, why not call it loss levels simply?} \lukas{Will call it loss level now.}
%\nael{We never mentioned thread model until here. Maybe worth a sentence or two in the introduction section? or here?}
%This set of experiment exploits the appealing characteristics of PGD attacks, as already described in Subsec.~\ref{subsec:WhiteBox}, to gain an intuition about the adversarial robustness of varying models trained specifically on CIFAR-10, that are exposed to the first-order attack PGD. 

% Experimental Setup
All models are pre-trained on CIFAR-10 without any adversarial examples, to distinguish the influence of varying compression techniques on adversarial robustness. 
Subsequently, each model is exposed to PGD attacks from $\tau_{PGD}$=$\{ \epsilon$=2, $~\alpha$=0.5, $~i$=1000$\}$.
%\nael{Ah, I see it here now. Maybe move this earlier: .. with regard to some PGD thread model $\tau = \{ \epsilon, \alpha, i\}$, where $\epsilon = 2$, $\alpha = 0.5$...}. 
Following the method of Carlini et al.~\cite{EvaluatingAdvRobust2019}, $i$ was increased to verify convergence, ensuring local-maxima, representing potentially worst-case adversarial examples for the underlying model with respect to the applied threat model $\tau_{PGD}$. However, results are only shown up to $i$=100, since $\tau_{PGD}$ showed convergence for all investigated models in this range.
%\alex{Also: Detecting high perturbation attacks can be easily done by reactive filtering methods} \manoj{In this analysis, we use fixed $\epsilon$=2. Strength is same. However, we can use this point in stress strain graphs as we play with varying strength. }
The loss value and the corresponding accuracy of the models to the adversary were tracked every $5^{th}$-iteration.
%Over the range of iterations, the experiment showed convergence of loss and accuracy of PGD over all models, that ensures finding local-maxima representing potential worst-case adversarial examples for the specific model. 
% Old Plots with Box-Plot
%\input{img/PGD_Loss_Eval/PGD_Loss_Eval.tex}
%1.Axis: Loss, 2. Axis: Acc
%\input{img/PGD_Loss_Eval/PGD_Loss_Acc.tex}
% 1. Axis: Acc 2. Axis: Loss
\input{img/PGD_Loss_Eval/PGD_Acc_Loss.tex}
% Justification for tracking Loss and Accuracy.
In the following, the adversarial robustness of a model against PGD attacks is evaluated by (1) the overall loss level the PGD attack is converging to and as a consequence the resulting accuracy (2) the number of iterations a model can sustain until it breaks. We can consider a CNN model broken, if its accuracy indicates that the classification is random (10\% for CIFAR-10 dataset), represented by model accuracy graphs dropping below the breaking line (BL). 
%\lukas{Maybe remove 0.1\% for ImageNet since we never used it}).
% However, depending on the application, the term can be used at a higher threshold.
% \manoj{Lets define this as the breaking point.} 
% Justification for five reruns. 
Fig.~\ref{fig:pgd_eval} shows the mean over five reruns of PGD attack for all models to exploit random initialization, which ensures random exploration of the underlying non-concave maximization problem as described in Sec.~\ref{sec:adversarial_attacks}.

% Observations
Consistently, all investigated pruning techniques harm adversarial robustness against PGD attack with respect to its vanilla versions of ResNet20/56, when considering (1) the loss and accuracy after a converged attack and (2) the speed of breaking. Vanilla and pruned versions of ResNet20 break within five iterations, whereas the respective ResNet56 versions break within ten iterations.
%- $i_{\text{break}}\leq 5$ for ResNet20 versions and $i_\text{break} \leq 10$ for ResNet56 versions. 
KD shows greater resilience to the PGD attack since (1) its accuracy after the converged attack is higher compared to both the ResNet20/56 vanilla variants and (2) breaking at a higher number of iterations. KD-KL breaks at $i$=15 for its ResNet20 variant and at $i$=40 at its ResNet56 variant.
Binarization can improve the robustness against the defined PGD attack, materializing in (1) the higher loss and accuracy after a converged attack and (2) the greater resilience for a longer period of PGD iterations. XNOR-Net and ABC(5$\times$5) break at $i$=20, while ABC(3$\times$3) and ABC(1$\times$1) break at around $i$=60 for their ResNet20 variants. For the ResNet56 variants, ABC(1$\times$1) and ABC(5$\times$5) break at $i$=20, whereas ABC(3$\times$3) sustains up to $i$=40. The ResNet56 variant of XNOR-Net outperforms all other models in (1) accuracy after converged attack ($\sim$14\%) and (2) being the only model that never breaks throuhgout this experiment (see Fig.~\ref{fig:pgd_eval} right).
% Explanation
%\manoj{Comment on any correlation between robustness and prune regularity}. 

\noindent
\textbf{Stress-Strain Evaluation:}
To facilitate the interpretation of the data generated from the experiments, we propose a
% \manoj{this is the only method we propose}
method for evaluating robustness. Different models such as ResNet20 and ResNet56 have different baseline accuracies, making it difficult to directly compare the robustness of different training or compression schemes. Existing metrics, such as attack success-rate \cite{GenAttack} or accuracy degradation, fail to capture the differences of the baseline accuracy of a network. Taking inspiration from the field of mechanics, we use formulas of stress and strain to make an analogy with the robustness of networks before they \textit{break}.
% We can consider a CNN broken, 
% \manoj{We need to make the defination of the "broken" clear} 
% \nael{agreed, @lukas the bound of random guessing has a formal def?} 
% if its accuracy indicates that the classification is random. However, depending on the application, the term can be used at a higher threshold. \manoj{break definition added it to PGD evaluation }
Applying a certain amount of stress on an object causes a certain measure of deformity or strain. We adapt the strain formula to our problem as $\varepsilon$ = $\frac{\mathcal{A}-\mathcal{A}^*}{\mathcal{A}}$,
%\nael{sorry, didnt notice lukas' comment}
%\alex{@Nael why no comma after equation?}
%\lukas{Add ", " after formula since equations are part of the sentence }
%\lukas{Give intuition about behavior of NN w.r.t. $\boldsymbol{\varepsilon}$, e.g. small values of $\boldsymbol{\varepsilon}$ means that NN is able to maintain accuracy under attacks and large values represent degradation of accuracy. }
where $\boldsymbol{\varepsilon}$ is the strain, $\mathcal{A}$ is the accuracy before attack and $\mathcal{A}^*$ the deteriorated accuracy. Note that, we use  $\epsilon$ and $\boldsymbol{\varepsilon}$ to represent perturbation amplitude and strain respectively. 
%\manoj{Conflict: we use $\epsilon$ for both strength and strain} 
A network which sustains higher strain $\boldsymbol{\varepsilon}$ w.r.t. an attack is less robust. The rate of change in $\boldsymbol{\varepsilon}$ with increased stress indicates the resilience or fragility of the CNN under heavier forms of the same attack.
Similar to the different types of mechanical stress (compressive, tensile or shear), iterative and amplitude based attacks can represent different types of attack-stress $\sigma$. 

%In mechanics, Young's modulus $\mathcal{Y}$ refers to the degree of stiffness of the material, i.e. the amount of deformity it sustains under a certain amount of stress in the elastic region. The robustness of the network at a specific stress-type and stress-level, is then measured by an analogy of Young's modulus:

%\begin{equation}
%    \mathcal{Y} = \frac{\sigma}{\boldsymbol{\varepsilon}}
%\end{equation}

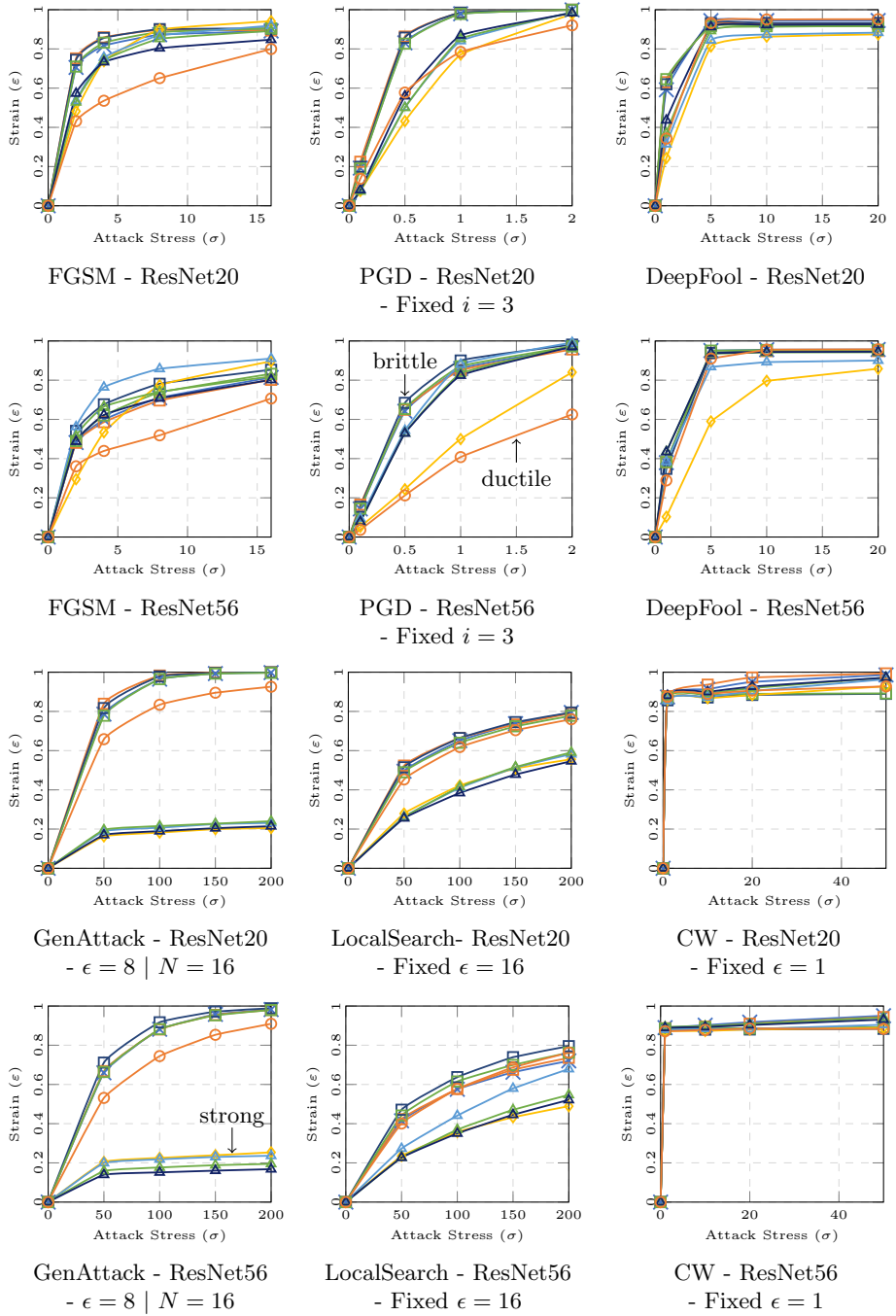
\begin{figure*}
\captionsetup[subfigure]{justification=centering, labelformat=empty}
%\centering
\subfloat{
%\hspace{10ex}
       \begin{tikzpicture}	
	\begin{axis}[
		name=legend,
		width=\textwidth,
		height=20ex,
		hide axis,
        legend style={at={(0,0)},anchor=west, legend columns=-1, draw = none, nodes={scale=0.75, transform shape}, column sep=2pt},
		ymin=0,
        ymax=1,
        xmin=0,
        xmax=1,
		]
		\addlegendimage{Pal_1, mark=x,mark options={scale=1.25}, line width=1pt, only marks}
       	\addlegendentry{Vanilla};
       	\addlegendimage{Pal_2, mark=square,mark options={scale=1.25}, line width=1pt, only marks}
		\addlegendentry{Ch.Prune};
		%\addlegendimage{Pal_7, mark=square,mark options={scale=1}, line width=1pt, only marks}
		%\addlegendentry{F.Prune};
		\addlegendimage{Pal_7,mark=square,mark options={scale=1.25}, line width=1pt, only marks}
        \addlegendentry{K.Prune};
        \addlegendimage{Pal_6,mark=square,mark options={scale=1.25}, line width=1pt, only marks}
		\addlegendentry{W.Prune};
		\addlegendimage{Pal_4, mark=diamond,mark options={scale=1.25}, line width=1pt, only marks}
		\addlegendentry{XNOR};
		\addlegendimage{Pal_5, mark=triangle,mark options={scale=1.25}, line width=1pt, only marks}
		\addlegendentry{ABC($1\times1$)};
		\addlegendimage{Pal_6,mark=triangle,mark options={scale=1.25}, line width=1pt, only marks}
		\addlegendentry{ABC($3\times3$)};
		\addlegendimage{Pal_8,mark=triangle,mark options={scale=1.25}, line width=1pt, only marks}
		\addlegendentry{ABC($5\times5$)};
		%\addlegendimage{Pal_3,mark=triangle,mark options={scale=1}, line width=1pt, only marks }
		%\addlegendentry{ABC($7\times7$)};
		%\addlegendimage{Pal_1 ,mark=o,mark options={scale=1}, line width=1pt, only marks}
		%\addlegendentry{KD-CE};
		\addlegendimage{Pal_2,mark=o,mark options={scale=1.25}, line width=1pt, only marks}
		\addlegendentry{KD-KL};
	\end{axis}
\end{tikzpicture}
} \\\vspace{-2ex}
\subfloat[FGSM - ResNet20]{
       \pgfplotstableread[col sep=space]{%
epsilon	Vanilla KDCE KDKL AMCCH AMCF AMCK AMCW XNOR ABC1 ABC2 ABC3 ABC4
0   92.46	92.70   93.25   89.76   89.91 90.73 91.99  82.71	83.42   88.94   90.64   91.34
2	27.21   26.56   53.01   22.31   21.53 23.21 26.68  42.84   38.78   41.98   38.70   40.11
4	16.96   16.36   43.30   12.63   11.69 13.10 15.46  21.71   20.32   22.99   24.18   27.45
8	11.79   11.73   32.54   9.29    8.45  8.99  10.49  8.40   10.93   13.20   17.78   20.94
16	10.04   9.95    18.69   9.50    8.14  8.72  8.77   4.85   6.80    9.62    13.86   16.61
}\fgsmresnettwentycifar

\begin{tikzpicture}
    \begin{axis}[
          height=0.22\textwidth,
          width=0.25\textwidth, % Scale the plot to \linewidth
          grid=major, 
          grid style={dashed,gray!30},
          xlabel= Attack Stress ($\sigma$), % Set the labels
          ylabel= Strain ($\varepsilon$),
          xmin=0,
          xmax=16,
          ymin=0,
          ymax=1,
          smooth,
          tension=0.2,
          yticklabel style = {font=\tiny, rotate=90, yshift=-0.3ex},
          xticklabel style = {font=\tiny},
		  ylabel style = {font=\tiny, yshift=-5.75ex},
		  xlabel style = {font=\tiny, yshift=2ex},
		  legend style={at={(1,0.32)},anchor=east, legend columns=1, fill=none ,draw = none, nodes={scale=0.3, transform shape}, column sep=2pt, mark size=1.5pt},
		  scale only axis,
		  every axis plot/.append style={line width=0.7pt, mark options={scale=1}},
          cycle list name=mycolorlist,
        ]
       \addplot table[x index = 0, y expr = {(92.46-\thisrow{Vanilla})/92.46}] {\fgsmresnettwentycifar};
       
       \addplot table[x index = 0, y expr = {(89.76-\thisrow{AMCCH})/89.76}] {\fgsmresnettwentycifar}; 
       
       %\addplot table[x index = 0, y expr = {(89.91-\thisrow{AMCF})/89.91}] {\fgsmresnettwentycifar};  
       
       \addplot table[x index = 0, y expr = {(90.73-\thisrow{AMCK})/90.73}] {\fgsmresnettwentycifar};  
       
       \addplot table[x index = 0, y expr = {(91.99-\thisrow{AMCW})/91.99}] {\fgsmresnettwentycifar};  
       
       \addplot table[x index = 0, y expr = {(82.71-\thisrow{XNOR})/82.71}] {\fgsmresnettwentycifar};  
       
       \addplot table[x index = 0, y expr = {(83.42-\thisrow{ABC1})/83.42}] {\fgsmresnettwentycifar};  
       
       \addplot table[x index = 0, y expr = {(88.94-\thisrow{ABC2})/88.94}] {\fgsmresnettwentycifar};  
       
       \addplot table[x index = 0, y expr = {(90.64-\thisrow{ABC3})/90.64}] {\fgsmresnettwentycifar};  
       %\pgfplotsset{cycle list shift=3}
       %\addplot table[x index = 0, y expr = {(91.34-\thisrow{ABC4})/91.34}] {\fgsmresnettwentycifar};  
       
       %\addplot table[x index = 0, y expr = {(92.70-\thisrow{KDCE})/92.70}] {\fgsmresnettwentycifar};  
       
       \addplot table[x index = 0, y expr = {(93.25-\thisrow{KDKL})/93.25}] {\fgsmresnettwentycifar}; 
       
       	\addlegendentry{Vanilla};
		\addlegendentry{AMC-CH};
		%\addlegendentry{AMC-F};
        \addlegendentry{AMC-K};
		\addlegendentry{AMC-W};
		\addlegendentry{XNOR};
		\addlegendentry{ABC($1\times1$)};
		\addlegendentry{ABC($3\times3$)};
		\addlegendentry{ABC($5\times5$)};
		%\addlegendentry{ABC($7\times7$)};
		%\addlegendentry{KD-CE};
		\addlegendentry{KD-KL};
		\legend{};
    \end{axis}

\end{tikzpicture} 
       }
\subfloat[PGD - ResNet20\\ - Fixed $i=3$]{
       \pgfplotstableread[col sep=space]{%
epsilon	Vanilla KDCE KDKL AMCCH AMCF AMCK AMCW XNOR ABC1 ABC2 ABC3 ABC4
0   92.46	92.70   93.25	89.76	89.91   90.73  91.99  82.71	83.42   88.94   90.64   91.34
0.1	74.69   74.24   80.39   69.60   70.86   72.69  74.49  76.47   76.48   82.40   83.35   83.32
0.5	15.61   14.12   39.40   12.30   13.57   12.93  15.92  47.01   41.54   44.47   39.93   38.76
1	1.92    1.97    20.06   1.41    1.49    1.36   1.97  18.93   13.27   13.19   11.82   12.79
2	0.09    0.11    7.42    0.04    0.07    0.04   0.04  1.86    1.03    1.37    1.54    1.82
}\pgdresnettwentycifarfixediter

\begin{tikzpicture}
    \begin{axis}[
          height=0.22\textwidth,
          width=0.25\textwidth, % Scale the plot to \linewidth
          grid=major, 
          grid style={dashed,gray!30},
          xlabel= Attack Stress ($\sigma$), % Set the labels
          ylabel= Strain ($\varepsilon$),
          xmin=0,
          xmax=2,
          ymin=0,
          ymax=1,
          smooth,
          tension=0.2,
          yticklabel style = {font=\tiny, rotate=90, yshift=-0.3ex},
          xticklabel style = {font=\tiny},
		  ylabel style = {font=\tiny, yshift=-5.75ex},
		  xlabel style = {font=\tiny, yshift=2ex},
		  legend style={at={(1,0.32)},anchor=east, legend columns=1, fill=none ,draw = none, nodes={scale=0.45, transform shape}, column sep=5pt},
		  scale only axis,
		  every axis plot/.append style={line width=0.7pt, mark options={scale=1}},
          cycle list name=mycolorlist,
        ]
       \addplot table[x index = 0, y expr = {(92.46-\thisrow{Vanilla})/92.46}] {\pgdresnettwentycifarfixediter};
       
       \addplot table[x index = 0, y expr = {(89.76-\thisrow{AMCCH})/89.76}] {\pgdresnettwentycifarfixediter};  
       
       %\addplot table[x index = 0, y expr = {(89.91-\thisrow{AMCF})/89.91}] {\pgdresnettwentycifarfixediter};  
       
       \addplot table[x index = 0, y expr = {(90.73-\thisrow{AMCK})/90.73}] {\pgdresnettwentycifarfixediter};  
       
       \addplot table[x index = 0, y expr = {(91.99-\thisrow{AMCW})/91.99}] {\pgdresnettwentycifarfixediter}; 
       
       \addplot table[x index = 0, y expr = {(82.71-\thisrow{XNOR})/82.71}] {\pgdresnettwentycifarfixediter};  
       
       \addplot table[x index = 0, y expr = {(83.42-\thisrow{ABC1})/83.42}] {\pgdresnettwentycifarfixediter};  
       
       \addplot table[x index = 0, y expr = {(88.94-\thisrow{ABC2})/88.94}] {\pgdresnettwentycifarfixediter};  
       
       \addplot table[x index = 0, y expr = {(90.64-\thisrow{ABC3})/90.64}] {\pgdresnettwentycifarfixediter};  
       %\pgfplotsset{cycle list shift=3}
       %\addplot table[x index = 0, y expr = {(91.34-\thisrow{ABC4})/91.34}] {\pgdresnettwentycifarfixediter};  
       
       %\addplot table[x index = 0, y expr = {(92.70-\thisrow{KDCE})/92.70}] {\pgdresnettwentycifarfixediter};  
       
       \addplot table[x index = 0, y expr = {(93.25-\thisrow{KDKL})/93.25}] {\pgdresnettwentycifarfixediter};   
       
       	\addlegendentry{Vanilla};
		\addlegendentry{AMC-CH};
		%\addlegendentry{AMC-F};
		\addlegendentry{AMC-K};
		\addlegendentry{AMC-W};
		\addlegendentry{XNOR};
		\addlegendentry{ABC($1\times1$)};
		\addlegendentry{ABC($3\times3$)};
		\addlegendentry{ABC($5\times5$)};
		%\addlegendentry{ABC($7\times7$)};
		%\addlegendentry{KD-CE};
		\addlegendentry{KD-KL};
		\legend{};

    \end{axis}

\end{tikzpicture} 
       }
\subfloat[DeepFool - ResNet20]{
       \pgfplotstableread[col sep=space]{%
iter	Vanilla KDCE KDKL AMCCH AMCF AMCK AMCW XNOR ABC1 ABC2 ABC3 ABC4
0   92.46	92.70   93.25	89.76	89.91   90.73  91.99	82.71	83.42   88.94   90.64   91.34
1	37.75   38.10   61.28   33.18   32.23   34.50  32.58   62.59   57.17   55.98   51.10   48.26
5	5.15    4.78    6.35    6.79    6.39    6.19   7.75   15.51   13.00   9.60    7.96    6.96
10	5.11    4.75    4.64    6.74    6.27    6.16   7.75   11.38   10.59   7.73    6.99    6.39
20	5.09    4.80    4.53    6.66    6.25    6.15   7.73   10.40   9.77    7.39    6.80    6.31    
}\deepfoolresnettwentycifar

\begin{tikzpicture}
    \begin{axis}[
          height=0.22\textwidth,
          width=0.25\textwidth, % Scale the plot to \linewidth
          grid=major, 
          grid style={dashed,gray!30},
          xlabel= Attack Stress ($\sigma$), % Set the labels
          ylabel= Strain ($\varepsilon$),
          xmin=0,
          xmax=20,
          ymin=0,
          ymax=1,
          smooth,
          tension=0.2,
          yticklabel style = {font=\tiny, rotate=90, yshift=-0.3ex},
          xticklabel style = {font=\tiny},
		  ylabel style = {font=\tiny, yshift=-5.75ex},
		  xlabel style = {font=\tiny, yshift=2ex},
		  legend style={at={(1,0.24)},anchor=east, legend columns=1, fill=none ,draw = none, nodes={scale=0.38, transform shape}, column sep=5pt},
		  scale only axis,
		  every axis plot/.append style={line width=0.7pt, mark options={scale=1}},
          cycle list name=mycolorlist,
        ]
       \addplot table[x index = 0, y expr = {(92.46-\thisrow{Vanilla})/92.46}] {\deepfoolresnettwentycifar};
       
       \addplot table[x index = 0, y expr = {(89.76-\thisrow{AMCCH})/89.76}] {\deepfoolresnettwentycifar};  
       
       %\addplot table[x index = 0, y expr = {(89.91-\thisrow{AMCF})/89.91}] {\deepfoolresnettwentycifar};  
       
       \addplot table[x index = 0, y expr = {(90.73-\thisrow{AMCK})/90.73}] {\deepfoolresnettwentycifar};  
       
       \addplot table[x index = 0, y expr = {(91.99-\thisrow{AMCW})/91.99}] {\deepfoolresnettwentycifar}; 
       
       \addplot table[x index = 0, y expr = {(82.71-\thisrow{XNOR})/82.71}] {\deepfoolresnettwentycifar};  
       
       \addplot table[x index = 0, y expr = {(83.42-\thisrow{ABC1})/83.42}] {\deepfoolresnettwentycifar};  
       
       \addplot table[x index = 0, y expr = {(88.94-\thisrow{ABC2})/88.94}] {\deepfoolresnettwentycifar};  
       
       \addplot table[x index = 0, y expr = {(90.64-\thisrow{ABC3})/90.64}] {\deepfoolresnettwentycifar};  
       %\pgfplotsset{cycle list shift=3}
       %\addplot table[x index = 0, y expr = {(91.34-\thisrow{ABC4})/91.34}] {\deepfoolresnettwentycifar};  
       
       %\addplot table[x index = 0, y expr = {(92.70-\thisrow{KDCE})/92.70}] {\deepfoolresnettwentycifar};  
       
       \addplot table[x index = 0, y expr = {(93.25-\thisrow{KDKL})/93.25}] {\deepfoolresnettwentycifar};  
       
       	\addlegendentry{Vanilla};
		\addlegendentry{AMC-CH};
		%\addlegendentry{AMC-F};
		\addlegendentry{AMC-K};
		\addlegendentry{AMC-W};
		\addlegendentry{XNOR};
		\addlegendentry{ABC($1\times1$)};
		\addlegendentry{ABC($3\times3$)};
		\addlegendentry{ABC($5\times5$)};
		%\addlegendentry{ABC($7\times7$)};
		%\addlegendentry{KD-CE};
		\addlegendentry{KD-KL};
		\legend{};
		%\node[align=center, text width=2cm] (N) at (axis cs: 8,0.6) %{brittle};
		%\draw[->] (N) -- (axis cs: 6,0.78);
    \end{axis}

\end{tikzpicture} 
       }\\\vspace{-2ex}
\subfloat[FGSM - ResNet56]{
       \pgfplotstableread[col sep=space]{%
epsilon	Vanilla KDCE KDKL AMCCH AMCF AMCK AMCW XNOR ABC1 ABC2 ABC3
0	93.88 93.68 94.24	92.93   93.08   93.04   93.54   83.24	86.29	92.48	92.82
2	48.63 51.92 60.24	48.27   48.12   42.65   47.75	58.68	37.81   44.96	47.62
4	37.67 42.02 52.86	38.47   37.72   30.08   35.55	38.77	20.37   30.93	35.02
8	27.18 32.92 45.36	28.21   26.99   20.31   24.51	18.77	12.28   23.97	27.14
16	17.45 19.85 27.65	18.42   17.15   13.67   15.68	8.63 	7.75    16.74	18.42
}\fgsmresnetcifar

\begin{tikzpicture}
    \begin{axis}[
          height=0.22\textwidth,
          width=0.25\textwidth, % Scale the plot to \linewidth
          grid=major, 
          grid style={dashed,gray!30},
          xlabel= Attack Stress ($\sigma$), % Set the labels
          ylabel= Strain ($\varepsilon$),
          xmin=0,
          xmax=16,
          ymin=0,
          ymax=1,
          smooth,
          tension=0.2,
          yticklabel style = {font=\tiny, rotate=90, yshift=-0.3ex},
          xticklabel style = {font=\tiny},
		  ylabel style = {font=\tiny, yshift=-5.75ex},
		  xlabel style = {font=\tiny, yshift=2ex},
		  legend style={at={(1.04,0.28)},anchor=east, legend columns=1, fill=none ,draw = none, nodes={scale=0.3, transform shape}, column sep=2pt, mark size=1.5pt},
		  scale only axis,
		  every axis plot/.append style={line width=0.7pt, mark options={scale=1}},
          cycle list name=mycolorlist,
        ]
       \addplot table[x index = 0, y expr = {(93.88-\thisrow{Vanilla})/93.88}] {\fgsmresnetcifar};
       
       \addplot table[x index = 0, y expr = {(92.93-\thisrow{AMCCH})/92.93}] {\fgsmresnetcifar};  
       
       %\addplot table[x index = 0, y expr = {(93.08-\thisrow{AMCF})/93.08}] {\fgsmresnetcifar};  
       
       \addplot table[x index = 0, y expr = {(93.04-\thisrow{AMCK})/93.04}] {\fgsmresnetcifar};  
       
       \addplot table[x index = 0, y expr = {(93.54-\thisrow{AMCW})/93.54}] {\fgsmresnetcifar};
       
       \addplot table[x index = 0, y expr = {(83.24-\thisrow{XNOR})/83.24}] {\fgsmresnetcifar};  
       
       \addplot table[x index = 0, y expr = {(86.29-\thisrow{ABC1})/86.29}] {\fgsmresnetcifar};  
       
       \addplot table[x index = 0, y expr = {(92.48-\thisrow{ABC2})/92.48}] {\fgsmresnetcifar};  
       
       \addplot table[x index = 0, y expr = {(92.82-\thisrow{ABC3})/92.82}] {\fgsmresnetcifar};  
       %\pgfplotsset{cycle list shift=3}
       %\addplot table[x index = 0, y expr = {(93.68-\thisrow{KDCE})/93.68}] {\fgsmresnetcifar};  
       
       \addplot table[x index = 0, y expr = {(94.24-\thisrow{KDKL})/94.24}] {\fgsmresnetcifar};
       
       	\addlegendentry{Vanilla};
		\addlegendentry{AMC-CH};
		%\addlegendentry{AMC-F};
		\addlegendentry{AMC-K};
		\addlegendentry{AMC-W};
		\addlegendentry{XNOR};
		\addlegendentry{ABC($1\times1$)};
		\addlegendentry{ABC($3\times3$)};
		\addlegendentry{ABC($5\times5$)};
		%\addlegendentry{KD-CE};
		\addlegendentry{KD-KL};
		\legend{};
    \end{axis}

\end{tikzpicture} 
       }
\subfloat[PGD - ResNet56\\ - Fixed $i=3$]{
       \pgfplotstableread[col sep=space]{%
epsilon Vanilla KDCE KDKL AMCCH AMCF AMCK AMCW XNOR ABC1 ABC2 ABC3
0	93.88   93.68  94.24	92.93   93.08   93.04  93.54	83.24	86.29	92.48	92.82
0.1	80.44   89.96  90.78	77.59   78.35   78.71  80.01	78.94	78.21   85.33	85.56
0.5	33.65   72.35  74.23	32.58   32.64   29.38  32.45	62.97	39.41   43.47	43.75
1	13.19   53.40  55.81	13.88   13.08   9.28   12.10	41.65	10.31   15.35	16.35
2	3.21    25.14  35.33	4.03    3.51    1.74    2.59    13.30 	0.74    2.29	2.69
}\pgdresnetcifarfixediter

\begin{tikzpicture}
    \begin{axis}[
          height=0.22\textwidth,
          width=0.25\textwidth, % Scale the plot to \linewidth
          grid=major, 
          grid style={dashed,gray!30},
          xlabel= Attack Stress ($\sigma$), % Set the labels
          ylabel= Strain ($\varepsilon$),
          xmin=0,
          xmax=2,
          ymin=0,
          ymax=1,
          smooth,
          tension=0.2,
          yticklabel style = {font=\tiny, rotate=90, yshift=-0.3ex},
          xticklabel style = {font=\tiny},
		  ylabel style = {font=\tiny, yshift=-5.75ex},
		  xlabel style = {font=\tiny, yshift=2ex},
		  legend style={at={(0.98,0.24)},anchor=east, legend columns=1, fill=none ,draw = none, nodes={scale=0.45, transform shape}, column sep=5pt},
		  scale only axis,
		  every axis plot/.append style={line width=0.7pt, mark options={scale=1}},
          cycle list name=mycolorlist,
        ]
       \addplot table[x index = 0, y expr = {(93.88-\thisrow{Vanilla})/93.88}] {\pgdresnetcifarfixediter};
       
       \addplot table[x index = 0, y expr = {(92.93-\thisrow{AMCCH})/92.93}] {\pgdresnetcifarfixediter};  
       
       %\addplot table[x index = 0, y expr = {(93.08-\thisrow{AMCF})/93.08}] {\pgdresnetcifarfixediter};  
       
       \addplot table[x index = 0, y expr = {(93.04-\thisrow{AMCK})/93.04}] {\pgdresnetcifarfixediter};  
       
       \addplot table[x index = 0, y expr = {(93.54-\thisrow{AMCW})/93.54}] {\pgdresnetcifarfixediter};  
       
       \addplot table[x index = 0, y expr = {(83.24-\thisrow{XNOR})/83.24}] {\pgdresnetcifarfixediter};  
       
       \addplot table[x index = 0, y expr = {(86.29-\thisrow{ABC1})/86.29}] {\pgdresnetcifarfixediter};  
       
       \addplot table[x index = 0, y expr = {(92.48-\thisrow{ABC2})/92.48}] {\pgdresnetcifarfixediter};  
       
       \addplot table[x index = 0, y expr = {(92.82-\thisrow{ABC3})/92.82}] {\pgdresnetcifarfixediter}; 
        %\pgfplotsset{cycle list shift=3}
       %\addplot table[x index = 0, y expr = {(93.68-\thisrow{KDCE})/93.68}] {\pgdresnetcifarfixediter};  
       
       \addplot table[x index = 0, y expr = {(94.24-\thisrow{KDKL})/94.24}] {\pgdresnetcifarfixediter};  
       
       	\addlegendentry{Vanilla};
		\addlegendentry{AMC-CH};
		%\addlegendentry{AMC-F};
		\addlegendentry{AMC-K};
		\addlegendentry{AMC-W};
		\addlegendentry{XNOR};
		\addlegendentry{ABC($1\times1$)};
		\addlegendentry{ABC($3\times3$)};
		\addlegendentry{ABC($5\times5$)};
		%\addlegendentry{KD-CE};
		\addlegendentry{KD-KL};
		\legend{};
	\node[align=center] (N) at (axis cs: 0.5,0.9) {\footnotesize brittle};
	\draw[->] (axis cs: 0.5,0.83) -- (axis cs: 0.5,0.72);
	\node[align=center] (N) at (axis cs: 1.5,0.30) {\footnotesize ductile};
	\draw[->] (axis cs: 1.5,0.38) -- (axis cs: 1.5,0.50);
    \end{axis}

\end{tikzpicture} 
       }
\subfloat[DeepFool - ResNet56]{
       \pgfplotstableread[col sep=space]{%
iter Vanilla KDCE KDKL AMCCH AMCF AMCK AMCW XNOR ABC1 ABC2 ABC3
0	93.88   93.68  94.24	92.93   93.08   93.04  93.54	83.24	86.29	92.48	92.82
1	58.27   60.70  67.03	57.50   57.68   60.70  57.78	74.66	53.06   52.22	52.56
5	4.59    5.69   8.46     5.60    5.31    5.71    4.62	34.14	11.42   6.05	5.80
10	4.18    4.27   4.16     4.82    4.72    4.29    4.36	16.94	9.29    5.41     5.22
20	4.09    4.29   4.01     4.78    4.66    4.29    4.28	11.79 	8.60    5.34	5.10
}\deepfoolresnetcifar

\begin{tikzpicture}
    \begin{axis}[
          height=0.22\textwidth,
          width=0.25\textwidth, % Scale the plot to \linewidth
          grid=major, 
          grid style={dashed,gray!30},
          xlabel= Attack Stress ($\sigma$), % Set the labels
          ylabel= Strain ($\varepsilon$),
          xmin=0,
          xmax=20,
          ymin=0,
          ymax=1,
          %smooth,
          %enlargelimits=false,
          %tension=0.2,
          yticklabel style = {font=\tiny, rotate=90, yshift=-0.3ex},
          xticklabel style = {font=\tiny},
		  ylabel style = {font=\tiny, yshift=-5.75ex},
		  xlabel style = {font=\tiny, yshift=2ex},
		  legend style={at={(0.98,0.22)},anchor=east, legend columns=1, fill=none ,draw = none, nodes={scale=0.5, transform shape}, column sep=5pt},
		  scale only axis,
		  every axis plot/.append style={line width=0.7pt, mark options={scale=1}},
          cycle list name=mycolorlist,
          %only
        ]
       \addplot table[x index = 0, y expr = {(93.88-\thisrow{Vanilla})/93.88}] {\deepfoolresnetcifar};
       
       \addplot table[x index = 0, y expr = {(92.93-\thisrow{AMCCH})/92.93}] {\deepfoolresnetcifar};  
       
       %\addplot table[x index = 0, y expr = {(93.08-\thisrow{AMCF})/93.08}] {\deepfoolresnetcifar};  
       
       \addplot table[x index = 0, y expr = {(93.04-\thisrow{AMCK})/93.04}] {\deepfoolresnetcifar};  
       
       \addplot table[x index = 0, y expr = {(93.54-\thisrow{AMCW})/93.54}] {\deepfoolresnetcifar}; 
       
       \addplot table[x index = 0, y expr = {(83.24-\thisrow{XNOR})/83.24}] {\deepfoolresnetcifar};  
       
       \addplot table[x index = 0, y expr = {(86.29-\thisrow{ABC1})/86.29}] {\deepfoolresnetcifar};  
       
       \addplot table[x index = 0, y expr = {(92.48-\thisrow{ABC2})/92.48}] {\deepfoolresnetcifar};  
       
       \addplot table[x index = 0, y expr = {(92.82-\thisrow{ABC3})/92.82}] {\deepfoolresnetcifar}; 
       %\pgfplotsset{cycle list shift=3}
       %\addplot table[x index = 0, y expr = {(93.68-\thisrow{KDCE})/93.68}] {\deepfoolresnetcifar};  
       
       \addplot table[x index = 0, y expr = {(94.24-\thisrow{KDKL})/94.24}] {\deepfoolresnetcifar}; 
       
       	\addlegendentry{Vanilla};
		\addlegendentry{AMC-CH};
		%\addlegendentry{AMC-F};
		\addlegendentry{AMC-K};
		\addlegendentry{AMC-W};
		\addlegendentry{XNOR};
		\addlegendentry{ABC($1\times1$)};
		\addlegendentry{ABC($3\times3$)};
		\addlegendentry{ABC($5\times5$)};
        %\addlegendentry{KD-CE};
		\addlegendentry{KD-KL};
		\legend{};

    \end{axis}

\end{tikzpicture} 
       }
       \\\vspace{-2ex}
\subfloat[GenAttack - ResNet20\\ - $\epsilon=8$ $|$ $N=16$]{
       \pgfplotstableread[col sep=space]{%
gen	Vanilla KDCE KDKL AMCCH AMCF AMCK AMCW XNOR ABC1 ABC2 ABC3 ABC4
0   92.46	92.70   93.25	89.76	89.91   90.73   91.99	82.71	83.42   88.94   90.64   91.34
50	19.77   17.66   31.86   14.45   15.52   16.61   20.41   69.26    67.67   71.52   75.24   77.48
100	3.28    1.24    15.48   1.54    1.53    1.94    3.24   67.67    66.18   69.79   73.47   75.99
150	0.64    0.12    9.76    0.18    0.18    0.36    0.66   66.19    64.60   68.69   72.10   74.87
200	0.19    0.00    6.91    0.03    0.05    0.05    0.31   65.72    64.13   67.59   71.19   74.35
}\genresnettwentycifar

\begin{tikzpicture}
    \begin{axis}[
          height=0.22\textwidth,
          width=0.25\textwidth, % Scale the plot to \linewidth
          grid=major, 
          grid style={dashed,gray!30},
          xlabel= Attack Stress ($\sigma$), % Set the labels
          ylabel= Strain ($\varepsilon$),
          xmin=0,
          xmax=200,
          ymin=0,
          ymax=1,
          smooth,
          tension=0.2,
          yticklabel style = {font=\tiny, rotate=90, yshift=-0.3ex},
          xticklabel style = {font=\tiny},
		  ylabel style = {font=\tiny, yshift=-5.75ex},
		  xlabel style = {font=\tiny, yshift=2ex},
		  legend style={at={(1,0.5)},anchor=east, legend columns=1, fill=none ,draw = none, nodes={scale=0.38, transform shape}, column sep=5pt},
		  scale only axis,
		  every axis plot/.append style={line width=0.7pt, mark options={scale=1}},
          cycle list name=mycolorlist,
        ]
       \addplot table[x index = 0, y expr = {(92.46-\thisrow{Vanilla})/92.46}] {\genresnettwentycifar};
       
       \addplot table[x index = 0, y expr = {(89.76-\thisrow{AMCCH})/89.76}] {\genresnettwentycifar};  
       
       %\addplot table[x index = 0, y expr = {(89.91-\thisrow{AMCF})/89.91}] {\genresnettwentycifar};  

       \addplot table[x index = 0, y expr = {(90.73-\thisrow{AMCK})/90.73}] {\genresnettwentycifar};  
       
       \addplot table[x index = 0, y expr = {(91.99-\thisrow{AMCW})/91.99}] {\genresnettwentycifar}; 
       
       \addplot table[x index = 0, y expr = {(82.71-\thisrow{XNOR})/82.71}] {\genresnettwentycifar};  
       
       \addplot table[x index = 0, y expr = {(83.42-\thisrow{ABC1})/83.42}] {\genresnettwentycifar};  
       
       \addplot table[x index = 0, y expr = {(88.94-\thisrow{ABC2})/88.94}] {\genresnettwentycifar};  
       
       \addplot table[x index = 0, y expr = {(90.64-\thisrow{ABC3})/90.64}] {\genresnettwentycifar};  
       %\pgfplotsset{cycle list shift=3}
       %\addplot table[x index = 0, y expr = {(91.34-\thisrow{ABC4})/91.34}] {\genresnettwentycifar};  
       
       %\addplot table[x index = 0, y expr = {(92.70-\thisrow{KDCE})/92.70}] {\genresnettwentycifar};  
       
       \addplot table[x index = 0, y expr = {(93.25-\thisrow{KDKL})/93.25}] {\genresnettwentycifar};  
       
       	\addlegendentry{Vanilla};
		\addlegendentry{AMC-CH};
		%\addlegendentry{AMC-F};
		\addlegendentry{AMC-K};
		\addlegendentry{AMC-W};
		\addlegendentry{XNOR};
		\addlegendentry{ABC($1\times1$)};
		\addlegendentry{ABC($3\times3$)};
		\addlegendentry{ABC($5\times5$)};
		%\addlegendentry{ABC($7\times7$)};
		%\addlegendentry{KD-CE};
		\addlegendentry{KD-KL};
		\legend{};
	%	\node[align=center, text width=2cm] (N) at (axis cs: 150,0.45) %{strong};
	%\draw[->] (N) -- (axis cs: 150,0.3);
    \end{axis}

\end{tikzpicture} 
       }\hspace{-2ex}
\subfloat[LocalSearch- ResNet20\\ - Fixed $\epsilon=16$]{
       \pgfplotstableread[col sep=space]{%
gen	Vanilla KDCE KDKL AMCCH AMCF AMCK AMCW XNOR ABC1 ABC2 ABC3 ABC4
0   92.46	92.70   93.25	89.76	89.91   90.74    91.99	82.71	83.42   88.94   90.64   91.34
50	46.70   43.37   51.00   42.76   41.49   44.00    46.93   59.62   61.54   65.84   67.49   66.99
100	32.26   27.49   35.60   30.05   28.10   30.47    33.10   47.80   49.22   52.04   55.82   54.62
150	24.27   20.44   27.58   23.94   20.92   23.10    25.38   40.58   40.57   42.99   47.29   46.21
200	18.99   15.74   22.37   18.80   16.62   18.66    20.45   36.73   34.85   36.37   41.13   40.48
}\LSresnettwentycifar

\begin{tikzpicture}
    \begin{axis}[
          height=0.22\textwidth,
          width=0.25\textwidth, % Scale the plot to \linewidth
          grid=major, 
          grid style={dashed,gray!30},
          xlabel= Attack Stress ($\sigma$), % Set the labels
          ylabel= Strain ($\varepsilon$),
          xmin=0,
          xmax=200,
          ymin=0,
          ymax=1,
          smooth,
          tension=0.2,
          yticklabel style = {font=\tiny, rotate=90, yshift=-0.3ex},
          xticklabel style = {font=\tiny},
		  ylabel style = {font=\tiny, yshift=-5.75ex},
		  xlabel style = {font=\tiny, yshift=2ex},
		  legend style={at={(1,0.5)},anchor=east, legend columns=1, fill=none ,draw = none, nodes={scale=0.38, transform shape}, column sep=5pt},
		  scale only axis,
		  every axis plot/.append style={line width=0.7pt, mark options={scale=1}},
          cycle list name=mycolorlist,
        ]
       \addplot table[x index = 0, y expr = {(92.46-\thisrow{Vanilla})/92.46}] {\LSresnettwentycifar};
       
       \addplot table[x index = 0, y expr = {(89.76-\thisrow{AMCCH})/89.76}] {\LSresnettwentycifar};  
       
       %\addplot table[x index = 0, y expr = {(89.91-\thisrow{AMCF})/89.91}] {\LSresnettwentycifar};  
       
       \addplot table[x index = 0, y expr = {(90.74-\thisrow{AMCK})/90.74}] {\LSresnettwentycifar};  
       
       \addplot table[x index = 0, y expr = {(91.99-\thisrow{AMCW})/91.99}] {\LSresnettwentycifar};
       
       \addplot table[x index = 0, y expr = {(82.71-\thisrow{XNOR})/82.71}] {\LSresnettwentycifar};  
       
       \addplot table[x index = 0, y expr = {(83.42-\thisrow{ABC1})/83.42}] {\LSresnettwentycifar};  
       
       \addplot table[x index = 0, y expr = {(88.94-\thisrow{ABC2})/88.94}] {\LSresnettwentycifar};  
       
       \addplot table[x index = 0, y expr = {(90.64-\thisrow{ABC3})/90.64}] {\LSresnettwentycifar};  
       %\pgfplotsset{cycle list shift=3}
       %\addplot table[x index = 0, y expr = {(91.34-\thisrow{ABC4})/91.34}] {\LSresnettwentycifar};  
       
       %\addplot table[x index = 0, y expr = {(92.70-\thisrow{KDCE})/92.70}] {\LSresnettwentycifar};  
       
       \addplot table[x index = 0, y expr = {(93.25-\thisrow{KDKL})/93.25}] {\LSresnettwentycifar};  
       
       	\addlegendentry{Vanilla};
		\addlegendentry{AMC-CH};
		%\addlegendentry{AMC-F};
		\addlegendentry{AMC-K};
		\addlegendentry{AMC-W};
		\addlegendentry{XNOR};
		\addlegendentry{ABC($1\times1$)};
		\addlegendentry{ABC($3\times3$)};
		\addlegendentry{ABC($5\times5$)};
		%\addlegendentry{ABC($7\times7$)};
		%\addlegendentry{KD-CE};
		\addlegendentry{KD-KL};
		\legend{};
    \end{axis}

\end{tikzpicture} 
       }
\subfloat[CW - ResNet20\\ - Fixed $\epsilon=1$]{
       \pgfplotstableread[col sep=space]{%
gen	Vanilla KDCE KDKL AMCCH AMCF AMCK AMCW XNOR ABC1 ABC2 ABC3 ABC4
0   92.46	92.70   93.25	89.76	89.91   90.73  91.99	82.71	83.42   88.94   90.64   91.34
1	11.67   10.63   11.70   10.96   11.07   12.99  11.96   11.68   12.18   11.18   10.82   10.55
10	8.02    9.05    10.34   5.70    5.77    11.70  10.78   10.96   10.64   9.71    9.10    8.55
20	4.58    6.22    8.73    2.42    2.07    10.51  10.14   9.66    8.01    7.17    6.63    5.16
50	1.28    2.13    6.75    0.55    0.28    10.01  9.94   5.93    2.99    2.42    2.51    1.53
}\CWresnettwentycifar

\begin{tikzpicture}
    \begin{axis}[
          height=0.22\textwidth,
          width=0.25\textwidth, % Scale the plot to \linewidth
          grid=major, 
          grid style={dashed,gray!30},
          xlabel= Attack Stress ($\sigma$), % Set the labels
          ylabel= Strain ($\varepsilon$),
          xmin=0,
          xmax=50,
          ymin=0,
          ymax=1,
          smooth,
          tension=0.2,
          yticklabel style = {font=\tiny, rotate=90, yshift=-0.3ex},
          xticklabel style = {font=\tiny},
		  ylabel style = {font=\tiny, yshift=-5.75ex},
		  xlabel style = {font=\tiny, yshift=2ex},
		  legend style={at={(1,0.5)},anchor=east, legend columns=1, fill=none ,draw = none, nodes={scale=0.38, transform shape}, column sep=5pt},
		  scale only axis,
		  every axis plot/.append style={line width=0.7pt, mark options={scale=1}},
          cycle list name=mycolorlist,
        ]
       \addplot table[x index = 0, y expr = {(92.46-\thisrow{Vanilla})/92.46}] {\CWresnettwentycifar};
       
       \addplot table[x index = 0, y expr = {(89.76-\thisrow{AMCCH})/89.76}] {\CWresnettwentycifar};  
       
       %\addplot table[x index = 0, y expr = {(89.91-\thisrow{AMCF})/89.91}] {\CWresnettwentycifar};  
       
       \addplot table[x index = 0, y expr = {(90.73-\thisrow{AMCK})/90.73}] {\CWresnettwentycifar};  
       
       \addplot table[x index = 0, y expr = {(91.99-\thisrow{AMCW})/91.99}] {\CWresnettwentycifar}; 
       
       \addplot table[x index = 0, y expr = {(82.71-\thisrow{XNOR})/82.71}] {\CWresnettwentycifar};  
       
       \addplot table[x index = 0, y expr = {(83.42-\thisrow{ABC1})/83.42}] {\CWresnettwentycifar};  
       
       \addplot table[x index = 0, y expr = {(88.94-\thisrow{ABC2})/88.94}] {\CWresnettwentycifar};  
       
       \addplot table[x index = 0, y expr = {(90.64-\thisrow{ABC3})/90.64}] {\CWresnettwentycifar};  
       %\pgfplotsset{cycle list shift=3}
       %\addplot table[x index = 0, y expr = {(91.34-\thisrow{ABC4})/91.34}] {\CWresnettwentycifar};  
       
       %\addplot table[x index = 0, y expr = {(92.70-\thisrow{KDCE})/92.70}] {\CWresnettwentycifar};  
       
       \addplot table[x index = 0, y expr = {(93.25-\thisrow{KDKL})/93.25}] {\CWresnettwentycifar};  
       
       	\addlegendentry{Vanilla};
		\addlegendentry{AMC-CH};
		%\addlegendentry{AMC-F};
		\addlegendentry{AMC-K};
		\addlegendentry{AMC-W};
		\addlegendentry{XNOR};
		\addlegendentry{ABC($1\times1$)};
		\addlegendentry{ABC($3\times3$)};
		\addlegendentry{ABC($5\times5$)};
		%\addlegendentry{ABC($7\times7$)};
		%\addlegendentry{KD-CE};
		\addlegendentry{KD-KL};
		\legend{};
    \end{axis}

\end{tikzpicture} 
       }\hspace{3ex}
       \\\vspace{-2ex}
\subfloat[GenAttack - ResNet56\\ - $\epsilon=8$ $|$ $N=16$]{
       \pgfplotstableread[col sep=space]{%
gen Vanilla KDCE KDKL AMCCH AMCF AMCK AMCW XNOR ABC1 ABC2 ABC3
0	93.88   93.67   94.24   92.93   93.08   93.04   93.54   83.24	86.29	92.48	92.82
50	31.97   37.75   44.16	30.86   32.21   26.85   31.46  	66.19	69.10   77.82   79.87
100	11.12   16.04   24.07	11.09   12.88   7.72    11.03  	64.48	67.46   76.15   78.75
150	4.00    7.25    13.83	4.21    5.63    2.65    4.40    63.34	66.41   75.07	77.93
200	1.69    3.62    8.54	1.84    2.83    1.07    1.91    62.12 	65.89   74.45	77.23
}\genresnetcifar

\begin{tikzpicture}
    \begin{axis}[
          height=0.22\textwidth,
          width=0.25\textwidth, % Scale the plot to \linewidth
          grid=major, 
          grid style={dashed,gray!30},
          xlabel= Attack Stress ($\sigma$), % Set the labels
          ylabel= Strain ($\varepsilon$),
          xmin=0,
          xmax=200,
          ymin=0,
          ymax=1,
          smooth,
          tension=0.2,
          yticklabel style = {font=\tiny, rotate=90, yshift=-0.3ex},
          xticklabel style = {font=\tiny},
		  ylabel style = {font=\tiny, yshift=-5.75ex},
		  xlabel style = {font=\tiny, yshift=2ex},
		  legend style={at={(0.98,0.5)},anchor=east, legend columns=1, fill=none ,draw = none, nodes={scale=0.5, transform shape}, column sep=5pt},
		  scale only axis,
		  every axis plot/.append style={line width=0.7pt, mark options={scale=1}},
          cycle list name=mycolorlist,
        ]
       \addplot table[x index = 0, y expr = {(93.88-\thisrow{Vanilla})/93.88}] {\genresnetcifar};
       
       \addplot table[x index = 0, y expr = {(92.93-\thisrow{AMCCH})/92.93}] {\genresnetcifar};  
       
       %\addplot table[x index = 0, y expr = {(93.08-\thisrow{AMCF})/93.08}] {\genresnetcifar};  
       
       \addplot table[x index = 0, y expr = {(93.04-\thisrow{AMCK})/93.04}] {\genresnetcifar};  
       
       \addplot table[x index = 0, y expr = {(93.54-\thisrow{AMCW})/93.54}] {\genresnetcifar}; 
       
       \addplot table[x index = 0, y expr = {(83.24-\thisrow{XNOR})/83.24}] {\genresnetcifar};  
       
       \addplot table[x index = 0, y expr = {(86.29-\thisrow{ABC1})/86.29}] {\genresnetcifar};  
       
       \addplot table[x index = 0, y expr = {(92.48-\thisrow{ABC2})/92.48}] {\genresnetcifar};  
       
       \addplot table[x index = 0, y expr = {(92.82-\thisrow{ABC3})/92.82}] {\genresnetcifar};  
       %\pgfplotsset{cycle list shift=3}
       %\addplot table[x index = 0, y expr = {(93.67-\thisrow{KDCE})/93.67}] {\genresnetcifar};  
       
       \addplot table[x index = 0, y expr = {(94.24-\thisrow{KDKL})/94.24}] {\genresnetcifar}; 
       
       	\addlegendentry{Vanilla};
		\addlegendentry{AMC-CH};
		%\addlegendentry{AMC-F};
		\addlegendentry{AMC-K};
		\addlegendentry{AMC-W};
		\addlegendentry{XNOR};
		\addlegendentry{ABC($1\times1$)};
		\addlegendentry{ABC($3\times3$)};
		\addlegendentry{ABC($5\times5$)};
		%\addlegendentry{KD-CE};
		\addlegendentry{KD-KL};
		\legend{};
	\node[align=center] (N) at (axis cs: 165,0.43) {\footnotesize strong};
	\draw[->] (axis cs: 165,0.38) -- (axis cs: 165,0.25);
    \end{axis}

\end{tikzpicture} 
       }\hspace{-2.25ex}
\subfloat[LocalSearch - ResNet56\\ - Fixed $\epsilon=16$]{
       \pgfplotstableread[col sep=space]{%
gen Vanilla KDCE KDKL AMCCH AMCF AMCK AMCW XNOR ABC1 ABC2 ABC3
0	93.88   93.68  94.24   92.93   93.08    93.04  93.54   83.24	86.29	92.48	92.82
50	54.94   52.89  56.48   53.65   53.18    49.20  52.24   63.70   62.73   71.45   71.78
100	39.88   37.23  39.83   39.34   37.95    33.73  36.00   53.50   48.23   58.32   60.32
150	31.70   28.62  29.48   30.17   28.66    24.27  28.12   47.22   36.27   48.95   51.47
200	26.27   22.73  22.38   24.56   23.49    19.04  22.25   42.52   27.71   41.92   44.45
}\LSresnetcifar

\begin{tikzpicture}
    \begin{axis}[
          height=0.22\textwidth,
          width=0.25\textwidth, % Scale the plot to \linewidth
          grid=major, 
          grid style={dashed,gray!30},
          xlabel= Attack Stress ($\sigma$), % Set the labels
          ylabel= Strain ($\varepsilon$),
          xmin=0,
          xmax=200,
          ymin=0,
          ymax=1,
          smooth,
          tension=0.2,
          yticklabel style = {font=\tiny, rotate=90, yshift=-0.3ex},
          xticklabel style = {font=\tiny},
		  ylabel style = {font=\tiny, yshift=-5.75ex},
		  xlabel style = {font=\tiny, yshift=2ex},
		  legend style={at={(0.48,0.78)},anchor=east, legend columns=1, fill=none ,draw = none, nodes={scale=0.5, transform shape}, column sep=5pt},
		  scale only axis,
		  every axis plot/.append style={line width=0.7pt, mark options={scale=1}},
          cycle list name=mycolorlist,
        ]
       \addplot table[x index = 0, y expr = {(93.88-\thisrow{Vanilla})/93.88}] {\LSresnetcifar};
       
       \addplot table[x index = 0, y expr = {(92.93-\thisrow{AMCCH})/92.93}] {\LSresnetcifar}; 
       
       %\addplot table[x index = 0, y expr = {(93.08-\thisrow{AMCF})/93.08}] {\LSresnetcifar};  
       
       \addplot table[x index = 0, y expr = {(93.04-\thisrow{AMCK})/93.04}] {\LSresnetcifar}; 
       
       \addplot table[x index = 0, y expr = {(93.54-\thisrow{AMCW})/93.54}] {\LSresnetcifar};  
       
       \addplot table[x index = 0, y expr = {(83.24-\thisrow{XNOR})/83.24}] {\LSresnetcifar};  
       
       \addplot table[x index = 0, y expr = {(86.29-\thisrow{ABC1})/86.29}] {\LSresnetcifar};  
       
       \addplot table[x index = 0, y expr = {(92.48-\thisrow{ABC2})/92.48}] {\LSresnetcifar};  
       
       \addplot table[x index = 0, y expr = {(92.82-\thisrow{ABC3})/92.82}] {\LSresnetcifar};  
       %\pgfplotsset{cycle list shift=3}
       %\addplot table[x index = 0, y expr = {(93.68-\thisrow{KDCE})/93.68}] {\LSresnetcifar};  
       
       \addplot table[x index = 0, y expr = {(94.24-\thisrow{KDKL})/94.24}] {\LSresnetcifar};  
       
       	\addlegendentry{Vanilla};
		\addlegendentry{AMC-CH};
		%\addlegendentry{AMC-F};
		\addlegendentry{AMC-K};
		\addlegendentry{AMC-W};
		\addlegendentry{XNOR};
		\addlegendentry{ABC($1\times1$)};
		\addlegendentry{ABC($3\times3$)};
		\addlegendentry{ABC($5\times5$)};
		%\addlegendentry{KD-CE};
		\addlegendentry{KD-KL};
		\legend{};
    \end{axis}

\end{tikzpicture} 
       }%\hspace{-2.5ex}
\subfloat[CW - ResNet56\\ - Fixed $\epsilon=1$]{
       \pgfplotstableread[col sep=space]{%
iter Vanilla KDCE KDKL AMCCH AMCF AMCK AMCW XNOR ABC1 ABC2 ABC3
0	93.88   93.68  94.24	92.93   93.08   93.04  93.54	83.24	86.29	92.48	92.82
1	10.35   12.20  12.12	10.53   10.23   10.95  10.76	10.58	10.39   9.85	10.27
10	9.17    10.79  11.50     9.81    9.03   11.20  10.14	10.58	9.99    9.48    9.93
20	7.71    11.14  10.84     8.34    8.01   11.12  10.71	9.94	10.16   8.42	9.01
50	4.64    10.38  10.97     5.25    3.94   10.81  10.05   9.29 	8.23    5.40	6.41
}\cwresnetcifar

\begin{tikzpicture}
    \begin{axis}[
          height=0.22\textwidth,
          width=0.25\textwidth, % Scale the plot to \linewidth
          grid=major, 
          grid style={dashed,gray!30},
          xlabel= Attack Stress ($\sigma$), % Set the labels
          ylabel= Strain ($\varepsilon$),
          xmin=0,
          xmax=50,
          ymin=0,
          ymax=1,
          %smooth,
          %enlargelimits=false,
          %tension=0.2,
          yticklabel style = {font=\tiny, rotate=90, yshift=-0.3ex},
          xticklabel style = {font=\tiny},
		  ylabel style = {font=\tiny, yshift=-5.75ex},
		  xlabel style = {font=\tiny, yshift=2ex},
		  legend style={at={(0.98,0.22)},anchor=east, legend columns=1, fill=none ,draw = none, nodes={scale=0.5, transform shape}, column sep=5pt},
		  scale only axis,
		  every axis plot/.append style={line width=0.7pt, mark options={scale=1}},
          cycle list name=mycolorlist,
        ]
       \addplot table[x index = 0, y expr = {(93.88-\thisrow{Vanilla})/93.88}] {\cwresnetcifar};
       
       \addplot table[x index = 0, y expr = {(92.93-\thisrow{AMCCH})/92.93}] {\cwresnetcifar};  
       
       %\addplot table[x index = 0, y expr = {(93.08-\thisrow{AMCF})/93.08}] {\cwresnetcifar};  
       
       \addplot table[x index = 0, y expr = {(93.04-\thisrow{AMCK})/93.04}] {\cwresnetcifar};  
       
       \addplot table[x index = 0, y expr = {(93.54-\thisrow{AMCW})/93.54}] {\cwresnetcifar};  
       
       \addplot table[x index = 0, y expr = {(83.24-\thisrow{XNOR})/83.24}] {\cwresnetcifar};  
       
       \addplot table[x index = 0, y expr = {(86.29-\thisrow{ABC1})/86.29}] {\cwresnetcifar};  
       
       \addplot table[x index = 0, y expr = {(92.48-\thisrow{ABC2})/92.48}] {\cwresnetcifar};  
       
       \addplot table[x index = 0, y expr = {(92.82-\thisrow{ABC3})/92.82}] {\cwresnetcifar};   
       %\pgfplotsset{cycle list shift=3}
       %\addplot table[x index = 0, y expr = {(93.68-\thisrow{KDCE})/93.68}] {\cwresnetcifar};  
       
       \addplot table[x index = 0, y expr = {(94.24-\thisrow{KDKL})/94.24}] {\cwresnetcifar};   
       
       	\addlegendentry{Vanilla};
		\addlegendentry{AMC-CH};
		%\addlegendentry{AMC-F};
		\addlegendentry{AMC-K};
		\addlegendentry{AMC-W};
		\addlegendentry{XNOR};
		\addlegendentry{ABC($1\times1$)};
		\addlegendentry{ABC($3\times3$)};
		\addlegendentry{ABC($5\times5$)};
        %\addlegendentry{KD-CE};
		\addlegendentry{KD-KL};
		\legend{}
    \end{axis}

\end{tikzpicture} 
       }\hspace{2ex}%\vspace{-2ex}
\caption{Stress-strain graphs for various attacks on compressed variants of ResNet20 (Top) and ResNet56 (Bottom).
%\alex{REMOVE: BatchNorm~\cite{bn} layers are in $\mathtt{statistics}$ mode.}
}
\vspace{-1ex}
\label{fig:Attack_graphs}
\end{figure*}

%where $\mathcal{Y}$ is the robustness modulus and $\sigma$ is the degree of the attack's stress, i.e. amplitude or number of iterations. It is important to note that $\mathcal{Y}$ is simply a ratio for a specific stress-type/level against the sustained strain of the network. Comparisons of $\mathcal{Y}$ can \textit{only} be made under the same stress criteria.
Using $\sigma$ and $\boldsymbol{\varepsilon}$, we can compare the degree of robustness \textit{between} networks, relative to their base accuracies. We can use \textit{inverted} stress-strain graphs to better visualize the robustness of networks accordingly. Given the behavior of a network under a certain attack, we can classify its robustness in terms of material properties. A network that sustains a high attack stress before breaking is a \textit{strong} network. On the other hand, a network which gradually degrades with increased attack stress is a \textit{ductile} network. Lastly, a network which breaks before it deforms can be considered a \textit{brittle} network. Fig.~\ref{fig:Attack_graphs} shows a set of stress-strain graphs for all the networks and attacks investigated.
%\lukas{These types of characteristics can be identified/seen in Fig.~\ref{fig:Attack_graphs}}. FIg ref done in the below sentences

%\begin{figure}
%\captionsetup[subfigure]{font=scriptsize,labelfont=scriptsize}
%\begin{subfigure}{0.49\columnwidth}
%     \begin{subfigure}{\columnwidth}
%       \input{img/stress_strain/FGSM_ResNet20_stress_strain.tex}
%       \caption{FGSM - ResNet20}
%       \end{subfigure}
%       \begin{subfigure}{\columnwidth}
%       \input{img/stress_strain/Deepfool_ResNet20_stress_strain.tex}
%       \caption{DeepFool - ResNet20}
%       \end{subfigure}
%\end{subfigure}
%\begin{subfigure}{0.49\columnwidth}
%     \begin{subfigure}{\columnwidth}
%       \input{img/stress_strain/PGD_ResNet20_stress_strain_fixedamp.tex}
%       \caption{PGD - ResNet20 - Fixed $\epsilon=0.5$}
%       \end{subfigure}
%       \begin{subfigure}{\columnwidth}
%       \input{img/stress_strain/PGD_ResNet20_stress_strain_fixediter.tex}
%       \caption{PGD - ResNet20 - Fixed $i=3$}
%       \end{subfigure}
%\end{subfigure}
%\begin{subfigure}{0.49\columnwidth}
%     \begin{subfigure}{\columnwidth}
%        \input{img/stress_strain/GenAttack_ResNet20_stress_strain.tex}
%        \caption{GenAttack - ResNet20 - $\epsilon=8$ $|$ $pop=16$}
%       \end{subfigure}
%       \begin{subfigure}{\columnwidth}
%       \input{img/stress_strain/LocalSearch_ResNet20_stress_strain.tex}
%       \caption{LocalSearch - ResNet20 - Fixed $\epsilon=16$}
%       \end{subfigure}
%\end{subfigure}
%\end{figure}

\noindent
\textbf{Fast Gradient Sign Method: }
For FGSM attacks, the results show that the KD-KL variant is more resilient compared to other compression techniques, as its strain $\boldsymbol{\varepsilon}$ increases at a slower rate with intensified attack stress.
%$\sigma$\nael{not true for ResNet20}. \manoj{Can you double check the stress strain plot of res20?}\nael{Looks like data is wrong in new files (maybe deleted by mistake?) I'm recovering them}
During the training, the distillation is performed using higher temperature (T = 30). The attack perturbations are generated using cross-entropy loss with T = 1, resulting in saturated gradients and therefore weakening the attack.
%\lukas{What KPI is used to identify superior behavior? $\boldsymbol{\varepsilon}$ or $\mathcal{Y}$}. \nael{both ;)}
Fig.~\ref{fig:Attack_graphs} shows an interesting effect of increased FGSM stress on the XNOR-Net variant. The robustness of ResNet56-XNOR is higher than other variants under low stress of up to $\sigma = 4$. Beyond that point, further attack stress severely harms the robustness of the network, making it the second-worst variant, following ABC(1$\times1$). Generally, a boost in robustness is observed when the base CNN is the larger ResNet56 model. This increases their ductility, as they sustain more attack stress before breaking, when compared to the more brittle ResNet20 models. Interestingly, the same does not apply for the binarized ABC models, as they show similar robustness, irrespective of being ResNet20 or ResNet56 variants.

%\lukas{Define "Breaking" - at what point are you saying that a NN is breaking? - at a certain degradationrate (drastic change in $\boldsymbol{\varepsilon}$)?}\nael{We could not define it properly since its related to whether the attack is targeted or not. I'll try to put some general definition at the top}

\noindent
\textbf{Projected Gradient Descent: }
For PGD, increased attack stress can be interpreted as higher perturbation amplitude $\epsilon$ or more iterations $i$. 
%As such, we plot two stress-strain graphs for this attack, fixing one type of stress while varying the other. 
Fig.~\ref{fig:Attack_graphs} shows the attack stress $\sigma=\epsilon$, with iterations fixed to 3. The CNNs show various characteristics for this attack hyper-parameter setting. We observe KD-KL and XNOR variants of ResNet56 having a lower slope compared to other compressed CNNs indicating the ductile behavior. 
%However, ResNet20 variants exhibits steeper curves indicating more brittle characteristics. When fixing $\epsilon$ to 0.5, all variants can sustain multiple iterations of this amplitude, while gradually showing increased strain, i.e. accuracy degradation (Fig.~\textcolor{red}{S1} in supplementary material). The XNOR variants show more ductile behavior, having a lower slope compared to other compressed CNNs. An exception here are the ResNet56-KD variants, which are strong w.r.t. the considered attack stress range.
%\manoj{Nael can you check the changes? Replaced the plots and adapted text with varying strength}
% When fixing the iterations $i=3$ and varying the stress $\sigma = \epsilon$, slightly steeper curves are observed (figure in supplementary material). \manoj{We need to replace it with the stress strain graphs which captures $\epsilon$, to differentiate with PGD loss plots}%For either type of attack stress, the binary variants, particularly XNOR, maintain the highest robustness under PGD. Out of the non-binarized CNNs, only ResNet20-KD-KL shows comparable performance.

\noindent
\textbf{Carlini \& Wagner: }
For the C\&W method, we set the attack stress $\sigma$ to search iterations over $\epsilon=1$ (see Eq.~\ref{eq:CW}). The results show the strength of this method, rendering all our networks brittle. This is characterized by the steep ascent in strain, breaking all CNNs with minimal attack stress.

%\nael{Why is resnet56 behavior so different}

\noindent
\textbf{DeepFool: }
Similar to the C\&W attack, DeepFool renders most of the considered CNNs brittle. One exception is the ResNet56-XNOR, which can sustain some amount of stress before completely breaking. It is worth noting that the other binary CNNs do not perform as well as ResNet56-XNOR in this case.

\noindent
\textbf{LocalSearch: }
The LocalSearch attack can also offer two types of stress: amplitude and iterations. In Fig.~\ref{fig:Attack_graphs}, the stress-strain curves for a fixed amplitude of $\epsilon = 16$. For this amplitude, none of the networks completely break, even after 200 iterations of the attack. However, it is worth noting that binarized CNNs outperform the full-precision variants for both ResNet20 and ResNet56 experiments. %\nael{can you get the local search graphs to main paper?} %For $\epsilon = 16$, the attack is milder than others. None of the networks completely break, even after an attack stress of 200 iterations, maintaining a ductile behavior. Nevertheless, the binarized CNNs perform better than the full-precision variants. When fixing amplitude $\epsilon$ to $32$, the gap between BNNs and the full-precision variants increases.

\noindent
\textbf{GenAttack: }
For GenAttack, we take the number of generations $i$ as the measure of attack stress, and fix amplitude $\epsilon = 8$ and population $N = 16$. In Fig.~\ref{fig:Attack_graphs}, a clear difference between the robustness of BNNs and other variants is observed. 
%This observation holds, even when increasing the amplitude to $\epsilon = 32$ (figure in supplementary).
We can classify BNNs as strong against GenAttacks, and all other variants, as brittle.

%A general trend which can be observed for all attacks is the robustness of BNNs and the brittleness of vanilla and pruned CNNs. \nael{Improve concluding remarks across attacks}

\iffalse
\noindent
\textbf{Robustness Modulus $\mathcal{Y}$: }
Different to Young's modulus, our adapted robustness modulus $\mathcal{Y}$ does not apply to a linear region of the stress-strain graph. Instead, we use it as a ratio value at each attack-stress point. %\lukas{Add intuition about high and low values of $\mathcal{Y}$ and to what behavior of NN they lead w.r.t. robustness}.
$\mathcal{Y}$ increases with higher $\sigma$ for all networks, but a deterioration in accuracy (i.e. increased strain) results in a lower slope.
Fig~\ref{fig:ymod} presents an example of visualizing the robustness of the stress-strain plot in Fig.~\ref{fig:Attack_graphs}a. The observation made for ResNet20-XNOR under FGSM is clearly visualized, showing relatively high robustness for $\sigma = 2$ compared to other compression methods, but the lowest robustness when $\sigma = 16$.

\begin{figure}[h]
    \centering
    \input{img/stress_strain/FGSM_ResNet20_Y_Modulus.tex}
    \caption{$\mathcal{Y}$-modulus of compressed ResNet20 variants at different attack-stress levels for FGSM.}
    \label{fig:ymod}
    \vspace{-3ex}
\end{figure}
\fi
    \begin{figure*}[t]
\captionsetup[subfigure]{labelformat=empty}
\centering
    \subfloat{
       %\hspace{-7ex}
       \begin{tikzpicture}	
	\begin{axis}[
		name=legend,
		width=\textwidth,
		height=20ex,
		hide axis,
        legend style={at={(-0.01,0)},anchor=west, legend columns=-1, draw = none, nodes={scale=0.75, transform shape}, column sep=2pt},
		ymin=0,
        ymax=1,
        xmin=0,
        xmax=1,
		]
		\addlegendimage{Pal_1, mark=square,mark options={scale=1}, line width=1pt, only marks}
       	\addlegendentry{Vanilla};
       	\addlegendimage{Pal_2, mark=square,mark options={scale=1}, line width=1pt, densely dashed, only marks}
		\addlegendentry{Ch.Prune};
% 		\addlegendimage{Pal_7, mark=square,mark options={scale=1}, line width=1pt, densely dashed, only marks}
% 		\addlegendentry{F.Prune};
 		\addlegendimage{Pal_7, mark=square,mark options={scale=1}, line width=1pt, densely dashed, only marks}
        \addlegendentry{K.Prune};
        \addlegendimage{Pal_6, mark=square,mark options={scale=1}, line width=1pt, densely dashed, only marks}
		\addlegendentry{W.Prune};
		\addlegendimage{Pal_4, mark=square,mark options={scale=1}, line width=1pt, only marks}
		\addlegendentry{XNOR};
		\addlegendimage{Pal_5, mark=square,mark options={scale=1}, line width=1pt, only marks}
		\addlegendentry{ABC($1\times1$)};
		\addlegendimage{Pal_6, mark=square,mark options={scale=1}, line width=1pt, only marks}
		\addlegendentry{ABC($3\times3$)};
		\addlegendimage{Pal_8, mark=square,mark options={scale=1}, line width=1pt, only marks}
		\addlegendentry{ABC($5\times5$)};
% 		\addlegendimage{Pal_3, mark=square,mark options={scale=1}, line width=1pt, only marks}
% 		\addlegendentry{ABC($7\times7$)};
% 		\addlegendimage{Pal_1, mark=square,mark options={scale=1}, line width=1pt, only marks}
% 		\addlegendentry{KD-CE};
		\addlegendimage{Pal_2, mark=square,mark options={scale=1}, line width=1pt, only marks}
		\addlegendentry{KD-KL};
	\end{axis}
\end{tikzpicture}
       %\captionFGSM - ResNet20}
       }\\\vspace{-1ex}
       \subfloat[FGSM]{
       \pgfplotstableread[col sep=comma]{%
name,FGSMmin,FGSMq1,FGSMq2,FGSMq3,FGSMmax,PGDmin,PGDq1,PGDq2,PGDq3,PGDmax,Deepmin,Deepq1,Deepq2,Deepq3,Deepmax,CWmin,CWq1,CWq2,CWq3,CWmax,Localmin,Localq1,Localq2,Localq3,Localmax,Genmin,Genq1,Genq2,Genq3,Genmax
Vanilla,10.04,15.6675,22.315,29.825,48.63,0.09,1.99,13.555,57.635,85.71,4.09,4.4875,5.1,13.3,58.27,0.2,4.195,7.96,10.3525,22.7,0.28,10.7525,31.98,51.28,71.41,0.09,1.6375,6.825,17.1475,50.35
Ch.Prune,9.29,11.8475,20.365,30.775,48.27,0.02,1.7875,13.205,52.0925,83.87,4.78,5.405,6.7,13.3875,57.5,0.11,2.5625,7.595,10.065,21.48,0.29,8.88,30.11,48.7675,68.57,0.01,1.3925,4.89,16.2725,50.18
K.Prune,8.72,12.0725,16.99,24.9275,42.65,0.02,1.4125,9.325,54.3975,84.36,4.29,5.355,6.155,13.2675,60.7,9.96,10.21,10.81,11.375,22.35,0.55,7.0225,27.37,48.5175,65.06,0.02,1.015,4.905,13.22,47
W.Prune,8.77,14.2175,20.095,25.0525,35.55,0.04,2.01,12.39,57.42,85.54,4.28,4.555,7.74,13.9575,57.78,9.66,9.98,10.29,10.79,23.34,0.48,8.2325,30.61,51.82,67.26,0.06,1.8,7.565,17.1775,49.48
XNOR,4.85,8.5725,20.24,39.7875,58.68,0.02,12.885,39.445,70.525,79.76,10.4,11.6875,16.225,41.2525,74.66,4.33,9.87,10.38,11.305,23.21,29.32,39.4175,45.82,54.1525,64.88,55.47,61.385,64.665,67.1225,72
ABC(1x1),6.8,10.135,16.3,24.73,38.78,0.01,3.505,24.86,64.22,81.24,8.6,9.65,11.005,23.015,57.17,1.16,7.835,9.855,10.395,23.78,22.79,34.6425,40.84,51.815,64.73,55.13,60.2525,64.18,67.84,72.23
ABC(3x3),9.62,15.855,23.48,33.6925,44.96,0.03,3.955,23.735,65.29,87.66,5.34,5.89,7.56,20.255,55.98,0.96,5.9075,8.715,9.8575,21.72,23.04,39.385,47.9,59.075,73.79,57.28,64.4375,69.51,74.51,80.96
ABC(5x5),13.86,18.26,25.66,35.94,47.62,0.03,3.905,21.92,65.47,88.18,5.1,5.655,6.895,18.745,52.56,0.81,5.865,9.01,9.995,22.34,23.73,43.5825,52.07,60.4325,75.16,59.59,66.9825,71.965,77.4,82.51
KD_KL,18.69,31.3175,44.33,52.8975,60.24,1.05,24.6075,53.09,74.3375,90.88,4.01,4.4375,5.495,21.665,67.03,1.81,9.8775,10.795,11.6025,23.39,2.2,18.365,32.54,51.82,68.43,3.43,9.775,17.525,29.2275,59.05
}\boxplotdata

\begin{tikzpicture}
\begin{axis}[boxplot/draw direction=y,
width=0.35\textwidth,
height=0.3\textwidth,
grid=major, 
grid style={dashed,gray!30},
legend style={at={(1,0)},anchor=south west, legend columns=1, fill=none ,draw = none, nodes={scale=0.5, transform shape}, column sep=5pt},
ymin=0,
ymax=109,
ytick={0,20,40,60,80,100},
yticklabel style = {font=\tiny, rotate=90, yshift=-0.3ex},
xticklabel style = {font=\tiny},
ylabel= Acc. after Attack,
ylabel style = {font=\tiny, yshift=-5.75ex},
xtick=\empty,
cycle list name=mycolorlist3,
every axis plot/.append style={line width=1pt},
]
\pgfplotstablegetrowsof{\boxplotdata}
\pgfmathtruncatemacro\TotalRows{\pgfplotsretval-1}
\pgfplotsinvokeforeach{0,...,\TotalRows}
{
  \addplot+[
  boxplot prepared from table={
    table=\boxplotdata,
    row=#1,
    lower whisker=FGSMmin,
    upper whisker=FGSMmax,
    lower quartile=FGSMq1,
    upper quartile=FGSMq3,
    median=FGSMq2
  },
  boxplot prepared,
  % to get a more useful legend
  area legend
  ]
  coordinates {};

  % add legend entry 
  \pgfplotstablegetelem{#1}{name}\of\boxplotdata
  \addlegendentryexpanded{\pgfplotsretval}
}
\legend{};

\end{axis}
\end{tikzpicture}
       \vspace{-1.2ex}
       %\caption{FGSM}
       \vspace{-0.2ex}
       \label{fig:FGSM_box}
       }
       \subfloat[PGD]{
       \pgfplotstableread[col sep=comma]{%
name,FGSMmin,FGSMq1,FGSMq2,FGSMq3,FGSMmax,PGDmin,PGDq1,PGDq2,PGDq3,PGDmax,Deepmin,Deepq1,Deepq2,Deepq3,Deepmax,CWmin,CWq1,CWq2,CWq3,CWmax,Localmin,Localq1,Localq2,Localq3,Localmax,Genmin,Genq1,Genq2,Genq3,Genmax
Vanilla,10.04,15.6675,22.315,29.825,48.63,0.09,1.99,13.555,57.635,85.71,4.09,4.4875,5.1,13.3,58.27,0.2,4.195,7.96,10.3525,22.7,0.28,10.7525,31.98,51.28,71.41,0.09,1.6375,6.825,17.1475,50.35
Ch.Prune,9.29,11.8475,20.365,30.775,48.27,0.02,1.7875,13.205,52.0925,83.87,4.78,5.405,6.7,13.3875,57.5,0.11,2.5625,7.595,10.065,21.48,0.29,8.88,30.11,48.7675,68.57,0.01,1.3925,4.89,16.2725,50.18
K.Prune,8.72,12.0725,16.99,24.9275,42.65,0.02,1.4125,9.325,54.3975,84.36,4.29,5.355,6.155,13.2675,60.7,9.96,10.21,10.81,11.375,22.35,0.55,7.0225,27.37,48.5175,65.06,0.02,1.015,4.905,13.22,47
W.Prune,8.77,14.2175,20.095,25.0525,35.55,0.04,2.01,12.39,57.42,85.54,4.28,4.555,7.74,13.9575,57.78,9.66,9.98,10.29,10.79,23.34,0.48,8.2325,30.61,51.82,67.26,0.06,1.8,7.565,17.1775,49.48
XNOR,4.85,8.5725,20.24,39.7875,58.68,0.02,12.885,39.445,70.525,79.76,10.4,11.6875,16.225,41.2525,74.66,4.33,9.87,10.38,11.305,23.21,29.32,39.4175,45.82,54.1525,64.88,55.47,61.385,64.665,67.1225,72
ABC(1x1),6.8,10.135,16.3,24.73,38.78,0.01,3.505,24.86,64.22,81.24,8.6,9.65,11.005,23.015,57.17,1.16,7.835,9.855,10.395,23.78,22.79,34.6425,40.84,51.815,64.73,55.13,60.2525,64.18,67.84,72.23
ABC(3x3),9.62,15.855,23.48,33.6925,44.96,0.03,3.955,23.735,65.29,87.66,5.34,5.89,7.56,20.255,55.98,0.96,5.9075,8.715,9.8575,21.72,23.04,39.385,47.9,59.075,73.79,57.28,64.4375,69.51,74.51,80.96
ABC(5x5),13.86,18.26,25.66,35.94,47.62,0.03,3.905,21.92,65.47,88.18,5.1,5.655,6.895,18.745,52.56,0.81,5.865,9.01,9.995,22.34,23.73,43.5825,52.07,60.4325,75.16,59.59,66.9825,71.965,77.4,82.51
KD_KL,18.69,31.3175,44.33,52.8975,60.24,1.05,24.6075,53.09,74.3375,90.88,4.01,4.4375,5.495,21.665,67.03,1.81,9.8775,10.795,11.6025,23.39,2.2,18.365,32.54,51.82,68.43,3.43,9.775,17.525,29.2275,59.05
}\boxplotdata

\begin{tikzpicture}
\begin{axis}[boxplot/draw direction=y,
width=0.35\textwidth,
height=0.3\textwidth,
grid=major, 
grid style={dashed,gray!30},
legend style={at={(1,0)},anchor=south west, legend columns=1, fill=none ,draw = none, nodes={scale=0.5, transform shape}, column sep=5pt},
ymin=0,
ymax=109,
ytick={0,20,40,60,80,100},
yticklabel style = {font=\tiny, rotate=90, yshift=-0.3ex},
xticklabel style = {font=\tiny},
ylabel= Acc. after Attack,
ylabel style = {font=\tiny, yshift=-5.75ex},
xtick=\empty,
cycle list name=mycolorlist3,
every axis plot/.append style={line width=1pt},
]
\pgfplotstablegetrowsof{\boxplotdata}
\pgfmathtruncatemacro\TotalRows{\pgfplotsretval-1}
\pgfplotsinvokeforeach{0,...,\TotalRows}
{
  \addplot+[
  boxplot prepared from table={
    table=\boxplotdata,
    row=#1,
    lower whisker=PGDmin,
    upper whisker=PGDmax,
    lower quartile=PGDq1,
    upper quartile=PGDq3,
    median=PGDq2
  },
  boxplot prepared,
  % to get a more useful legend
  area legend
  ]
  coordinates {};

  % add legend entry 
  \pgfplotstablegetelem{#1}{name}\of\boxplotdata
  \addlegendentryexpanded{\pgfplotsretval}
}
\legend{};
%\node[align=center, text width=2cm] (N) at (axis cs: 6.6,105) {\tiny
%ductile};
%\draw [decorate,decoration={brace,amplitude=7pt,raise=4pt},yshift=0pt]
% (axis cs: 0.5,80) -- (axis cs: 12.6,80) node [black,midway,xshift=0.8cm];

\node[align=center] (N) at (axis cs: 5,97) {\tiny ductile};
\draw[->] (N) -- (axis cs: 5,82);

\node[align=center] (N1) at (axis cs: 3,105) {\tiny brittle};
\draw[->] (N1) -- (axis cs: 3,86);

\node[align=center] (N2) at (axis cs: 9,105) {\tiny strong};
\draw[->] (N2) -- (axis cs: 9,92);

\end{axis}
\end{tikzpicture}
       \vspace{-1.2ex}
       %\caption{PGD}
       \label{fig:PGD_box}
       }
       \subfloat[CW]{
       \pgfplotstableread[col sep=comma]{%
name,FGSMmin,FGSMq1,FGSMq2,FGSMq3,FGSMmax,PGDmin,PGDq1,PGDq2,PGDq3,PGDmax,Deepmin,Deepq1,Deepq2,Deepq3,Deepmax,CWmin,CWq1,CWq2,CWq3,CWmax,Localmin,Localq1,Localq2,Localq3,Localmax,Genmin,Genq1,Genq2,Genq3,Genmax
Vanilla,10.04,15.6675,22.315,29.825,48.63,0.09,1.99,13.555,57.635,85.71,4.09,4.4875,5.1,13.3,58.27,0.2,4.195,7.96,10.3525,22.7,0.28,10.7525,31.98,51.28,71.41,0.09,1.6375,6.825,17.1475,50.35
Ch.Prune,9.29,11.8475,20.365,30.775,48.27,0.02,1.7875,13.205,52.0925,83.87,4.78,5.405,6.7,13.3875,57.5,0.11,2.5625,7.595,10.065,21.48,0.29,8.88,30.11,48.7675,68.57,0.01,1.3925,4.89,16.2725,50.18
K.Prune,8.72,12.0725,16.99,24.9275,42.65,0.02,1.4125,9.325,54.3975,84.36,4.29,5.355,6.155,13.2675,60.7,9.96,10.21,10.81,11.375,22.35,0.55,7.0225,27.37,48.5175,65.06,0.02,1.015,4.905,13.22,47
W.Prune,8.77,14.2175,20.095,25.0525,35.55,0.04,2.01,12.39,57.42,85.54,4.28,4.555,7.74,13.9575,57.78,9.66,9.98,10.29,10.79,23.34,0.48,8.2325,30.61,51.82,67.26,0.06,1.8,7.565,17.1775,49.48
XNOR,4.85,8.5725,20.24,39.7875,58.68,0.02,12.885,39.445,70.525,79.76,10.4,11.6875,16.225,41.2525,74.66,4.33,9.87,10.38,11.305,23.21,29.32,39.4175,45.82,54.1525,64.88,55.47,61.385,64.665,67.1225,72
ABC(1x1),6.8,10.135,16.3,24.73,38.78,0.01,3.505,24.86,64.22,81.24,8.6,9.65,11.005,23.015,57.17,1.16,7.835,9.855,10.395,23.78,22.79,34.6425,40.84,51.815,64.73,55.13,60.2525,64.18,67.84,72.23
ABC(3x3),9.62,15.855,23.48,33.6925,44.96,0.03,3.955,23.735,65.29,87.66,5.34,5.89,7.56,20.255,55.98,0.96,5.9075,8.715,9.8575,21.72,23.04,39.385,47.9,59.075,73.79,57.28,64.4375,69.51,74.51,80.96
ABC(5x5),13.86,18.26,25.66,35.94,47.62,0.03,3.905,21.92,65.47,88.18,5.1,5.655,6.895,18.745,52.56,0.81,5.865,9.01,9.995,22.34,23.73,43.5825,52.07,60.4325,75.16,59.59,66.9825,71.965,77.4,82.51
KD_KL,18.69,31.3175,44.33,52.8975,60.24,1.05,24.6075,53.09,74.3375,90.88,4.01,4.4375,5.495,21.665,67.03,1.81,9.8775,10.795,11.6025,23.39,2.2,18.365,32.54,51.82,68.43,3.43,9.775,17.525,29.2275,59.05

}\boxplotdata

\begin{tikzpicture}
\begin{axis}[boxplot/draw direction=y,
width=0.35\textwidth,
height=0.3\textwidth,
grid=major, 
grid style={dashed,gray!30},
legend style={at={(1,0)},anchor=south west, legend columns=1, fill=none ,draw = none, nodes={scale=0.5, transform shape}, column sep=5pt},
ymin=0,
ymax=109,
ytick={0,20,40,60,80,100},
yticklabel style = {font=\tiny, rotate=90, yshift=-0.3ex},
xticklabel style = {font=\tiny},
ylabel= Acc. after Attack,
ylabel style = {font=\tiny, yshift=-5.75ex},
xtick=\empty,
cycle list name=mycolorlist3,
every axis plot/.append style={line width=1pt},
]
\pgfplotstablegetrowsof{\boxplotdata}
\pgfmathtruncatemacro\TotalRows{\pgfplotsretval-1}
\pgfplotsinvokeforeach{0,...,\TotalRows}
{
  \addplot+[
  boxplot prepared from table={
    table=\boxplotdata,
    row=#1,
    lower whisker=CWmin,
    upper whisker=CWmax,
    lower quartile=CWq1,
    upper quartile=CWq3,
    median=CWq2
  },
  boxplot prepared,
  % to get a more useful legend
  area legend
  ]
  coordinates {};

  % add legend entry 
  \pgfplotstablegetelem{#1}{name}\of\boxplotdata
  \addlegendentryexpanded{\pgfplotsretval}
}
\legend{};
%\node[align=center, text width=2cm] (N) at (axis cs: 6.5,62) {\tiny
%brittle};
%\draw [decorate,decoration={brace,amplitude=7pt,raise=4pt},yshift=0pt]
% (axis cs: 0.5,35) -- (axis cs: 12.5,35) node [black,midway,xshift=0.8cm];
\end{axis}
\end{tikzpicture}
       \vspace{-1.2ex}
       %\caption{CW}
       \vspace{-0.2ex}
       \label{fig:CW_box}
       }\\\vspace{-2ex}
       \subfloat[LocalSearch]{
       \pgfplotstableread[col sep=comma]{%
name,FGSMmin,FGSMq1,FGSMq2,FGSMq3,FGSMmax,PGDmin,PGDq1,PGDq2,PGDq3,PGDmax,Deepmin,Deepq1,Deepq2,Deepq3,Deepmax,CWmin,CWq1,CWq2,CWq3,CWmax,Localmin,Localq1,Localq2,Localq3,Localmax,Genmin,Genq1,Genq2,Genq3,Genmax
Vanilla,10.04,15.6675,22.315,29.825,48.63,0.09,1.99,13.555,57.635,85.71,4.09,4.4875,5.1,13.3,58.27,0.2,4.195,7.96,10.3525,22.7,0.28,10.7525,31.98,51.28,71.41,0.09,1.6375,6.825,17.1475,50.35
Ch.Prune,9.29,11.8475,20.365,30.775,48.27,0.02,1.7875,13.205,52.0925,83.87,4.78,5.405,6.7,13.3875,57.5,0.11,2.5625,7.595,10.065,21.48,0.29,8.88,30.11,48.7675,68.57,0.01,1.3925,4.89,16.2725,50.18
K.Prune,8.72,12.0725,16.99,24.9275,42.65,0.02,1.4125,9.325,54.3975,84.36,4.29,5.355,6.155,13.2675,60.7,9.96,10.21,10.81,11.375,22.35,0.55,7.0225,27.37,48.5175,65.06,0.02,1.015,4.905,13.22,47
W.Prune,8.77,14.2175,20.095,25.0525,35.55,0.04,2.01,12.39,57.42,85.54,4.28,4.555,7.74,13.9575,57.78,9.66,9.98,10.29,10.79,23.34,0.48,8.2325,30.61,51.82,67.26,0.06,1.8,7.565,17.1775,49.48
XNOR,4.85,8.5725,20.24,39.7875,58.68,0.02,12.885,39.445,70.525,79.76,10.4,11.6875,16.225,41.2525,74.66,4.33,9.87,10.38,11.305,23.21,29.32,39.4175,45.82,54.1525,64.88,55.47,61.385,64.665,67.1225,72
ABC(1x1),6.8,10.135,16.3,24.73,38.78,0.01,3.505,24.86,64.22,81.24,8.6,9.65,11.005,23.015,57.17,1.16,7.835,9.855,10.395,23.78,22.79,34.6425,40.84,51.815,64.73,55.13,60.2525,64.18,67.84,72.23
ABC(3x3),9.62,15.855,23.48,33.6925,44.96,0.03,3.955,23.735,65.29,87.66,5.34,5.89,7.56,20.255,55.98,0.96,5.9075,8.715,9.8575,21.72,23.04,39.385,47.9,59.075,73.79,57.28,64.4375,69.51,74.51,80.96
ABC(5x5),13.86,18.26,25.66,35.94,47.62,0.03,3.905,21.92,65.47,88.18,5.1,5.655,6.895,18.745,52.56,0.81,5.865,9.01,9.995,22.34,23.73,43.5825,52.07,60.4325,75.16,59.59,66.9825,71.965,77.4,82.51
KD_KL,18.69,31.3175,44.33,52.8975,60.24,1.05,24.6075,53.09,74.3375,90.88,4.01,4.4375,5.495,21.665,67.03,1.81,9.8775,10.795,11.6025,23.39,2.2,18.365,32.54,51.82,68.43,3.43,9.775,17.525,29.2275,59.05
}\boxplotdata

\begin{tikzpicture}
\begin{axis}[boxplot/draw direction=y,
width=0.35\textwidth,
height=0.3\textwidth,
grid=major, 
grid style={dashed,gray!30},
legend style={at={(1,0)},anchor=south west, legend columns=1, fill=none ,draw = none, nodes={scale=0.5, transform shape}, column sep=5pt},
ymin=0,
ymax=109,
ytick={0,20,40,60,80,100},
yticklabel style = {font=\tiny, rotate=90, yshift=-0.3ex},
xticklabel style = {font=\tiny},
ylabel= Acc. after Attack,
ylabel style = {font=\tiny, yshift=-5.75ex},
xtick=\empty,
cycle list name=mycolorlist3,
every axis plot/.append style={line width=1pt},
]
\pgfplotstablegetrowsof{\boxplotdata}
\pgfmathtruncatemacro\TotalRows{\pgfplotsretval-1}
\pgfplotsinvokeforeach{0,...,\TotalRows}
{
  \addplot+[
  boxplot prepared from table={
    table=\boxplotdata,
    row=#1,
    lower whisker=Localmin,
    upper whisker=Localmax,
    lower quartile=Localq1,
    upper quartile=Localq3,
    median=Localq2
  },
  boxplot prepared,
  % to get a more useful legend
  area legend
  ]
  coordinates {};

  % add legend entry 
  \pgfplotstablegetelem{#1}{name}\of\boxplotdata
  \addlegendentryexpanded{\pgfplotsretval}
}
\legend{};
%\node[align=center, text width=2cm] (N) at (axis cs: 8,93) {\tiny
%strong};
%\draw [decorate,decoration={brace,amplitude=5pt,raise=4pt},yshift=0pt]
% (axis cs: 5.52,70) -- (axis cs: 10.5,70) node [black,midway,xshift=0.8cm];
%\node[align=center, text width=2cm] (N) at (axis cs: 3,93) {\tiny
%ductile};
%\draw [decorate,decoration={brace,amplitude=5pt,raise=4pt},yshift=0pt]
% (axis cs: 0.5,70) -- (axis cs: 5.48,70) node [black,midway,xshift=0.8cm];
\end{axis}
\end{tikzpicture}
       \vspace{-1.2ex}
       %\caption{LocalSearch}
       \label{fig:ls_box}
       }
       \subfloat[DeepFool]{
       \pgfplotstableread[col sep=comma]{%
name,FGSMmin,FGSMq1,FGSMq2,FGSMq3,FGSMmax,PGDmin,PGDq1,PGDq2,PGDq3,PGDmax,Deepmin,Deepq1,Deepq2,Deepq3,Deepmax,CWmin,CWq1,CWq2,CWq3,CWmax,Localmin,Localq1,Localq2,Localq3,Localmax,Genmin,Genq1,Genq2,Genq3,Genmax
Vanilla,10.04,15.6675,22.315,29.825,48.63,0.09,1.99,13.555,57.635,85.71,4.09,4.4875,5.1,13.3,58.27,0.2,4.195,7.96,10.3525,22.7,0.28,10.7525,31.98,51.28,71.41,0.09,1.6375,6.825,17.1475,50.35
Ch.Prune,9.29,11.8475,20.365,30.775,48.27,0.02,1.7875,13.205,52.0925,83.87,4.78,5.405,6.7,13.3875,57.5,0.11,2.5625,7.595,10.065,21.48,0.29,8.88,30.11,48.7675,68.57,0.01,1.3925,4.89,16.2725,50.18
K.Prune,8.72,12.0725,16.99,24.9275,42.65,0.02,1.4125,9.325,54.3975,84.36,4.29,5.355,6.155,13.2675,60.7,9.96,10.21,10.81,11.375,22.35,0.55,7.0225,27.37,48.5175,65.06,0.02,1.015,4.905,13.22,47
W.Prune,8.77,14.2175,20.095,25.0525,35.55,0.04,2.01,12.39,57.42,85.54,4.28,4.555,7.74,13.9575,57.78,9.66,9.98,10.29,10.79,23.34,0.48,8.2325,30.61,51.82,67.26,0.06,1.8,7.565,17.1775,49.48
XNOR,4.85,8.5725,20.24,39.7875,58.68,0.02,12.885,39.445,70.525,79.76,10.4,11.6875,16.225,41.2525,74.66,4.33,9.87,10.38,11.305,23.21,29.32,39.4175,45.82,54.1525,64.88,55.47,61.385,64.665,67.1225,72
ABC(1x1),6.8,10.135,16.3,24.73,38.78,0.01,3.505,24.86,64.22,81.24,8.6,9.65,11.005,23.015,57.17,1.16,7.835,9.855,10.395,23.78,22.79,34.6425,40.84,51.815,64.73,55.13,60.2525,64.18,67.84,72.23
ABC(3x3),9.62,15.855,23.48,33.6925,44.96,0.03,3.955,23.735,65.29,87.66,5.34,5.89,7.56,20.255,55.98,0.96,5.9075,8.715,9.8575,21.72,23.04,39.385,47.9,59.075,73.79,57.28,64.4375,69.51,74.51,80.96
ABC(5x5),13.86,18.26,25.66,35.94,47.62,0.03,3.905,21.92,65.47,88.18,5.1,5.655,6.895,18.745,52.56,0.81,5.865,9.01,9.995,22.34,23.73,43.5825,52.07,60.4325,75.16,59.59,66.9825,71.965,77.4,82.51
KD_KL,18.69,31.3175,44.33,52.8975,60.24,1.05,24.6075,53.09,74.3375,90.88,4.01,4.4375,5.495,21.665,67.03,1.81,9.8775,10.795,11.6025,23.39,2.2,18.365,32.54,51.82,68.43,3.43,9.775,17.525,29.2275,59.05
}\boxplotdata

\begin{tikzpicture}
\begin{axis}[boxplot/draw direction=y,
width=0.35\textwidth,
height=0.3\textwidth,
grid=major, 
grid style={dashed,gray!30},
legend style={at={(1,0)},anchor=south west, legend columns=1, fill=none ,draw = none, nodes={scale=0.5, transform shape}, column sep=5pt},
ymin=0,
ymax=109,
ytick={0,20,40,60,80,100},
yticklabel style = {font=\tiny, rotate=90, yshift=-0.3ex},
xticklabel style = {font=\tiny},
ylabel= Acc. after Attack,
ylabel style = {font=\tiny, yshift=-5.75ex},
xtick=\empty,
cycle list name=mycolorlist3,
every axis plot/.append style={line width=1pt},
]
\pgfplotstablegetrowsof{\boxplotdata}
\pgfmathtruncatemacro\TotalRows{\pgfplotsretval-1}
\pgfplotsinvokeforeach{0,...,\TotalRows}
{
  \addplot+[
  boxplot prepared from table={
    table=\boxplotdata,
    row=#1,
    lower whisker=Deepmin,
    upper whisker=Deepmax,
    lower quartile=Deepq1,
    upper quartile=Deepq3,
    median=Deepq2
  },
  boxplot prepared,
  % to get a more useful legend
  area legend
  ]
  coordinates {};

  % add legend entry 
  \pgfplotstablegetelem{#1}{name}\of\boxplotdata
  \addlegendentryexpanded{\pgfplotsretval}
}
\legend{};

\end{axis}
\end{tikzpicture}
       \vspace{-1.2ex}
       %\caption{DeepFool}
       \vspace{-0.2ex}
       \label{fig:df_boxplot}
       }
       \subfloat[GenAttack]{
       \pgfplotstableread[col sep=comma]{%
name,FGSMmin,FGSMq1,FGSMq2,FGSMq3,FGSMmax,PGDmin,PGDq1,PGDq2,PGDq3,PGDmax,Deepmin,Deepq1,Deepq2,Deepq3,Deepmax,CWmin,CWq1,CWq2,CWq3,CWmax,Localmin,Localq1,Localq2,Localq3,Localmax,Genmin,Genq1,Genq2,Genq3,Genmax
Vanilla,10.04,15.6675,22.315,29.825,48.63,0.09,1.99,13.555,57.635,85.71,4.09,4.4875,5.1,13.3,58.27,0.2,4.195,7.96,10.3525,22.7,0.28,10.7525,31.98,51.28,71.41,0.09,1.6375,6.825,17.1475,50.35
Ch.Prune,9.29,11.8475,20.365,30.775,48.27,0.02,1.7875,13.205,52.0925,83.87,4.78,5.405,6.7,13.3875,57.5,0.11,2.5625,7.595,10.065,21.48,0.29,8.88,30.11,48.7675,68.57,0.01,1.3925,4.89,16.2725,50.18
K.Prune,8.72,12.0725,16.99,24.9275,42.65,0.02,1.4125,9.325,54.3975,84.36,4.29,5.355,6.155,13.2675,60.7,9.96,10.21,10.81,11.375,22.35,0.55,7.0225,27.37,48.5175,65.06,0.02,1.015,4.905,13.22,47
W.Prune,8.77,14.2175,20.095,25.0525,35.55,0.04,2.01,12.39,57.42,85.54,4.28,4.555,7.74,13.9575,57.78,9.66,9.98,10.29,10.79,23.34,0.48,8.2325,30.61,51.82,67.26,0.06,1.8,7.565,17.1775,49.48
XNOR,4.85,8.5725,20.24,39.7875,58.68,0.02,12.885,39.445,70.525,79.76,10.4,11.6875,16.225,41.2525,74.66,4.33,9.87,10.38,11.305,23.21,29.32,39.4175,45.82,54.1525,64.88,55.47,61.385,64.665,67.1225,72
ABC(1x1),6.8,10.135,16.3,24.73,38.78,0.01,3.505,24.86,64.22,81.24,8.6,9.65,11.005,23.015,57.17,1.16,7.835,9.855,10.395,23.78,22.79,34.6425,40.84,51.815,64.73,55.13,60.2525,64.18,67.84,72.23
ABC(3x3),9.62,15.855,23.48,33.6925,44.96,0.03,3.955,23.735,65.29,87.66,5.34,5.89,7.56,20.255,55.98,0.96,5.9075,8.715,9.8575,21.72,23.04,39.385,47.9,59.075,73.79,57.28,64.4375,69.51,74.51,80.96
ABC(5x5),13.86,18.26,25.66,35.94,47.62,0.03,3.905,21.92,65.47,88.18,5.1,5.655,6.895,18.745,52.56,0.81,5.865,9.01,9.995,22.34,23.73,43.5825,52.07,60.4325,75.16,59.59,66.9825,71.965,77.4,82.51
KD_KL,18.69,31.3175,44.33,52.8975,60.24,1.05,24.6075,53.09,74.3375,90.88,4.01,4.4375,5.495,21.665,67.03,1.81,9.8775,10.795,11.6025,23.39,2.2,18.365,32.54,51.82,68.43,3.43,9.775,17.525,29.2275,59.05
}\boxplotdata

\begin{tikzpicture}
\begin{axis}[boxplot/draw direction=y,
width=0.35\textwidth,
height=0.3\textwidth,
grid=major, 
grid style={dashed,gray!30},
legend style={at={(1,0)},anchor=south west, legend columns=1, fill=none ,draw = none, nodes={scale=0.5, transform shape}, column sep=5pt},
ymin=0,
ymax=109,
ytick={0,20,40,60,80,100},
yticklabel style = {font=\tiny, rotate=90, yshift=-0.3ex},
xticklabel style = {font=\tiny},
ylabel= Acc. after Attack,
ylabel style = {font=\tiny, yshift=-5.75ex},
xtick=\empty,
cycle list name=mycolorlist3,
every axis plot/.append style={line width=1pt},
]
\pgfplotstablegetrowsof{\boxplotdata}
\pgfmathtruncatemacro\TotalRows{\pgfplotsretval-1}
\pgfplotsinvokeforeach{0,...,\TotalRows}
{
  \addplot+[
  boxplot prepared from table={
    table=\boxplotdata,
    row=#1,
    lower whisker=Genmin,
    upper whisker=Genmax,
    lower quartile=Genq1,
    upper quartile=Genq3,
    median=Genq2
  },
  boxplot prepared,
  % to get a more useful legend
  area legend
  ]
  coordinates {};

  % add legend entry 
  \pgfplotstablegetelem{#1}{name}\of\boxplotdata
  \addlegendentryexpanded{\pgfplotsretval}
}
\legend{};

%\node[align=center, text width=2cm] (N) at (axis cs: 8,35) {\tiny
%strong};
%\draw [decorate,decoration={brace,amplitude=5pt,raise=4pt, mirror},yshift=0pt]
% (axis cs: 5.5,58) -- (axis cs: 10.5,58) node [black,midway,xshift=0.8cm];

\end{axis}
\end{tikzpicture}
       \vspace{-1.2ex}
       %\caption{GenAttack}
       \label{fig:gen_attack}
       }
\vspace{-2ex}
\caption{Box-plots for attacks on compressed variants of ResNet20 and ResNet56.}
%\alex{PGD KD-KL is not ductile -> should be strong ... also XNOR is kind of strong} 
%\nael{Ductile refers to a large distribution of accuracies before reaching zero/random guess. Strong is when they dont reach zero at all. At least that was my interpretation}\alex{1. Fig 2 shows something different -> messege would not aling}\alex{2. every model breaks under enough preasure (permutations)}}\alex{Yes they are more soft than in GenAttack but still different than the other ones of pgd}
\vspace{-2ex}
\label{fig:boxplot_cluster}
\end{figure*}
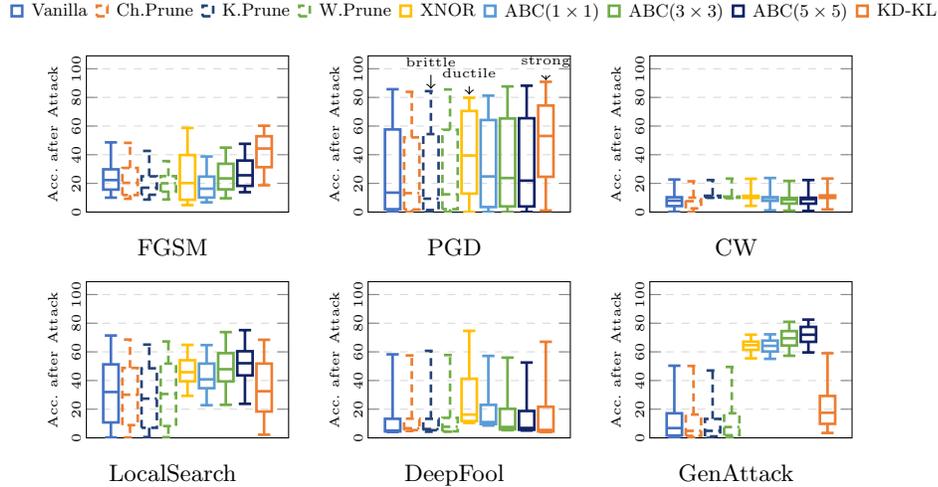

% \alex{we dont want to say, that individual attacks are better than others -> we compare an ensemble of different AA applied to the compressed CNNs. We show that the compressed CNNs (BED) behave differently on the AA} \\
% \manoj{In the case of White box attacks, the behaviour of different compressed models looks the same. This observation is somewhat similar to stress strain graphs}\\
% \alex{what does it mean when the curve of a compressed model is bend more strong than another one}\\
% \manoj{This pattern is seen in the blackbox attacks. I think its only because of local search and genattack produce different accuracy for same norm AAS.}\\
% \alex{does a rectangular surface is more robust as a banana shape one? observation: banana shape more strong than recangular? see also whitebox}
% \manoj{rectangle means : same Norm AAS, different Attack Acc}\\
% \manoj{banana mean : As Norm AAS increases, the attack acc decreases}\\
% \manoj{ABC nets 3x3 and 5x5 have more observations towards greater Attack Acc and this represents their strong behaviour mentioned in stress strain graphs.}
\noindent
\textbf{Box-Plots:} In Fig.~\ref{fig:boxplot_cluster}, we present box-plots from data collected over a range of experiments. For each attack, we sweep over the respective strength and iterations mentioned in Tab.~\ref{tab:minmaxsweep}. The exact definition of strength and iteration for each attack can be recalled from Sec.~\ref{sec:adversarial_attacks}.
The data includes both models, ResNet20 and ResNet56. %Each attack is swept across its stress parameters, as shown in Tab.~\ref{tab:minmaxsweep}.

\begin{table*}[h]
  \centering
  \resizebox{0.6\textwidth}{!}{
  \renewcommand{\arraystretch}{1.0}
  \begin{tabular}{l|c|c} 
    \toprule
    \textbf{Attack} &\textbf{Strength} $\epsilon$ & \textbf{Iterations} $i$ \\
    \midrule
    \midrule
    FGSM  & 2, 4, 8, 16 & N/A\\
    \midrule
    PGD   & 0.1, 0.5, 1.0, 2.0 & 2, 3, 4, 5\\
    \midrule
    CW    & 0.01, 0.1, 1.0, 5.0, 10.0 & 1,10, 20, 50\\
    \midrule
    DeepFool & N/A & 1, 5, 10, 20\\
    \midrule
    Local Search & 8, 16, 32 & 50, 100, 150, 200\\
    \midrule
    \multirow{2}{*}{GenAttack} & \multirow{2}{*}{8, 12} & 50,100, 150, 200\\
    & & $pop size$={6, 16}\\
    \bottomrule                                
    \end{tabular}}
  \vspace{1em}
  \caption{All strength and iteration combinations tested for ResNet20 and ResNet56 variants (vanilla, pruned, binary, and distilled). Strength and iteration definitions for each attack are explained in Sec.~\ref{sec:adversarial_attacks}.}
  \label{tab:minmaxsweep}
\end{table*}

Each plot shows the distribution of all the accuracies achieved by the compression technique, after being attacked by the corresponding method, over all the considered strengths and iterations, as well as their combinations. %over two models (ResNet20 and 56), for all explored attack variations in Tab.~\ref{tab:minmaxsweep} .
% For FGSM, we measure the attacked accuracy for by sweeping the strength from $\epsilon=2$ to $\epsilon=16$ in Eq.~\ref{eq:FGSM}. For PGD based attacked accuracy, we sweep the attack strength from  $\epsilon=0.1$ to $\epsilon=2.0$ and iterations $i=1$ to $i=5$ in Eq.~\ref{eq:PGD}. 
%\nael{We should mention what ranges of attack are included in each boxplot since supplementary material is gone.}. \manoj{yeah}
% effect of the attacks on different model variants of ResNet20/56. 
%The distribution of accuracy after attack is based on the experiments in the supplementary material from Tab.~\textcolor{red}{S2} to Tab.~\textcolor{red}{S17}.
%\manoj{Adapt from supp after changes}
The box-plots reveal the strength of BNNs against both black-box attacks (GenAttack and LocalSearch), when compared to other variants. Different compression techniques produce different distributions for the PGD attack (marked in Fig.~\ref{fig:boxplot_cluster}). CW proves to be the strongest adversarial attack scheme across all the compressed variants. \\% Based on the total distribution, we classify the models' robustness into brittle, ductile and strong w.r.t. other variants for the same attack hyper-parameters (Fig~\ref{fig:PGD_box}). Vanilla and pruned variants are generally less robust w.r.t. other variants for all attacks.\\

    \subsection{Class Activation Mapping on Attacked CNNs}
    %CAM experimental setup - RoI
We use class activation maps (CAM)~\cite{cam} to determine the region of interest (RoI) for the prediction class using clean and attacked images. The output feature maps of the last convolutional layer and the weight tensor of the fully-connected layer is considered as the input to the CAM. The CAM highlights regions of the image that influence the CNN's prediction to a specific class. Similar to heat-maps, \textcolor{red}{red} regions indicate those with the highest contribution, while \textcolor{blue}{blue} indicates the ones with the least. We applied CAM on various compressed variants of ResNet20 and ResNet56, trained on CIFAR-10, which are attacked by DeepFool (Tab.~\ref{tab:cam_resnet56}). As mentioned in Sec.~\ref{sec:adversarial_attacks}, DeepFool attempts to find the adversarial perturbation which leads the CNN to the closest decision boundary. Once a perturbation is found, it is reinforced to push the prediction beyond that boundary. Through the CAM visualizations in this section, we attempt to capture this behaviour over the attack iterations.

\begin{table*}[ht]
\centering
\setlength{\tabcolsep}{2pt}
\resizebox{0.95\textwidth}{!}{
\begin{tabular}{lcc|c|c|ccc|cccc}
\toprule
\multirow{2}{*}{Model}&\multicolumn{2}{c|}{\multirow{2}{*}{\textbf{Image$\rightarrow$ \fcolorbox{red}{white}{$I^{Adv}$}}}}&\multicolumn{1}{c|}{\multirow{2}{*}{\textbf{Vanilla}}}&\multicolumn{1}{c|}{\textbf{Distilled}}&\multicolumn{3}{c|}{\textbf{Pruned}}&\multicolumn{4}{c}{\textbf{Binary}}\\
%\multirow{2}{*}{\textbf{Model}}&No AA & \multicolumn{2}{c}{DeepFool\cite{Deepfool}}&No AA&\multicolumn{2}{c}{DeepFool\cite{Deepfool}} \\
%&$L$=$Car$&$i$=$1$&$i$=$5$&$L$=$Cat$&$i$=$1$&$i$=$5$\\
%\textbf{DeepFool}
&&&&
%\textbf{KD\_CE}&
\footnotesize{KD-KL}&
\footnotesize{Ch./F.}&
% Filter&
\footnotesize{Kernel}&
\footnotesize{Weight}&
\footnotesize{XNOR}&
\footnotesize{ABC(1$\times$1)}&
\footnotesize{ABC(3$\times$3)}&
\footnotesize{ABC(5$\times$5)}\\
% &&(93.88 \%)&
% (94.24 \%)&
% (92.86 \%)&
% (92.86\%)&
% (93.09\%)&
% (93.04\%)&
% (93.54\%)&
% (83.24 \%)&
% (86.29 \%)&
% (92.48 \%)\\
\midrule
\midrule
\multirow{3}{*}{\rotatebox{90}{\makecell{ResNet20 - CIFAR10}}}&
\adjustbox{valign=t}{\rotatebox{90}{\makecell{No AA}}}&
\adjustbox{valign=t}{\includegraphics[width=0.15\columnwidth]{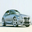}}&
\adjustbox{valign=t}{\includegraphics[width=0.15\columnwidth]{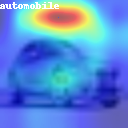}}&
%\adjustbox{valign=t}{\includegraphics[width=0.15\columnwidth]{img/CAM_ResNet56/auto/ResNet56_KD_CE_overlay_1604.png}}&
\adjustbox{valign=t}{\includegraphics[width=0.15\columnwidth]{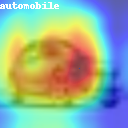}}&
\adjustbox{valign=t}{\includegraphics[width=0.15\columnwidth]{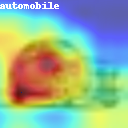}}&
% \adjustbox{valign=t}{\includegraphics[width=0.15\columnwidth]{img/CAM_ResNet20/auto/ResNet20_AMC_filter_prune_overlay_1604.png}}&
\adjustbox{valign=t}{\includegraphics[width=0.15\columnwidth]{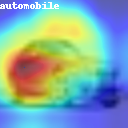}}&
\adjustbox{valign=t}{\includegraphics[width=0.15\columnwidth]{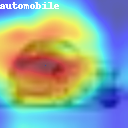}}&
\adjustbox{valign=t}{\includegraphics[width=0.15\columnwidth]{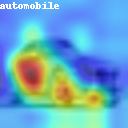}}&
\adjustbox{valign=t}{\includegraphics[width=0.15\columnwidth]{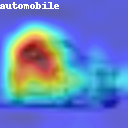}}&
\adjustbox{valign=t}{\includegraphics[width=0.15\columnwidth]{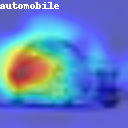}}&
\adjustbox{valign=t}{\includegraphics[width=0.15\columnwidth]{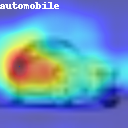}}\\
% \adjustbox{valign=t}{\includegraphics[width=0.15\columnwidth]{img/CAM_ResNet56/auto/ResNet56_ABC7x7_overlay_1604.png}}\\
&\adjustbox{valign=t}{\rotatebox{90}{\makecell{$i=1$}}}&
\fcolorbox{red}{red}{\adjustbox{valign=t}{\includegraphics[width=0.15\columnwidth]{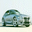}}}&
\adjustbox{valign=t}{\includegraphics[width=0.15\columnwidth]{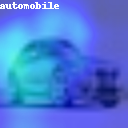}}&
%\adjustbox{valign=t}{\includegraphics[width=0.15\columnwidth]{img/CAM_ResNet56/aa_auto_i1/ResNet56_KD_CE_overlay_1604.png}}&
\adjustbox{valign=t}{\includegraphics[width=0.15\columnwidth]{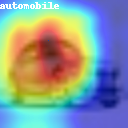}}&
\fcolorbox{red}{red}{\adjustbox{valign=t}{\includegraphics[width=0.15\columnwidth]{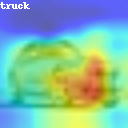}}}&
% \adjustbox{valign=t}{\includegraphics[width=0.15\columnwidth]{img/CAM_ResNet20/aa_auto_i1/ResNet20_AMC_filter_prune_overlay_1604.png}}&
\adjustbox{valign=t}{\includegraphics[width=0.15\columnwidth]{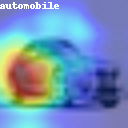}}&
\adjustbox{valign=t}{\includegraphics[width=0.15\columnwidth]{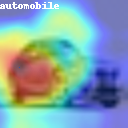}}&
\adjustbox{valign=t}{\includegraphics[width=0.15\columnwidth]{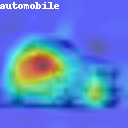}}&
\adjustbox{valign=t}{\includegraphics[width=0.15\columnwidth]{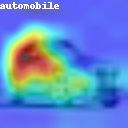}}&
\adjustbox{valign=t}{\includegraphics[width=0.15\columnwidth]{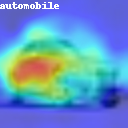}}&
\fcolorbox{red}{red}{\adjustbox{valign=t}{\includegraphics[width=0.15\columnwidth]{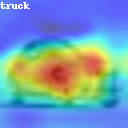}}}\\
%\adjustbox{valign=t}{\includegraphics[width=0.15\columnwidth]{img/CAM_ResNet56/aa_auto_i1/ResNet56_ABC7x7_overlay_1604.png}}\\
&\adjustbox{valign=t}{\rotatebox{90}{\makecell{$i=5$}}}&
\fcolorbox{red}{red}{\adjustbox{valign=t}{\includegraphics[width=0.15\columnwidth]{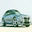}}}&
\fcolorbox{red}{red}{\adjustbox{valign=t}{\includegraphics[width=0.15\columnwidth]{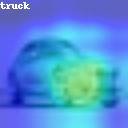}}}&
%\fcolorbox{red}{red}{\adjustbox{valign=t}{\includegraphics[width=0.15\columnwidth]{img/CAM_ResNet56/aa_auto_i5/ResNet56_KD_CE_overlay_1604.png}}}&
\fcolorbox{red}{red}{\adjustbox{valign=t}{\includegraphics[width=0.15\columnwidth]{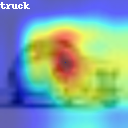}}}&
\fcolorbox{red}{red}{\adjustbox{valign=t}{\includegraphics[width=0.15\columnwidth]{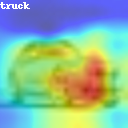}}}&
% \fcolorbox{red}{red}{\adjustbox{valign=t}{\includegraphics[width=0.15\columnwidth]{img/CAM_ResNet20/aa_auto_i5/ResNet20_AMC_filter_prune_overlay_1604.png}}}&
\fcolorbox{red}{red}{\adjustbox{valign=t}{\includegraphics[width=0.15\columnwidth]{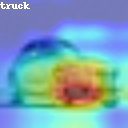}}}&
\fcolorbox{red}{red}{\adjustbox{valign=t}{\includegraphics[width=0.15\columnwidth]{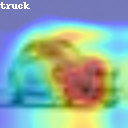}}}&
\adjustbox{valign=t}{\includegraphics[width=0.15\columnwidth]{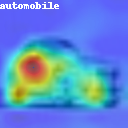}}&
\fcolorbox{red}{red}{\adjustbox{valign=t}{\includegraphics[width=0.15\columnwidth]{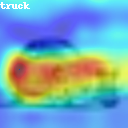}}}&
\fcolorbox{red}{red}{\adjustbox{valign=t}{\includegraphics[width=0.15\columnwidth]{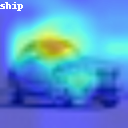}}}&
\fcolorbox{red}{red}{\adjustbox{valign=t}{\includegraphics[width=0.15\columnwidth]{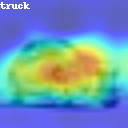}}}\\
\midrule
\multirow{3}{*}{\rotatebox{90}{\makecell{ResNet56 - CIFAR10}}}&
%\multirow{1}{*}{\rotatebox[origin=c]{90}{\makecell[c]{No AA}}}&
\adjustbox{valign=t}{\rotatebox{90}{\makecell{No AA}}}&
\adjustbox{valign=t}{\includegraphics[width=0.15\columnwidth]{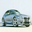}}&
\adjustbox{valign=t}{\includegraphics[width=0.15\columnwidth]{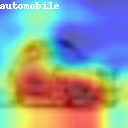}}&
%\adjustbox{valign=t}{\includegraphics[width=0.15\columnwidth]{img/CAM_ResNet56/auto/ResNet56_KD_CE_overlay_1604.png}}&
\adjustbox{valign=t}{\includegraphics[width=0.15\columnwidth]{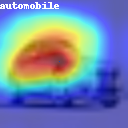}}&
\adjustbox{valign=t}{\includegraphics[width=0.15\columnwidth]{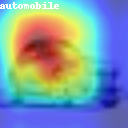}}&
% \adjustbox{valign=t}{\includegraphics[width=0.15\columnwidth]{img/CAM_ResNet56/auto/ResNet56_AMC_filter_prune_overlay_1604.png}}&
\adjustbox{valign=t}{\includegraphics[width=0.15\columnwidth]{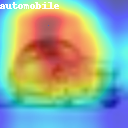}}&
\adjustbox{valign=t}{\includegraphics[width=0.15\columnwidth]{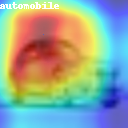}}&
\adjustbox{valign=t}{\includegraphics[width=0.15\columnwidth]{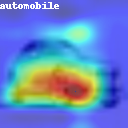}}&
\adjustbox{valign=t}{\includegraphics[width=0.15\columnwidth]{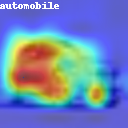}}&
\adjustbox{valign=t}{\includegraphics[width=0.15\columnwidth]{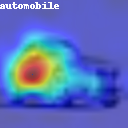}}&
\adjustbox{valign=t}{\includegraphics[width=0.15\columnwidth]{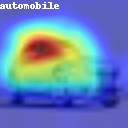}}\\
%\adjustbox{valign=t}{\includegraphics[width=0.15\columnwidth]{img/CAM_ResNet56/auto/ResNet56_ABC7x7_overlay_1604.png}}\\
&\adjustbox{valign=t}{\rotatebox{90}{\makecell{$i=1$}}}&
\fcolorbox{red}{red}{\adjustbox{valign=t}{\includegraphics[width=0.15\columnwidth]{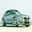}}}&
\adjustbox{valign=t}{\includegraphics[width=0.15\columnwidth]{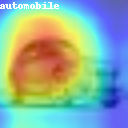}}&
%\adjustbox{valign=t}{\includegraphics[width=0.15\columnwidth]{img/CAM_ResNet56/aa_auto_i1/ResNet56_KD_CE_overlay_1604.png}}&
\adjustbox{valign=t}{\includegraphics[width=0.15\columnwidth]{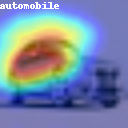}}&
\adjustbox{valign=t}{\includegraphics[width=0.15\columnwidth]{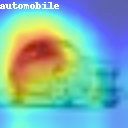}}&
% \adjustbox{valign=t}{\includegraphics[width=0.15\columnwidth]{img/CAM_ResNet56/aa_auto_i1/ResNet56_AMC_filter_prune_overlay_1604.png}}&
\adjustbox{valign=t}{\includegraphics[width=0.15\columnwidth]{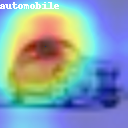}}&
\adjustbox{valign=t}{\includegraphics[width=0.15\columnwidth]{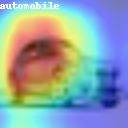}}&
\adjustbox{valign=t}{\includegraphics[width=0.15\columnwidth]{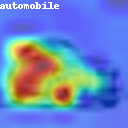}}&
\fcolorbox{red}{red}{\adjustbox{valign=t}{\includegraphics[width=0.15\columnwidth]{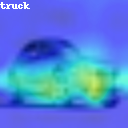}}}&
\adjustbox{valign=t}{\includegraphics[width=0.15\columnwidth]{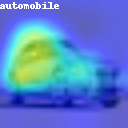}}&
\fcolorbox{red}{red}{\adjustbox{valign=t}{\includegraphics[width=0.15\columnwidth]{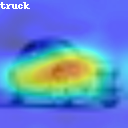}}}\\
%\adjustbox{valign=t}{\includegraphics[width=0.15\columnwidth]{img/CAM_ResNet56/aa_auto_i1/ResNet56_ABC7x7_overlay_1604.png}}\\
&\adjustbox{valign=t}{\rotatebox{90}{\makecell{$i=5$}}}&
\fcolorbox{red}{red}{\adjustbox{valign=t}{\includegraphics[width=0.15\columnwidth]{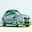}}}&
\fcolorbox{red}{red}{\adjustbox{valign=t}{\includegraphics[width=0.15\columnwidth]{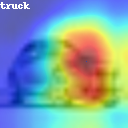}}}&
%\fcolorbox{red}{red}{\adjustbox{valign=t}{\includegraphics[width=0.15\columnwidth]{img/CAM_ResNet56/aa_auto_i5/ResNet56_KD_CE_overlay_1604.png}}}&
\fcolorbox{red}{red}{\adjustbox{valign=t}{\includegraphics[width=0.15\columnwidth]{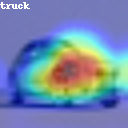}}}&
\fcolorbox{red}{red}{\adjustbox{valign=t}{\includegraphics[width=0.15\columnwidth]{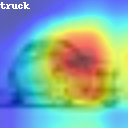}}}&
% \fcolorbox{red}{red}{\adjustbox{valign=t}{\includegraphics[width=0.15\columnwidth]{img/CAM_ResNet56/aa_auto_i5/ResNet56_AMC_filter_prune_overlay_1604.png}}}&
\fcolorbox{red}{red}{\adjustbox{valign=t}{\includegraphics[width=0.15\columnwidth]{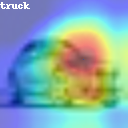}}}&
\fcolorbox{red}{red}{\adjustbox{valign=t}{\includegraphics[width=0.15\columnwidth]{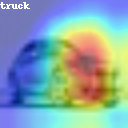}}}&
\adjustbox{valign=t}{\includegraphics[width=0.15\columnwidth]{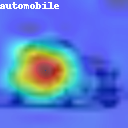}}&
\fcolorbox{red}{red}{\adjustbox{valign=t}{\includegraphics[width=0.15\columnwidth]{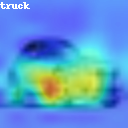}}}&
\fcolorbox{red}{red}{\adjustbox{valign=t}{\includegraphics[width=0.15\columnwidth]{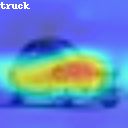}}}&
\fcolorbox{red}{red}{\adjustbox{valign=t}{\includegraphics[width=0.15\columnwidth]{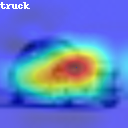}}}\\
%\fcolorbox{red}{red}{\adjustbox{valign=t}{\includegraphics[width=0.15\columnwidth]{img/CAM_ResNet56/aa_auto_i5/ResNet56_ABC7x7_overlay_1604.png}}}\\
\bottomrule
\end{tabular}}
\vspace{2ex}
\caption{CAM for ResNet20/56 and its compressed variants performed on non-attacked and DeepFool attacked images on the automobile image from CIFAR-10 dataset.
% \manoj{Can we consider removing the CAT based CAM images, so that we have no confusion in interpretation?}
}
\label{tab:cam_resnet56}
\end{table*}

All the compression techniques produce no mis-classification in the automobile example using the unattacked raw image in Tab.~\ref{tab:cam_resnet56}. 
% as the RoI is located at the center (passenger area/cabin). 
Three interpretations can be made from the heat maps. %, when the input image is attacked. 
We support our interpretation with quantitative analysis by measuring the third quartile value of the heat map intensity across all the pixels. 
%For an intensity range of (0,255) blue->red
%Our interpretations can also be supported using the quantitative analysis as suggested. For an intensity range of (0,255) blue->red, the vanilla and channel pruned models for the car example have the third quartile (Q3) of pixels with an intensity of 114, 96 respectively. When the attack is applied for one iteration, the Q3 value increases (150, 156) indicating the larger ROI. With more iterations, the Q3 value decreases (135, 121).  The pixels above the Q3 indicate reinforced red regions (bonnet) to fool to the nearest class (truck).
%First interpretation and supporting quantitative analysis
Observing the CAM output of ResNet56's vanilla and channel-pruned variants for the unattacked input image, the RoI has large focused interest regions. For an intensity range of (0,255) \textcolor{blue}{blue}$\rightarrow$\textcolor{red}{red}, the third quartile value of the heat map intensity across all pixels is 184 and 162 for vanilla and channel-pruned respectively, indicating a large RoI.
%Second and Third interpretation and supporting quantitative analysis
Second, the intensity of the interest regions decreases, after the attack is applied for one iteration. The third quartile value decreases (171, 152) indicating the lower interest regions. Third, after the attack is applied for five iterations, the focus on the attacked region (bonet) is reinforced to fool towards the nearest class ($truck$). The third quartile value further decreases (135, 121). Under DeepFool attacks, ResNet56 is more robust compared to ResNet20 which can be illustrated by the more distinct RoIs in the heat maps. 
The BNN variants have a small RoI compared to their vanilla model for unattacked images. The third quartile value for ResNet56-XNOR is 98 indicating this aspect. As the inherent RoI for BNNs are small and concentrated, it could reduce the chances of finding and perturbing the smaller set of critical pixels by the attack model.

    \subsection{Robustness Evaluation on ImageNet dataset}
    For the robustness evaluation on the ImageNet dataset~\cite{ILSVRC15}, we use pre-trained ResNet50 and ResNet18 models, and compressed variants of ResNet18. 
%For the robustness analysis, we use FGSM, PGD and GenSearch attack methods. 
We observe a higher attack search time for ImageNet compared to the CIFAR-10 dataset due to the larger image sizes and model complexity. Therefore, we limit our analysis to two white-box attacks (FGSM and PGD), and one black-box attack (GenAttack). We consider compressed variants such as Ch-Prune, XNOR, ABC(1$\times$1) and ABC(3$\times$3) specified in Tab.~\ref{tab:FGSMImageNet}-~\ref{tab:GenAttackImageNet_eval}. 

\noindent
\textbf{Fast Gradient Sign Method:} In Tab.~\ref{tab:FGSMImageNet}, we report the natural accuracy and attacked accuracy for different strengths ($\epsilon=\{2, 4, 8, 16\}$). ResNet50 achieves the highest natural accuracy and attacked accuracy for different strengths compared to other models. Among the compressed variants the channel pruned and ABC(3x3) models portray slightly higher robustness at different strengths. 
\begin{table*}[t]
  \centering
  \resizebox{0.6\textwidth}{!}{
  \renewcommand{\arraystretch}{1.0}
  \begin{tabular}{l|l|cccc} 
    \toprule
    \textbf{FGSM} &\textbf{Nat.Acc} & $\epsilon=2$ & $\epsilon=4$ & $\epsilon=8$ & $\epsilon=16$\\
    \midrule
    \midrule
    \textbf{ResNet50}~\cite{He2015DeepRL}    
    %&$\mathtt{mini}$-$\mathtt{batch}$  &23.02&15.87&11.27&8.23\\
                                  
    &75.43 \%             &22.18&16.24&12.08&7.46\\    
    %\midrule
    \textbf{ResNet18}~\cite{He2015DeepRL}    
    %&$\mathtt{mini}$-$\mathtt{batch}$  &11.99  &6.82   &4.49   &2.74\\
    &69.00 \%             &12.82  &8.16   &5.19   &2.95\\
    %\midrule
    \textbf{ResNet18-Ch.Prune}~\cite{AMC2019}    
    %&$\mathtt{mini}$-$\mathtt{batch}$  &14.95  &9.46   &5.52   &3.43\\ 
                                 
    &67.62 \%             &11.18  &6.64   &3.99   &2.34\\
    %\midrule
    \textbf{ResNet18-XNOR}~\cite{rastegari2016}       %&$\mathtt{mini}$-$\mathtt{batch}$  &11.54  &6.43   &3.19   &1.33\\
                                   
    &49.10 \%         &7.57   &4.54   &2.19   &0.93\\
    %\midrule
    \textbf{ResNet18-ABC(1$\times$1)}~\cite{lin2017}  %&$\mathtt{mini}$-$\mathtt{batch}$  &10.98&5.00&2.1 7&1.02\\
    &    51.07 \%                &9.11&4.65&2.30&1.13\\
    \textbf{ResNet18-ABC(3$\times$3)}~\cite{lin2017}  %&$\mathtt{mini}$-$\mathtt{batch}$  &12.12&5.61&2.56&1.36\\
    &59.83 \%             &11.33&5.73&2.65&1.43\\
    \bottomrule                                
    \end{tabular}}
  \vspace{1em}
  \caption{Accuracy (Top1) [\%] of CNNs after FGSM adversarial attacks for ImageNet.}
  \label{tab:FGSMImageNet}
\end{table*}

%STATISTICS MODE:
\begin{table*}[t]
    \centering
    \resizebox{0.6\textwidth}{!}{
    \renewcommand{\arraystretch}{0.6}
    \begin{tabular}{p{5cm}|c|cccc} 
    \toprule
    \textbf{PGD} &$\epsilon$    & $i=2$ &$i=3$ &$i=4$ &$i=5$ \\
    \midrule    
    \midrule
    \textbf{ResNet50}~\cite{He2015DeepRL}    &0.1     &25.77&16.07 & 9.83 & 5.91 \\  
    (75.43 \%)                      &0.5     &3.35&0.94&0.43& 0.27\\
    \midrule
    \textbf{ResNet18}~\cite{He2015DeepRL}    &0.1     &17.86 &10.32  &5.58   &3.11\\
    (69.00 \%)                      &0.5     &1.33  &0.17   &0.04   &0.01\\
    \midrule
    \textbf{ResNet18-Ch.Prune}~\cite{AMC2019}    &0.1     &17.02 &10.23  &5.92   &3.50\\
    (67.62 \%)                      &0.5     &1.40  &0.27   &0.06   &0.02\\
    \midrule
    \textbf{ResNet18-XNOR}~\cite{rastegari2016}       &0.1     &13.16 &11.46  &10.06  &8.84\\
    (49.10 \%)                      &0.5     &5.67  &3.07   &1.57   &0.78\\
    \midrule                                       
    \textbf{ResNet18-ABC(1$\times$1)}~\cite{lin2017}  &0.1     &18.35&16.22&14.20&12.37\\               
    (51.91)                         &0.5     &7.60&3.64&1.75&0.82\\
    \midrule                                       
    \textbf{ResNet18-ABC(3$\times$3)}~\cite{lin2017}  &0.1     &23.90&20.81&17.80&15.07\\                
    (59.83)                         &0.5     &8.31&3.70&1.59&0.66\\
    \bottomrule                                
    \end{tabular}}
    \vspace{1em}
    \caption{Accuracy [\%] of CNNs after PGD adversarial attacks for ImageNet.}
    
    \label{tab:PGDImageNet_eval}
\end{table*}
\noindent
\textbf{Projected Gradient Decent:} In Tab.~\ref{tab:PGDImageNet_eval}, we report the attacked accuracy for two strengths ($\epsilon=0.1$, $\epsilon=0.5$). The attacked accuracy decreases for all the models as we increase the number of iterations $i$. We observe 9.16\% higher attacked accuracy for binarized ResNet18 using ABC(3$\times$3) compared to the ResNet50 model at $i=5$ and $\epsilon=0.1$. Robustness at higher attack strength $\epsilon=0.5$ degrades the prediction accuracy for all the compressed variants. 
\begin{table*}[t]
\centering
\resizebox{0.9\textwidth}{!}{
\renewcommand{\arraystretch}{1.0}
\begin{tabular}{l|c|ccccc}
\toprule
\multirow{2}{*}{\textbf{GenAttack}}&\multirow{2}{*}{\textbf{$\epsilon$ }}             & $i=200$          & $i=400$         & $i=600$        & $i=800$       & $i=1000$     \\ 
                  &                                 & \textbf{OA/TA}   & \textbf{OA/TA}  & \textbf{OA/TA} & \textbf{OA/TA}& \textbf{OA/TA}\\ 
\midrule
\midrule
\textbf{ResNet50}\cite{He2015DeepRL}        &  8.0  & 21.29/12.80   & 11.64/34.46   &  6.87/51.94   &  4.67/64.08   &  3.06/72.82   \\         
(75.43 \%)                                          & 12.0  &  13.16/17.45  & 5.67/41.19    &  3.55/56.65   &  2.40/67.29   &  1.60/74.58      \\
\midrule                                            
\textbf{ResNet18}\cite{He2015DeepRL}        & 8.0   & 16.41/14.52   & 8.11/41.83    &  4.35/62.58   &  2.36/75.62   &  1.34/83.29 \\         
(69.00 \%)                                          & 12.0  & 10.24/22.44   & 5.13/50.74    &  2.70/68.85   &  1.58/80.21   &  1.04/86.62 \\
\midrule                                            
\textbf{ResNet18-Ch.Prune}\cite{AMC2019}  & 8.0   & 12.34/12.82   & 6.05/39.02    &  3.17/60.46   &  2.00/74.46   &  1.22/82.79 \\         
(67.62 \%)                                          & 12.0  & 7.33/20.25    & 3.29/49.44    &  1.84/68.97   &  1.08/80.11   &  0.88/86.80 \\
\midrule
\textbf{ResNet18-XNOR}\cite{rastegari2016}           & 8.0   & 13.06/0.64    & 12.86/0.72    & 12.64/0.84       & 12.68/0.86        &  12.68/0.94   \\         
(49.10 \%)                                          & 12.0  & 11.56/0.78         & 11.14/0.92         &  11.14/1.04        &  11.04/1.16        &  10.82/1.22  \\
\midrule                                       
\textbf{ResNet18-ABC(1$\times$1)}\cite{lin2017}&  8.0  & 17.59/1.48    & 17.67/1.62    & 17.37/1.76    &  17.23/1.88   &  16.89/1.98  \\         
(51.07 \%)                                          & 12.0  & 15.83/1.90    & 15.40/2.08    &  15.20/2.26   &  15.02/2.34   & 14.86/2.52  \\
\midrule
\textbf{ResNet18-ABC(3$\times$3)}\cite{lin2017}&  8.0  & 26.00/0.68    & 25.02/0.82    & 25.26/0.92    &  25.46/0.98   &  25.58/0.96  \\         
(59.83 \%)                                             & 12.0  & 22.50/0.74    & 22.04/0.94    &  22.36/1.02   &  21.75/1.08   & 21.90/1.14  \\
\bottomrule
\multicolumn{7}{r}{OA/TA = Accuracy to original label / Accuracy to target label.}\\ 
\end{tabular}}
\vspace{2ex}
\caption{Accuracy (Top1) [\%] of CNNs after GenAttack adversarial attacks for ImageNet. Pop size = 6.}
\label{tab:GenAttackImageNet_eval}
\end{table*}

%\textbf{Local Search:} In Tab.~\ref{tab:LocalSearchImageNet_eval}, we report natural accuracy in column 2. We report overall attacked accuracy and accuracy w.r.t the fooled target class at several iterations during the search ($i={50, 100, 150, 200}$). We observe that the  compressed variants loose hrobustness 
%\input{data/experiments/45_LocalSearch/453_Experiments_LocalSearch_ImageNet_eval.tex}

\noindent
\textbf{GenAttack:} We set an adaptive mutation rate $\rho$ and mutation range $\alpha$ for GenAttack based on the dataset configuration and set the population size to 6, as in~\cite{GenAttack}. In Tab.~\ref{tab:GenAttackImageNet_eval}, we report overall attacked accuracy and accuracy w.r.t. the fooled target class at several iterations during the attack search ($i$ = $\{200, 400, \allowbreak 600, 800, 1000\}$). We also analyze the robustness for two attack strengths ($\epsilon=8, 12$). Similar to previous observations, ABC models portray higher robustness with respect to their unattacked accuracy, when compared to other compressed variants and the vanilla ResNet50 and ResNet18 models. 
\subsection{Discussion}
    The robustness of distilled models can be attributed to their soft label training, which can be more informative than sheer, hard labels. The student is ideally able to learn both the correct classification \textit{and} the distribution of closeness among other classes. Furthermore, the student is distilled using a high temperature factor $T$, causing the magnitude of the predicted class to be $T$ times more confident than when trained on hard labels~\cite{CW}. Thus, white box attacks like FGSM, PGD and DeepFool would require strong adversarial perturbation for fooling the final prediction to its nearest class. However, the C\&W attack is able to fool the distilled model, even at higher temperatures as the attack is not focused on the cross-entropy loss directly.

The training scheme for BNNs is not as simple as vanilla or pruned models. It requires a straight-through-estimator, making the white-box attacks challenging compared to other variants. Introducing multiple scaling factors in case of ABC-Net eases the approximation to its full-precision model. Thus, XNOR-Nets appear to be more resilient against white-box attacks (Fig.~\ref{fig:Attack_graphs}, Fig.~\ref{fig:boxplot_cluster}). Moreover, the PGD loss levels in Fig.~\ref{fig:pgd_eval} demonstrate the robustness of XNOR-Net through lower loss convergence values and breaking speed.  The discretization of weights and activations also makes BNNs stronger against black-box attacks. The CAM results support the robustness for BNNs as they inherently possess smaller and concentrated RoI, reducing the chances of finding and perturbing the critical set of pixels. The BNN robustness is also observed for the ImageNet dataset when attacked with PGD and GenAttack (Tab.~\ref{tab:PGDImageNet_eval}, Tab.~\ref{tab:GenAttackImageNet_eval}).
%\alex{Kl i aggree}\alex{xnor the discretization of weights and activations brings the benifit, i am not sure if the ste for backpropagation is of importance, however, the scaling factors a make the model less robust, this becomes even more important when scale factors for the activations are introduced, see ABC }\manoj{discretization of weights could be a good justification for the robustness of BNNs against black box attacks. We can add in the box plot section as we tell bnns are strong with black box. I was trying to differentiate the gradient flow as CAM is generated using deepfool }\manoj{Due to scale factors, the gradient flow might get easier in case of ABC nets }\manoj{This could also be the reason why XNOR is good at white box (simple, no scale factors, gradient flow not easy) and ABC are better in black box(discretization)}

Pruning is the process of eliminating unused and/or redundant parameters. Here, balancing the compression rate and the accuracy is a key factor. 
%Pruned models inherently have a larger RoI, as seen in the CAM results, which directly increases their susceptibility to attacks.
Due to the reduced learning ability, pruned models are not automatically more robust than their full-precision counterpart. This would call for an extra objective function for improving the robustness. Existing works have shown that it is possible to remove more model parameters when pruning is applied in an unstructured manner~\cite{Han2015}. A similar behavior can be expected if the robustness is included in the pruning and fine-tuning process.
% \manoj{Include suggestions on which compression tech to use in which scenarios}
% \manoj{Include strong findings on PGD robustness}

\section{Conclusion}
    In this paper, we provided a comprehensive analysis on recent white-box and black-box adversarial attacks against state-of-the-art vanilla, distilled, pruned and binary neural networks.
We demonstrated that the robustness of CNNs not only depends on the adversarial attack but also on the compression technique at hand. By varying the attacks' hyper-parameters, strong, ductile and brittle CNNs were identified. Conclusions were made on robustness by analyzing PGD loss/accuracy levels, box-plots, stress-strain graphs and CNN heat maps with CAM. 
%The following three conclusions can be made from the analysis: First, knowledge distillation, i.e. by minimizing the KL divergence between a teacher and a student, inherently make the model more robust to various adversarial attacks. Second, on the tested black-box attacks, BNNs are more robust compared to other compressed neural networks. Finally, binary and efficient DNNs break differently on various adversarial attacks.
From the presented data, we show that knowledge about the expected adversarial attack or the used compression technique can help the designer or the attacker generate more robust applications or stronger attacks respectively.
    
{\small
\bibliographystyle{splncs03}
\bibliography{references}
}

\end{document}

\maketitle
%  ___      __   ___  __           ___      ___  __  
% |__  \_/ |__) |__  |__) |  |\/| |__  |\ |  |  /__` 
% |___ / \ |    |___ |  \ |  |  | |___ | \|  |  .__/ 

\section{Class Activation Mapping on Attacked CNNs}
    We apply class activation mapping (CAM) to compare the internal representations of different CNNs in a visual manner~\cite{cam}. We consider the output feature maps of the last convolutional layer and the weight tensor of the fully-connected layer as the input to the CAM. Results for compressed variants of ResNet20 can be found in Tab.~\ref{tab:cam_resnet20}. Tab.~\ref{tab:cam_imagenet} shows the results of $i$=1 and $i$=2 iterations of PGD attacks on ResNet18, ResNet50 and two compressed variants, namely a channel pruned and a binarized (XNOR). The models are trained on the ImageNet dataset.
    \begin{table}[ht]
\centering
\setlength{\tabcolsep}{2pt}
\resizebox{\textwidth}{!}{
\begin{tabular}{lc|c|c|cccc|cccc}
\toprule
&&\multicolumn{1}{c|}{\textbf{Vanilla}}&\multicolumn{1}{c|}{\textbf{Distilled}}&\multicolumn{4}{c|}{\textbf{Pruned}}&\multicolumn{4}{c}{\textbf{Binary}}\\
%\multirow{2}{*}{\textbf{Model}}&No AA & \multicolumn{2}{c}{DeepFool\cite{Deepfool}}&No AA&\multicolumn{2}{c}{DeepFool\cite{Deepfool}} \\
%&$L$=$Car$&$i$=$1$&$i$=$5$&$L$=$Cat$&$i$=$1$&$i$=$5$\\
%\textbf{DeepFool}
\multicolumn{2}{c|}{\textbf{Image$\rightarrow$ \fcolorbox{red}{white}{$I^{Adv}$}}}&
ResNet20&
%\textbf{KD\_CE}&
KD-KL&
Channel&
Filter&
Kernel&
Weight&
XNOR&
ABC(1$\times$1)&
ABC(3$\times$3)&
ABC(5$\times$5)\\
&&(92.46 \%)&
(92.70 \%)&
%(93.25 \%)&
(89.76 \%)&
(89.91 \%)&
(90.73 \%)&
(91.98 \%)&
(82.71 \%)&
(83.42 \%)&
(88.94 \%)&
(90.64 \%)\\
% \adjustbox{valign=t}{\makecell[l]{\textbf{Image}\\$I \rightarrow$ \fcolorbox{red}{white}{$I^{Adv}$}}} &
% \adjustbox{valign=t}{\makecell[l]{\textbf{ResNet20}~\cite{He2015DeepRL}\\(92.46 \%)}}&
% \adjustbox{valign=t}{\makecell[l]{\textbf{KD\_CE}~\cite{contrastive_kd}\\(92.70 \%)}}&
% \adjustbox{valign=t}{\makecell[l]{\textbf{KD\_KL}~\cite{contrastive_kd}\\(93.25 \%)}}&
% \adjustbox{valign=t}{\makecell[l]{\textbf{AMC(Ch.Prune)}~\cite{AMC2019}\\(89.76 \%)}}&
% \adjustbox{valign=t}{\makecell[l]{\textbf{XNOR}~\cite{rastegari2016}\\(82.71 \%)}}&
% \adjustbox{valign=t}{\makecell[l]{\textbf{ABC(1$\times$1)}~\cite{lin2017}\\(83.42 \%)}}&
% \adjustbox{valign=t}{\makecell[l]{\textbf{ABC(3$\times$3)}~\cite{lin2017}\\(88.94 \%)}}&
% \adjustbox{valign=t}{\makecell[l]{\textbf{ABC(5$\times$5)}~\cite{lin2017}\\(90.64 \%)}}&
% \adjustbox{valign=t}{\makecell[l]{\textbf{ABC(7$\times$7)}~\cite{lin2017}\\(91.34 \%)}}\\
\midrule
\midrule
\adjustbox{valign=t}{\rotatebox[origin=c]{90}{No AA}}&
\adjustbox{valign=t}{\includegraphics[width=0.15\columnwidth]{img/CAM_ResNet20/auto/ResNet20_raw_image_1604.png}}&
\adjustbox{valign=t}{\includegraphics[width=0.15\columnwidth]{img/CAM_ResNet20/auto/ResNet20_overlay_1604.png}}&
%\adjustbox{valign=t}{\includegraphics[width=0.15\columnwidth]{img/CAM_ResNet20/auto/ResNet20_KD_CE_overlay_1604.png}}&
\adjustbox{valign=t}{\includegraphics[width=0.15\columnwidth]{img/CAM_ResNet20/auto/ResNet20_KD_KL_overlay_1604.png}}&
\adjustbox{valign=t}{\includegraphics[width=0.15\columnwidth]{img/CAM_ResNet20/auto/ResNet20_AMC_channel_prune_overlay_1604.png}}&
\adjustbox{valign=t}{\includegraphics[width=0.15\columnwidth]{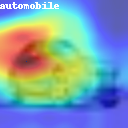}}&
\adjustbox{valign=t}{\includegraphics[width=0.15\columnwidth]{img/CAM_ResNet20/auto/ResNet20_AMC_kernel_prune_overlay_1604.png}}&
\adjustbox{valign=t}{\includegraphics[width=0.15\columnwidth]{img/CAM_ResNet20/auto/ResNet20_AMC_weight_prune_overlay_1604.png}}&
\adjustbox{valign=t}{\includegraphics[width=0.15\columnwidth]{img/CAM_ResNet20/auto/ResNet20_XNOR_overlay_1604.png}}&
\adjustbox{valign=t}{\includegraphics[width=0.15\columnwidth]{img/CAM_ResNet20/auto/ResNet20_ABC1x1_overlay_1604.png}}&
\adjustbox{valign=t}{\includegraphics[width=0.15\columnwidth]{img/CAM_ResNet20/auto/ResNet20_ABC3x3_overlay_1604.png}}&
\adjustbox{valign=t}{\includegraphics[width=0.15\columnwidth]{img/CAM_ResNet20/auto/ResNet20_ABC5x5_overlay_1604.png}}\\
%\adjustbox{valign=t}{\includegraphics[width=0.15\columnwidth]{img/CAM_ResNet20/auto/ResNet20_ABC7x7_overlay_1604.png}}\\
\adjustbox{valign=t}{\rotatebox{90}{\makecell{$i=1$}}}&
\fcolorbox{red}{red}{\adjustbox{valign=t}{\includegraphics[width=0.15\columnwidth]{img/CAM_ResNet20/aa_auto_i1/i1_ResNet20_Vanilla_adv_image_1604.png}}}&
\adjustbox{valign=t}{\includegraphics[width=0.15\columnwidth]{img/CAM_ResNet20/aa_auto_i1/ResNet20_overlay_1604.png}}&
%\adjustbox{valign=t}{\includegraphics[width=0.15\columnwidth]{img/CAM_ResNet20/aa_auto_i1/ResNet20_KD_CE_overlay_1604.png}}&
\adjustbox{valign=t}{\includegraphics[width=0.15\columnwidth]{img/CAM_ResNet20/aa_auto_i1/ResNet20_KD_KL_overlay_1604.png}}&
\fcolorbox{red}{red}{\adjustbox{valign=t}{\includegraphics[width=0.15\columnwidth]{img/CAM_ResNet20/aa_auto_i1/ResNet20_AMC_channel_prune_overlay_1604.png}}}&
\adjustbox{valign=t}{\includegraphics[width=0.15\columnwidth]{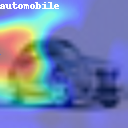}}&
\adjustbox{valign=t}{\includegraphics[width=0.15\columnwidth]{img/CAM_ResNet20/aa_auto_i1/ResNet20_AMC_kernel_prune_overlay_1604.png}}&
\adjustbox{valign=t}{\includegraphics[width=0.15\columnwidth]{img/CAM_ResNet20/aa_auto_i1/ResNet20_AMC_weight_prune_overlay_1604.png}}&
\adjustbox{valign=t}{\includegraphics[width=0.15\columnwidth]{img/CAM_ResNet20/aa_auto_i1/ResNet20_XNOR_overlay_1604.png}}&
\adjustbox{valign=t}{\includegraphics[width=0.15\columnwidth]{img/CAM_ResNet20/aa_auto_i1/ResNet20_ABC1x1_overlay_1604.png}}&
\adjustbox{valign=t}{\includegraphics[width=0.15\columnwidth]{img/CAM_ResNet20/aa_auto_i1/ResNet20_ABC3x3_overlay_1604.png}}&
\fcolorbox{red}{red}{\adjustbox{valign=t}{\includegraphics[width=0.15\columnwidth]{img/CAM_ResNet20/aa_auto_i1/ResNet20_ABC5x5_overlay_1604.png}}}\\
%\adjustbox{valign=t}{\includegraphics[width=0.15\columnwidth]{img/CAM_ResNet20/aa_auto_i1/ResNet20_ABC7x7_overlay_1604.png}}\\
\adjustbox{valign=t}{\rotatebox{90}{\makecell{$i=5$}}}&
\fcolorbox{red}{red}{\adjustbox{valign=t}{\includegraphics[width=0.15\columnwidth]{img/CAM_ResNet20/aa_auto_i5/i5_ResNet20_Vanilla_adv_image_1604.png}}}&
\fcolorbox{red}{red}{\adjustbox{valign=t}{\includegraphics[width=0.15\columnwidth]{img/CAM_ResNet20/aa_auto_i5/ResNet20_overlay_1604.png}}}&
%\fcolorbox{red}{red}{\adjustbox{valign=t}{\includegraphics[width=0.15\columnwidth]{img/CAM_ResNet20/aa_auto_i5/ResNet20_KD_CE_overlay_1604.png}}}&
\fcolorbox{red}{red}{\adjustbox{valign=t}{\includegraphics[width=0.15\columnwidth]{img/CAM_ResNet20/aa_auto_i5/ResNet20_KD_KL_overlay_1604.png}}}&
\fcolorbox{red}{red}{\adjustbox{valign=t}{\includegraphics[width=0.15\columnwidth]{img/CAM_ResNet20/aa_auto_i5/ResNet20_AMC_channel_prune_overlay_1604.png}}}&
\fcolorbox{red}{red}{\adjustbox{valign=t}{\includegraphics[width=0.15\columnwidth]{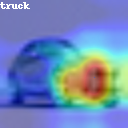}}}&
\fcolorbox{red}{red}{\adjustbox{valign=t}{\includegraphics[width=0.15\columnwidth]{img/CAM_ResNet20/aa_auto_i5/ResNet20_AMC_kernel_prune_overlay_1604.png}}}&
\fcolorbox{red}{red}{\adjustbox{valign=t}{\includegraphics[width=0.15\columnwidth]{img/CAM_ResNet20/aa_auto_i5/ResNet20_AMC_weight_prune_overlay_1604.png}}}&
\adjustbox{valign=t}{\includegraphics[width=0.15\columnwidth]{img/CAM_ResNet20/aa_auto_i5/ResNet20_XNOR_overlay_1604.png}}&
\fcolorbox{red}{red}{\adjustbox{valign=t}{\includegraphics[width=0.15\columnwidth]{img/CAM_ResNet20/aa_auto_i5/ResNet20_ABC1x1_overlay_1604.png}}}&
\fcolorbox{red}{red}{\adjustbox{valign=t}{\includegraphics[width=0.15\columnwidth]{img/CAM_ResNet20/aa_auto_i5/ResNet20_ABC3x3_overlay_1604.png}}}&
\fcolorbox{red}{red}{\adjustbox{valign=t}{\includegraphics[width=0.15\columnwidth]{img/CAM_ResNet20/aa_auto_i5/ResNet20_ABC5x5_overlay_1604.png}}}\\
%\fcolorbox{red}{red}{\adjustbox{valign=t}{\includegraphics[width=0.15\columnwidth]{img/CAM_ResNet20/aa_auto_i5/ResNet20_ABC7x7_overlay_1604.png}}}\\
\midrule
\adjustbox{valign=t}{\rotatebox{90}{\makecell{No AA}}}&
\adjustbox{valign=t}{\includegraphics[width=0.15\columnwidth]{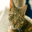}}&
\adjustbox{valign=t}{\includegraphics[width=0.15\columnwidth]{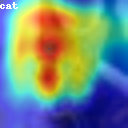}}&
%\adjustbox{valign=t}{\includegraphics[width=0.15\columnwidth]{img/CAM_ResNet20/cat/ResNet20_KD_CE_overlay_103.png}}&
\adjustbox{valign=t}{\includegraphics[width=0.15\columnwidth]{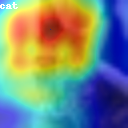}}&
\adjustbox{valign=t}{\includegraphics[width=0.15\columnwidth]{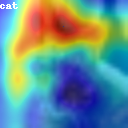}}&
\adjustbox{valign=t}{\includegraphics[width=0.15\columnwidth]{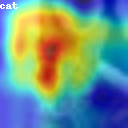}}&
\adjustbox{valign=t}{\includegraphics[width=0.15\columnwidth]{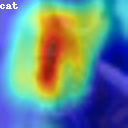}}&
\adjustbox{valign=t}{\includegraphics[width=0.15\columnwidth]{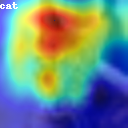}}&
\adjustbox{valign=t}{\includegraphics[width=0.15\columnwidth]{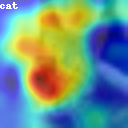}}&
\adjustbox{valign=t}{\includegraphics[width=0.15\columnwidth]{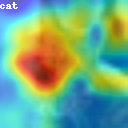}}&
\adjustbox{valign=t}{\includegraphics[width=0.15\columnwidth]{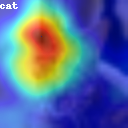}}&
\adjustbox{valign=t}{\includegraphics[width=0.15\columnwidth]{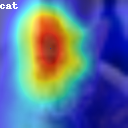}}\\
%\adjustbox{valign=t}{\includegraphics[width=0.15\columnwidth]{img/CAM_ResNet20/cat/ResNet20_ABC7x7_overlay_103.png}}\\
\adjustbox{valign=t}{\rotatebox{90}{\makecell{$i=1$}}}&
\fcolorbox{red}{red}{\adjustbox{valign=t}{\includegraphics[width=0.15\columnwidth]{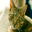}}}&
\fcolorbox{red}{red}{\adjustbox{valign=t}{\includegraphics[width=0.15\columnwidth]{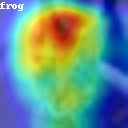}}}&
%\fcolorbox{red}{red}{\adjustbox{valign=t}{\includegraphics[width=0.15\columnwidth]{img/CAM_ResNet20/aa_cat_i1/ResNet20_KD_CE_overlay_103.png}}}&
\fcolorbox{red}{red}{\adjustbox{valign=t}{\includegraphics[width=0.15\columnwidth]{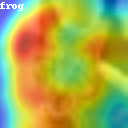}}}&
\fcolorbox{red}{red}{\adjustbox{valign=t}{\includegraphics[width=0.15\columnwidth]{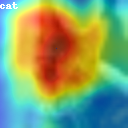}}}&
\adjustbox{valign=t}{\includegraphics[width=0.15\columnwidth]{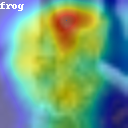}}&
\adjustbox{valign=t}{\includegraphics[width=0.15\columnwidth]{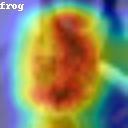}}&
\adjustbox{valign=t}{\includegraphics[width=0.15\columnwidth]{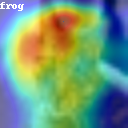}}&
\fcolorbox{red}{red}{\adjustbox{valign=t}{\includegraphics[width=0.15\columnwidth]{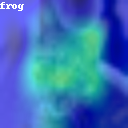}}}&
\fcolorbox{red}{red}{\adjustbox{valign=t}{\includegraphics[width=0.15\columnwidth]{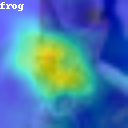}}}&
\fcolorbox{red}{red}{\adjustbox{valign=t}{\includegraphics[width=0.15\columnwidth]{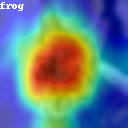}}}&
\fcolorbox{red}{red}{\adjustbox{valign=t}{\includegraphics[width=0.15\columnwidth]{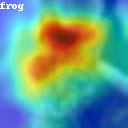}}}\\
%\fcolorbox{red}{red}{\adjustbox{valign=t}{\includegraphics[width=0.15\columnwidth]{img/CAM_ResNet20/aa_cat_i1/ResNet20_ABC7x7_overlay_103.png}}}\\
\adjustbox{valign=t}{\rotatebox{90}{\makecell{$i=5$}}}&
\fcolorbox{red}{red}{\adjustbox{valign=t}{\includegraphics[width=0.15\columnwidth]{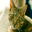}}}&
\fcolorbox{red}{red}{\adjustbox{valign=t}{\includegraphics[width=0.15\columnwidth]{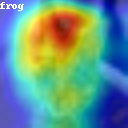}}}&
%\fcolorbox{red}{red}{\adjustbox{valign=t}{\includegraphics[width=0.15\columnwidth]{img/CAM_ResNet20/aa_cat_i5/ResNet20_KD_CE_overlay_103.png}}}&
\fcolorbox{red}{red}{\adjustbox{valign=t}{\includegraphics[width=0.15\columnwidth]{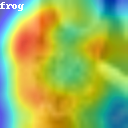}}}&
\fcolorbox{red}{red}{\adjustbox{valign=t}{\includegraphics[width=0.15\columnwidth]{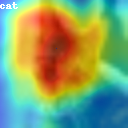}}}&
\adjustbox{valign=t}{\includegraphics[width=0.15\columnwidth]{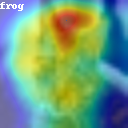}}&
\adjustbox{valign=t}{\includegraphics[width=0.15\columnwidth]{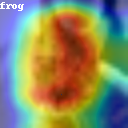}}&
\adjustbox{valign=t}{\includegraphics[width=0.15\columnwidth]{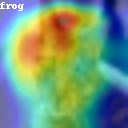}}&
\fcolorbox{red}{red}{\adjustbox{valign=t}{\includegraphics[width=0.15\columnwidth]{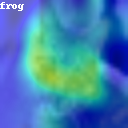}}}&
\fcolorbox{red}{red}{\adjustbox{valign=t}{\includegraphics[width=0.15\columnwidth]{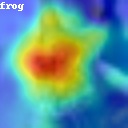}}}&
\fcolorbox{red}{red}{\adjustbox{valign=t}{\includegraphics[width=0.15\columnwidth]{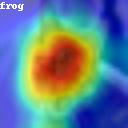}}}&
\fcolorbox{red}{red}{\adjustbox{valign=t}{\includegraphics[width=0.15\columnwidth]{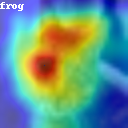}}}\\
\bottomrule
\end{tabular}}
\vspace{0.2ex}
\caption{CAM for ResNet20 and its compressed variants performed on non-attacked and DeepFool attacked images.}
\label{tab:cam_resnet20}
\end{table}
    \begin{table}[ht]
\centering
\setlength{\tabcolsep}{2pt}
\resizebox{\textwidth}{!}{
\begin{tabular}{lc|cc|c|c}
\toprule
&&\multicolumn{2}{c|}{\textbf{Vanilla}}&\multicolumn{1}{c|}{\textbf{Pruned}}&\multicolumn{1}{c}{\textbf{Binary}}\\
%\multirow{2}{*}{\textbf{Model}}&No AA & \multicolumn{2}{c}{DeepFool\cite{Deepfool}}&No AA&\multicolumn{2}{c}{DeepFool\cite{Deepfool}} \\
%&$L$=$Car$&$i$=$1$&$i$=$5$&$L$=$Cat$&$i$=$1$&$i$=$5$\\
%\textbf{DeepFool}
\multicolumn{2}{c|}{\textbf{Image$\rightarrow$ \fcolorbox{red}{white}{$I^{Adv}$}}}&
ResNet50&
ResNet18&
ResNet18 - Channel&
ResNet18 - XNOR\\
&&(75.43 \%)&
(69.00 \%)&
(67.62 \%)&
(49.10 \%)\\
\midrule
\midrule
\adjustbox{valign=t}{\rotatebox{90}{\makecell{No AA}}}&
\adjustbox{valign=t}{\includegraphics[width=0.25\columnwidth]{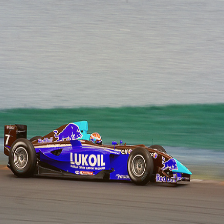}}&
\adjustbox{valign=t}{\includegraphics[width=0.25\columnwidth]{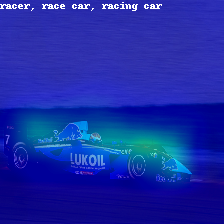}}&
\adjustbox{valign=t}{\includegraphics[width=0.25\columnwidth]{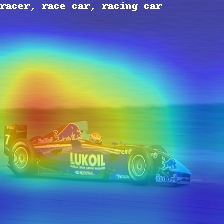}}&
\adjustbox{valign=t}{\includegraphics[width=0.25\columnwidth]{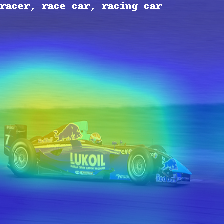}}&
\adjustbox{valign=t}{\includegraphics[width=0.25\columnwidth]{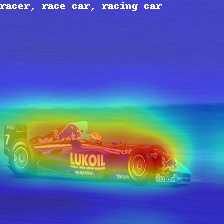}}\\
\adjustbox{valign=t}{\rotatebox{90}{\makecell{$i=1$}}}&
\fcolorbox{red}{red}{\adjustbox{valign=t}{\includegraphics[width=0.25\columnwidth]{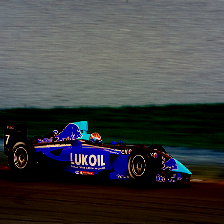}}}&
\adjustbox{valign=t}{\includegraphics[width=0.25\columnwidth]{img/CAM_ImageNet/aa_racecar_i1/ResNet50_overlay_28.png}}&
\adjustbox{valign=t}{\includegraphics[width=0.25\columnwidth]{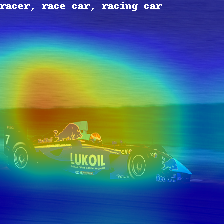}}&
\fcolorbox{red}{red}{\adjustbox{valign=t}{\includegraphics[width=0.25\columnwidth]{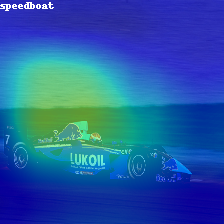}}}&
\fcolorbox{red}{red}{\adjustbox{valign=t}{\includegraphics[width=0.25\columnwidth]{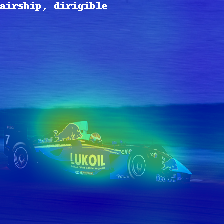}}}\\
\adjustbox{valign=t}{\rotatebox{90}{\makecell{$i=2$}}}&
\fcolorbox{red}{red}{\adjustbox{valign=t}{\includegraphics[width=0.25\columnwidth]{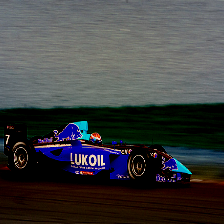}}}&
\fcolorbox{red}{red}{\adjustbox{valign=t}{\includegraphics[width=0.25\columnwidth]{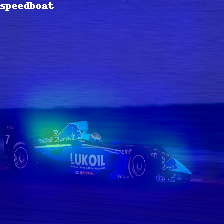}}}&
\adjustbox{valign=t}{\includegraphics[width=0.25\columnwidth]{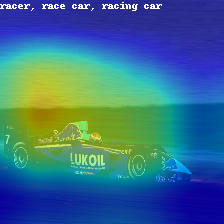}}&
\fcolorbox{red}{red}{\adjustbox{valign=t}{\includegraphics[width=0.25\columnwidth]{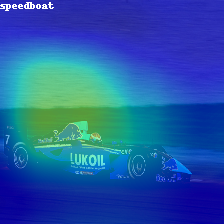}}}&
\fcolorbox{red}{red}{\adjustbox{valign=t}{\includegraphics[width=0.25\columnwidth]{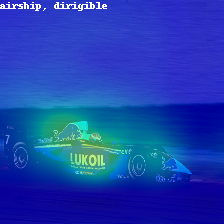}}}\\
\bottomrule
\end{tabular}}
\vspace{0.2ex}
\caption{CAM for CNNs trained on ImageNet performed on PGD attacked images.}
\label{tab:cam_imagenet}
\end{table}

\section{Stress-Strain Graphs}
    Fig.~\ref{fig:Attack_graphs_supp} shows additional stress-strain graphs for different attacks (a-e) on compressed variants of ResNet20  and ResNet56, as referred to in the paper.

\begin{figure}[b]
\captionsetup[subfigure]{labelformat=empty}
\centering
\subfloat{
\hspace{10ex}
       \begin{tikzpicture}	
	\begin{axis}[
		name=legend,
		width=\textwidth,
		height=20ex,
		hide axis,
        legend style={at={(0,0)},anchor=west, legend columns=-1, draw = none, nodes={scale=0.75, transform shape}, column sep=2pt},
		ymin=0,
        ymax=1,
        xmin=0,
        xmax=1,
		]
		\addlegendimage{Pal_1, mark=x,mark options={scale=1.25}, line width=1pt, only marks}
       	\addlegendentry{Vanilla};
       	\addlegendimage{Pal_2, mark=square,mark options={scale=1.25}, line width=1pt, only marks}
		\addlegendentry{Ch.Prune};
		%\addlegendimage{Pal_7, mark=square,mark options={scale=1}, line width=1pt, only marks}
		%\addlegendentry{F.Prune};
		\addlegendimage{Pal_7,mark=square,mark options={scale=1.25}, line width=1pt, only marks}
        \addlegendentry{K.Prune};
        \addlegendimage{Pal_6,mark=square,mark options={scale=1.25}, line width=1pt, only marks}
		\addlegendentry{W.Prune};
		\addlegendimage{Pal_4, mark=diamond,mark options={scale=1.25}, line width=1pt, only marks}
		\addlegendentry{XNOR};
		\addlegendimage{Pal_5, mark=triangle,mark options={scale=1.25}, line width=1pt, only marks}
		\addlegendentry{ABC($1\times1$)};
		\addlegendimage{Pal_6,mark=triangle,mark options={scale=1.25}, line width=1pt, only marks}
		\addlegendentry{ABC($3\times3$)};
		\addlegendimage{Pal_8,mark=triangle,mark options={scale=1.25}, line width=1pt, only marks}
		\addlegendentry{ABC($5\times5$)};
		%\addlegendimage{Pal_3,mark=triangle,mark options={scale=1}, line width=1pt, only marks }
		%\addlegendentry{ABC($7\times7$)};
		%\addlegendimage{Pal_1 ,mark=o,mark options={scale=1}, line width=1pt, only marks}
		%\addlegendentry{KD-CE};
		\addlegendimage{Pal_2,mark=o,mark options={scale=1.25}, line width=1pt, only marks}
		\addlegendentry{KD-KL};
	\end{axis}
\end{tikzpicture}
} \\\vspace{-2ex}
\subfloat[PGD - Fixed $\epsilon=0.5$ ResNet20]{
       \pgfplotstableread[col sep=space]{%
iter	Vanilla KDCE    KDKL    AMCCH   AMCF    AMCK   AMCW    XNOR    ABC1    ABC2    ABC3    ABC4
0   92.46   92.70   93.25   89.76   89.91   90.73   91.99   82.71   83.42   88.94   90.64   91.34
2   35.30   33.84   55.36   29.89   31.03   31.28   35.47   60.09   56.87   61.46   58.54   58.11
3   15.61   14.12   39.40   12.30   13.57   12.93   15.92   47.01   41.54   44.47   39.93   38.76
4   6.40    5.10    27.78   4.53    5.20    4.66    6.36    34.97   27.84   29.03   25.07   24.70
5   2.20    1.97    19.60   1.56    1.87    1.51    2.13    24.28   17.08   17.24   15.09   15.54
}\pgdresnettwentycifarfixedepsilon

\begin{tikzpicture}
    \begin{axis}[
          height=0.18\textwidth,
          width=0.2\textwidth, % Scale the plot to \linewidth
          grid=major, 
          grid style={dashed,gray!30},
          xlabel= Attack Stress ($\sigma$), % Set the labels
          ylabel= Strain ($\varepsilon$),
          xmin=0,
          xmax=5,
          ymin=0,
          ymax=1,
          smooth,
          tension=0.2,
          yticklabel style = {font=\tiny, rotate=90, yshift=-0.3ex},
          xticklabel style = {font=\tiny},
		  ylabel style = {font=\tiny, yshift=-5.75ex},
		  xlabel style = {font=\tiny, yshift=2ex},
		  legend style={at={(1,0.24)},anchor=east, legend columns=1, fill=none ,draw = none, nodes={scale=0.38, transform shape}, column sep=5pt},
		  scale only axis,
		  every axis plot/.append style={line width=0.7pt, mark options={scale=1}},
          cycle list name=mycolorlist,
        ]
       \addplot table[x index = 0, y expr = {(92.46-\thisrow{Vanilla})/92.46}] {\pgdresnettwentycifarfixedepsilon};
       
       \addplot table[x index = 0, y expr = {(89.76-\thisrow{AMCCH})/89.76}] {\pgdresnettwentycifarfixedepsilon};  
       
       %\addplot table[x index = 0, y expr = {(89.91-\thisrow{AMCF})/89.91}] {\pgdresnettwentycifarfixedepsilon};
       
       \addplot table[x index = 0, y expr = {(90.73-\thisrow{AMCK})/90.73}] {\pgdresnettwentycifarfixedepsilon};  
       
       \addplot table[x index = 0, y expr = {(91.99-\thisrow{AMCW})/91.99}] {\pgdresnettwentycifarfixedepsilon}; 
       
       \addplot table[x index = 0, y expr = {(82.71-\thisrow{XNOR})/82.71}] {\pgdresnettwentycifarfixedepsilon};  
       
       \addplot table[x index = 0, y expr = {(83.42-\thisrow{ABC1})/83.42}] {\pgdresnettwentycifarfixedepsilon};  
       
       \addplot table[x index = 0, y expr = {(88.94-\thisrow{ABC2})/88.94}] {\pgdresnettwentycifarfixedepsilon};  
       
       \addplot table[x index = 0, y expr = {(90.64-\thisrow{ABC3})/90.64}] {\pgdresnettwentycifarfixedepsilon};  
       %\pgfplotsset{cycle list shift=3}
       %\addplot table[x index = 0, y expr = {(91.34-\thisrow{ABC4})/91.34}] {\pgdresnettwentycifarfixedepsilon};  
       
       %\addplot table[x index = 0, y expr = {(92.70-\thisrow{KDCE})/92.70}] {\pgdresnettwentycifarfixedepsilon};  
       
       \addplot table[x index = 0, y expr = {(93.25-\thisrow{KDKL})/93.25}] {\pgdresnettwentycifarfixedepsilon};   
       
       	\addlegendentry{Vanilla};
		\addlegendentry{AMC-CH};
		%\addlegendentry{AMC-F};
		\addlegendentry{AMC-K};
		\addlegendentry{AMC-W};
		\addlegendentry{XNOR};
		\addlegendentry{ABC($1\times1$)};
		\addlegendentry{ABC($3\times3$)};
		\addlegendentry{ABC($5\times5$)};
		%\addlegendentry{ABC($7\times7$)};
		%\addlegendentry{KD-CE};
		\addlegendentry{KD-KL};
		\legend{};
		%		\node[align=center, text width=2cm] (N) at (axis cs: 4,0.3) %{ductile};
		%\draw[->] (N) -- (axis cs: 3.5,0.45);
    \end{axis}

\end{tikzpicture} 
       }
\subfloat[LocalSearch - ResNet20 Fixed $\epsilon=16$]{
       \pgfplotstableread[col sep=space]{%
gen	Vanilla KDCE KDKL AMCCH AMCF AMCK AMCW XNOR ABC1 ABC2 ABC3 ABC4
0   92.46	92.70   93.25	89.76	89.91   90.74    91.99	82.71	83.42   88.94   90.64   91.34
50	46.70   43.37   51.00   42.76   41.49   44.00    46.93   59.62   61.54   65.84   67.49   66.99
100	32.26   27.49   35.60   30.05   28.10   30.47    33.10   47.80   49.22   52.04   55.82   54.62
150	24.27   20.44   27.58   23.94   20.92   23.10    25.38   40.58   40.57   42.99   47.29   46.21
200	18.99   15.74   22.37   18.80   16.62   18.66    20.45   36.73   34.85   36.37   41.13   40.48
}\LSresnettwentycifar

\begin{tikzpicture}
    \begin{axis}[
          height=0.22\textwidth,
          width=0.25\textwidth, % Scale the plot to \linewidth
          grid=major, 
          grid style={dashed,gray!30},
          xlabel= Attack Stress ($\sigma$), % Set the labels
          ylabel= Strain ($\varepsilon$),
          xmin=0,
          xmax=200,
          ymin=0,
          ymax=1,
          smooth,
          tension=0.2,
          yticklabel style = {font=\tiny, rotate=90, yshift=-0.3ex},
          xticklabel style = {font=\tiny},
		  ylabel style = {font=\tiny, yshift=-5.75ex},
		  xlabel style = {font=\tiny, yshift=2ex},
		  legend style={at={(1,0.5)},anchor=east, legend columns=1, fill=none ,draw = none, nodes={scale=0.38, transform shape}, column sep=5pt},
		  scale only axis,
		  every axis plot/.append style={line width=0.7pt, mark options={scale=1}},
          cycle list name=mycolorlist,
        ]
       \addplot table[x index = 0, y expr = {(92.46-\thisrow{Vanilla})/92.46}] {\LSresnettwentycifar};
       
       \addplot table[x index = 0, y expr = {(89.76-\thisrow{AMCCH})/89.76}] {\LSresnettwentycifar};  
       
       %\addplot table[x index = 0, y expr = {(89.91-\thisrow{AMCF})/89.91}] {\LSresnettwentycifar};  
       
       \addplot table[x index = 0, y expr = {(90.74-\thisrow{AMCK})/90.74}] {\LSresnettwentycifar};  
       
       \addplot table[x index = 0, y expr = {(91.99-\thisrow{AMCW})/91.99}] {\LSresnettwentycifar};
       
       \addplot table[x index = 0, y expr = {(82.71-\thisrow{XNOR})/82.71}] {\LSresnettwentycifar};  
       
       \addplot table[x index = 0, y expr = {(83.42-\thisrow{ABC1})/83.42}] {\LSresnettwentycifar};  
       
       \addplot table[x index = 0, y expr = {(88.94-\thisrow{ABC2})/88.94}] {\LSresnettwentycifar};  
       
       \addplot table[x index = 0, y expr = {(90.64-\thisrow{ABC3})/90.64}] {\LSresnettwentycifar};  
       %\pgfplotsset{cycle list shift=3}
       %\addplot table[x index = 0, y expr = {(91.34-\thisrow{ABC4})/91.34}] {\LSresnettwentycifar};  
       
       %\addplot table[x index = 0, y expr = {(92.70-\thisrow{KDCE})/92.70}] {\LSresnettwentycifar};  
       
       \addplot table[x index = 0, y expr = {(93.25-\thisrow{KDKL})/93.25}] {\LSresnettwentycifar};  
       
       	\addlegendentry{Vanilla};
		\addlegendentry{AMC-CH};
		%\addlegendentry{AMC-F};
		\addlegendentry{AMC-K};
		\addlegendentry{AMC-W};
		\addlegendentry{XNOR};
		\addlegendentry{ABC($1\times1$)};
		\addlegendentry{ABC($3\times3$)};
		\addlegendentry{ABC($5\times5$)};
		%\addlegendentry{ABC($7\times7$)};
		%\addlegendentry{KD-CE};
		\addlegendentry{KD-KL};
		\legend{};
    \end{axis}

\end{tikzpicture} 
       }
\subfloat[LocalSearch - ResNet20 Fixed $\epsilon=32$]{
       \pgfplotstableread[col sep=space]{%
gen	Vanilla KDCE KDKL AMCCH AMCF AMCK AMCW XNOR ABC1 ABC2 ABC3 ABC4
0   92.46	92.70   93.25	89.76	89.91   90.74  91.99	 82.71	83.42   88.94   90.64   91.34
50	11.58   10.40   24.10   6.81    6.95    8.02    9.21   51.84   54.68   53.79   53.65   50.81
100	1.79    1.38    9.09    1.23    1.40    1.60    2.27   39.70   41.11   36.27   36.56   32.48
150	0.52    0.49    4.11    0.43    0.47    0.74    0.94   32.73   34.02   27.91   28.49   24.79
200	0.28    0.22    2.41    0.29    0.25    0.55    0.48   29.32   28.50   23.04   23.73   20.45
}\LSresnettwentycifarethirtytwo

\begin{tikzpicture}
    \begin{axis}[
          height=0.18\textwidth,
          width=0.2\textwidth, % Scale the plot to \linewidth
          grid=major, 
          grid style={dashed,gray!30},
          xlabel= Attack Stress ($\sigma$), % Set the labels
          ylabel= Strain ($\varepsilon$),
          xmin=0,
          xmax=200,
          ymin=0,
          ymax=1,
          smooth,
          tension=0.2,
          yticklabel style = {font=\tiny, rotate=90, yshift=-0.3ex},
          xticklabel style = {font=\tiny},
		  ylabel style = {font=\tiny, yshift=-5.75ex},
		  xlabel style = {font=\tiny, yshift=2ex},
		  legend style={at={(1,0.5)},anchor=east, legend columns=1, fill=none ,draw = none, nodes={scale=0.38, transform shape}, column sep=5pt},
		  scale only axis,
		  every axis plot/.append style={line width=0.7pt, mark options={scale=1}},
          cycle list name=mycolorlist,
        ]
       \addplot table[x index = 0, y expr = {(92.46-\thisrow{Vanilla})/92.46}] {\LSresnettwentycifarethirtytwo};
       
       \addplot table[x index = 0, y expr = {(89.76-\thisrow{AMCCH})/89.76}] {\LSresnettwentycifarethirtytwo};  
       
       %\addplot table[x index = 0, y expr = {(89.91-\thisrow{AMCF})/89.91}] {\LSresnettwentycifarethirtytwo};  
       
       \addplot table[x index = 0, y expr = {(90.74-\thisrow{AMCK})/90.74}] {\LSresnettwentycifarethirtytwo};  
       
       \addplot table[x index = 0, y expr = {(91.99-\thisrow{AMCW})/91.99}] {\LSresnettwentycifarethirtytwo};  
       
       \addplot table[x index = 0, y expr = {(82.71-\thisrow{XNOR})/82.71}] {\LSresnettwentycifarethirtytwo};  
       
       \addplot table[x index = 0, y expr = {(83.42-\thisrow{ABC1})/83.42}] {\LSresnettwentycifarethirtytwo};  
       
       \addplot table[x index = 0, y expr = {(88.94-\thisrow{ABC2})/88.94}] {\LSresnettwentycifarethirtytwo};  
       
       \addplot table[x index = 0, y expr = {(90.64-\thisrow{ABC3})/90.64}] {\LSresnettwentycifarethirtytwo};  
       %\pgfplotsset{cycle list shift=3}
       %\addplot table[x index = 0, y expr = {(91.34-\thisrow{ABC4})/91.34}] {\LSresnettwentycifarethirtytwo};  
       
       %\addplot table[x index = 0, y expr = {(92.70-\thisrow{KDCE})/92.70}] {\LSresnettwentycifarethirtytwo};  
       
       \addplot table[x index = 0, y expr = {(93.25-\thisrow{KDKL})/93.25}] {\LSresnettwentycifarethirtytwo};  
       
       	\addlegendentry{Vanilla};
		\addlegendentry{AMC-CH};
		%\addlegendentry{AMC-F};
		\addlegendentry{AMC-K};
		\addlegendentry{AMC-W};
		\addlegendentry{XNOR};
		\addlegendentry{ABC($1\times1$)};
		\addlegendentry{ABC($3\times3$)};
		\addlegendentry{ABC($5\times5$)};
		%\addlegendentry{ABC($7\times7$)};
		%\addlegendentry{KD-CE};
		\addlegendentry{KD-KL};
		\legend{};
    \end{axis}

\end{tikzpicture} 
       }
       \\\vspace{-2ex}
\subfloat[PGD - Fixed $\epsilon=0.5$ ResNet56]{
       \pgfplotstableread[col sep=space]{%
iter Vanilla KDCE KDKL AMCCH AMCF AMCK AMCW XNOR ABC1 ABC2 ABC3
0	93.88   93.68 94.24	92.93   93.08   93.04    93.54	83.24	86.29	92.48	92.82
2	51.89   73.88 76.83	49.46   50.87   47.97    51.06	70.29	57.50   61.74	61.71
3	33.65   72.35 74.23	32.58   32.64   29.38    32.45	62.97	39.41   43.47	43.75
4	20.92   73.13 73.49	20.40   20.07   16.86    19.98	55.16	24.86   28.99	29.22
5	12.19   72.40 73.11	12.53   12.38   9.29     11.35	47.95 	14.20   18.48	18.77
}\pgdresnetcifarfixedepsilon

\begin{tikzpicture}
    \begin{axis}[
          height=0.18\textwidth,
          width=0.2\textwidth, % Scale the plot to \linewidth
          grid=major, 
          grid style={dashed,gray!30},
          xlabel= Attack Stress ($\sigma$), % Set the labels
          ylabel= Strain ($\varepsilon$),
          xmin=0,
          xmax=5,
          ymin=0,
          ymax=1,
          smooth,
          tension=0.2,
          yticklabel style = {font=\tiny, rotate=90, yshift=-0.3ex},
          xticklabel style = {font=\tiny},
		  ylabel style = {font=\tiny, yshift=-5.75ex},
		  xlabel style = {font=\tiny, yshift=2ex},
		  legend style={at={(0.48,0.7)},anchor=east, legend columns=1, fill=none ,draw = none, nodes={scale=0.5, transform shape}, column sep=5pt},
		  scale only axis,
		  every axis plot/.append style={line width=0.7pt, mark options={scale=1}},
          cycle list name=mycolorlist,
        ]
       \addplot table[x index = 0, y expr = {(93.88-\thisrow{Vanilla})/93.88}] {\pgdresnetcifarfixedepsilon};
       
       \addplot table[x index = 0, y expr = {(92.93-\thisrow{AMCCH})/92.93}] {\pgdresnetcifarfixedepsilon};  
       
       %\addplot table[x index = 0, y expr = {(93.08-\thisrow{AMCF})/93.08}] {\pgdresnetcifarfixedepsilon};  
       
       \addplot table[x index = 0, y expr = {(93.04-\thisrow{AMCK})/93.04}] {\pgdresnetcifarfixedepsilon};  
       
       \addplot table[x index = 0, y expr = {(93.54-\thisrow{AMCW})/93.54}] {\pgdresnetcifarfixedepsilon};
       
       \addplot table[x index = 0, y expr = {(83.24-\thisrow{XNOR})/83.24}] {\pgdresnetcifarfixedepsilon};  
       
       \addplot table[x index = 0, y expr = {(86.29-\thisrow{ABC1})/86.29}] {\pgdresnetcifarfixedepsilon};  
       
       \addplot table[x index = 0, y expr = {(92.48-\thisrow{ABC2})/92.48}] {\pgdresnetcifarfixedepsilon};  
       
       \addplot table[x index = 0, y expr = {(92.82-\thisrow{ABC3})/92.82}] {\pgdresnetcifarfixedepsilon};
       %\pgfplotsset{cycle list shift=3}
       %\addplot table[x index = 0, y expr = {(93.68-\thisrow{KDCE})/93.68}] {\pgdresnetcifarfixedepsilon};  
       \addplot table[x index = 0, y expr = {(94.24-\thisrow{KDKL})/94.24}] {\pgdresnetcifarfixedepsilon};
       
       	\addlegendentry{Vanilla};
		\addlegendentry{AMC-CH};
		%\addlegendentry{AMC-F};
        \addlegendentry{AMC-K};
        \addlegendentry{AMC-W};
		\addlegendentry{XNOR};
		\addlegendentry{ABC($1\times1$)};
		\addlegendentry{ABC($3\times3$)};
		\addlegendentry{ABC($5\times5$)};
		%\addlegendentry{KD-CE};
		\addlegendentry{KD-KL};
		\legend{};
	\node[align=center] (N) at (axis cs: 4,0.07) {\tiny strong};
	\draw[->] (axis cs: 4,0.085) -- (axis cs: 4,0.2);
	
	\node[align=center] (N2) at (axis cs: 4,0.55) {\tiny ductile};
	\draw[->] (N2) -- (axis cs: 4,0.37);
	
	\node[align=center] (N3) at (axis cs: 3,0.88) {\tiny brittle};
	\draw[->] (axis cs: 3.45,0.86) -- (axis cs: 3.8,0.8);
    \end{axis}
\end{tikzpicture} 
       }
\subfloat[LocalSearch - ResNet56 Fixed $\epsilon=16$]{
       \pgfplotstableread[col sep=space]{%
gen Vanilla KDCE KDKL AMCCH AMCF AMCK AMCW XNOR ABC1 ABC2 ABC3
0	93.88   93.68  94.24   92.93   93.08    93.04  93.54   83.24	86.29	92.48	92.82
50	54.94   52.89  56.48   53.65   53.18    49.20  52.24   63.70   62.73   71.45   71.78
100	39.88   37.23  39.83   39.34   37.95    33.73  36.00   53.50   48.23   58.32   60.32
150	31.70   28.62  29.48   30.17   28.66    24.27  28.12   47.22   36.27   48.95   51.47
200	26.27   22.73  22.38   24.56   23.49    19.04  22.25   42.52   27.71   41.92   44.45
}\LSresnetcifar

\begin{tikzpicture}
    \begin{axis}[
          height=0.22\textwidth,
          width=0.25\textwidth, % Scale the plot to \linewidth
          grid=major, 
          grid style={dashed,gray!30},
          xlabel= Attack Stress ($\sigma$), % Set the labels
          ylabel= Strain ($\varepsilon$),
          xmin=0,
          xmax=200,
          ymin=0,
          ymax=1,
          smooth,
          tension=0.2,
          yticklabel style = {font=\tiny, rotate=90, yshift=-0.3ex},
          xticklabel style = {font=\tiny},
		  ylabel style = {font=\tiny, yshift=-5.75ex},
		  xlabel style = {font=\tiny, yshift=2ex},
		  legend style={at={(0.48,0.78)},anchor=east, legend columns=1, fill=none ,draw = none, nodes={scale=0.5, transform shape}, column sep=5pt},
		  scale only axis,
		  every axis plot/.append style={line width=0.7pt, mark options={scale=1}},
          cycle list name=mycolorlist,
        ]
       \addplot table[x index = 0, y expr = {(93.88-\thisrow{Vanilla})/93.88}] {\LSresnetcifar};
       
       \addplot table[x index = 0, y expr = {(92.93-\thisrow{AMCCH})/92.93}] {\LSresnetcifar}; 
       
       %\addplot table[x index = 0, y expr = {(93.08-\thisrow{AMCF})/93.08}] {\LSresnetcifar};  
       
       \addplot table[x index = 0, y expr = {(93.04-\thisrow{AMCK})/93.04}] {\LSresnetcifar}; 
       
       \addplot table[x index = 0, y expr = {(93.54-\thisrow{AMCW})/93.54}] {\LSresnetcifar};  
       
       \addplot table[x index = 0, y expr = {(83.24-\thisrow{XNOR})/83.24}] {\LSresnetcifar};  
       
       \addplot table[x index = 0, y expr = {(86.29-\thisrow{ABC1})/86.29}] {\LSresnetcifar};  
       
       \addplot table[x index = 0, y expr = {(92.48-\thisrow{ABC2})/92.48}] {\LSresnetcifar};  
       
       \addplot table[x index = 0, y expr = {(92.82-\thisrow{ABC3})/92.82}] {\LSresnetcifar};  
       %\pgfplotsset{cycle list shift=3}
       %\addplot table[x index = 0, y expr = {(93.68-\thisrow{KDCE})/93.68}] {\LSresnetcifar};  
       
       \addplot table[x index = 0, y expr = {(94.24-\thisrow{KDKL})/94.24}] {\LSresnetcifar};  
       
       	\addlegendentry{Vanilla};
		\addlegendentry{AMC-CH};
		%\addlegendentry{AMC-F};
		\addlegendentry{AMC-K};
		\addlegendentry{AMC-W};
		\addlegendentry{XNOR};
		\addlegendentry{ABC($1\times1$)};
		\addlegendentry{ABC($3\times3$)};
		\addlegendentry{ABC($5\times5$)};
		%\addlegendentry{KD-CE};
		\addlegendentry{KD-KL};
		\legend{};
    \end{axis}

\end{tikzpicture} 
       }
\subfloat[LocalSearch - ResNet56 Fixed $\epsilon=32$]{
       \pgfplotstableread[col sep=space]{%
gen Vanilla KDCE KDKL AMCCH AMCF AMCK AMCW XNOR ABC1 ABC2 ABC3
0	93.88   93.68   94.24   92.93   93.08   93.04  93.54	83.24	86.29	92.48	92.82
50	22.82   22.63   20.95   23.99   21.93   16.56  19.21	56.39   55.97   61.34   60.77
100	8.27    5.30    10.61   9.57    7.64    4.03    5.30	44.30   39.33   43.27   44.40
150	4.17    1.75    4.34    5.06    3.90    1.45    2.12	38.19   29.35   34.05   35.42
200	2.67    0.77    2.20    2.98    2.48    0.77    0.92	33.19   22.79   28.52   30.88
}\LSresnetcifarethirtytwo

\begin{tikzpicture}
    \begin{axis}[
          height=0.18\textwidth,
          width=0.2\textwidth, % Scale the plot to \linewidth
          grid=major, 
          grid style={dashed,gray!30},
          xlabel= Attack Stress ($\sigma$), % Set the labels
          ylabel= Strain ($\varepsilon$),
          xmin=0,
          xmax=200,
          ymin=0,
          ymax=1,
          smooth,
          tension=0.2,
          yticklabel style = {font=\tiny, rotate=90, yshift=-0.3ex},
          xticklabel style = {font=\tiny},
		  ylabel style = {font=\tiny, yshift=-5.75ex},
		  xlabel style = {font=\tiny, yshift=2ex},
		  legend style={at={(0.48,0.78)},anchor=east, legend columns=1, fill=none ,draw = none, nodes={scale=0.5, transform shape}, column sep=5pt},
		  scale only axis,
		  every axis plot/.append style={line width=0.7pt, mark options={scale=1}},
          cycle list name=mycolorlist,
        ]
       \addplot table[x index = 0, y expr = {(93.88-\thisrow{Vanilla})/93.88}] {\LSresnetcifarethirtytwo};
       
       \addplot table[x index = 0, y expr = {(92.93-\thisrow{AMCCH})/92.93}] {\LSresnetcifarethirtytwo}; 
       
       %\addplot table[x index = 0, y expr = {(93.08-\thisrow{AMCF})/93.08}] {\LSresnetcifarethirtytwo};  
       
       \addplot table[x index = 0, y expr = {(93.04-\thisrow{AMCK})/93.04}] {\LSresnetcifarethirtytwo}; 
       
       \addplot table[x index = 0, y expr = {(93.54-\thisrow{AMCW})/93.54}] {\LSresnetcifarethirtytwo}; 
       
       \addplot table[x index = 0, y expr = {(83.24-\thisrow{XNOR})/83.24}] {\LSresnetcifarethirtytwo};  
       
       \addplot table[x index = 0, y expr = {(86.29-\thisrow{ABC1})/86.29}] {\LSresnetcifarethirtytwo};  
       
       \addplot table[x index = 0, y expr = {(92.48-\thisrow{ABC2})/92.48}] {\LSresnetcifarethirtytwo};  
       
       \addplot table[x index = 0, y expr = {(92.82-\thisrow{ABC3})/92.82}] {\LSresnetcifarethirtytwo}; 
       %\pgfplotsset{cycle list shift=3}
       %\addplot table[x index = 0, y expr = {(93.68-\thisrow{KDCE})/93.68}] {\LSresnetcifarethirtytwo};  
       
       \addplot table[x index = 0, y expr = {(94.24-\thisrow{KDKL})/94.24}] {\LSresnetcifarethirtytwo}; 
       
       	\addlegendentry{Vanilla};
		\addlegendentry{AMC-CH};
		%\addlegendentry{AMC-F};
		\addlegendentry{AMC-K};
		\addlegendentry{AMC-W};
		\addlegendentry{XNOR};
		\addlegendentry{ABC($1\times1$)};
		\addlegendentry{ABC($3\times3$)};
		\addlegendentry{ABC($5\times5$)};
		%\addlegendentry{KD-CE};
		\addlegendentry{KD-KL};
		\legend{};
    \end{axis}

\end{tikzpicture} 
       }
       \\\vspace{-2ex}
\subfloat[CW - ResNet20 Fixed $\epsilon=1$]{
       \pgfplotstableread[col sep=space]{%
gen	Vanilla KDCE KDKL AMCCH AMCF AMCK AMCW XNOR ABC1 ABC2 ABC3 ABC4
0   92.46	92.70   93.25	89.76	89.91   90.73  91.99	82.71	83.42   88.94   90.64   91.34
1	11.67   10.63   11.70   10.96   11.07   12.99  11.96   11.68   12.18   11.18   10.82   10.55
10	8.02    9.05    10.34   5.70    5.77    11.70  10.78   10.96   10.64   9.71    9.10    8.55
20	4.58    6.22    8.73    2.42    2.07    10.51  10.14   9.66    8.01    7.17    6.63    5.16
50	1.28    2.13    6.75    0.55    0.28    10.01  9.94   5.93    2.99    2.42    2.51    1.53
}\CWresnettwentycifar

\begin{tikzpicture}
    \begin{axis}[
          height=0.22\textwidth,
          width=0.25\textwidth, % Scale the plot to \linewidth
          grid=major, 
          grid style={dashed,gray!30},
          xlabel= Attack Stress ($\sigma$), % Set the labels
          ylabel= Strain ($\varepsilon$),
          xmin=0,
          xmax=50,
          ymin=0,
          ymax=1,
          smooth,
          tension=0.2,
          yticklabel style = {font=\tiny, rotate=90, yshift=-0.3ex},
          xticklabel style = {font=\tiny},
		  ylabel style = {font=\tiny, yshift=-5.75ex},
		  xlabel style = {font=\tiny, yshift=2ex},
		  legend style={at={(1,0.5)},anchor=east, legend columns=1, fill=none ,draw = none, nodes={scale=0.38, transform shape}, column sep=5pt},
		  scale only axis,
		  every axis plot/.append style={line width=0.7pt, mark options={scale=1}},
          cycle list name=mycolorlist,
        ]
       \addplot table[x index = 0, y expr = {(92.46-\thisrow{Vanilla})/92.46}] {\CWresnettwentycifar};
       
       \addplot table[x index = 0, y expr = {(89.76-\thisrow{AMCCH})/89.76}] {\CWresnettwentycifar};  
       
       %\addplot table[x index = 0, y expr = {(89.91-\thisrow{AMCF})/89.91}] {\CWresnettwentycifar};  
       
       \addplot table[x index = 0, y expr = {(90.73-\thisrow{AMCK})/90.73}] {\CWresnettwentycifar};  
       
       \addplot table[x index = 0, y expr = {(91.99-\thisrow{AMCW})/91.99}] {\CWresnettwentycifar}; 
       
       \addplot table[x index = 0, y expr = {(82.71-\thisrow{XNOR})/82.71}] {\CWresnettwentycifar};  
       
       \addplot table[x index = 0, y expr = {(83.42-\thisrow{ABC1})/83.42}] {\CWresnettwentycifar};  
       
       \addplot table[x index = 0, y expr = {(88.94-\thisrow{ABC2})/88.94}] {\CWresnettwentycifar};  
       
       \addplot table[x index = 0, y expr = {(90.64-\thisrow{ABC3})/90.64}] {\CWresnettwentycifar};  
       %\pgfplotsset{cycle list shift=3}
       %\addplot table[x index = 0, y expr = {(91.34-\thisrow{ABC4})/91.34}] {\CWresnettwentycifar};  
       
       %\addplot table[x index = 0, y expr = {(92.70-\thisrow{KDCE})/92.70}] {\CWresnettwentycifar};  
       
       \addplot table[x index = 0, y expr = {(93.25-\thisrow{KDKL})/93.25}] {\CWresnettwentycifar};  
       
       	\addlegendentry{Vanilla};
		\addlegendentry{AMC-CH};
		%\addlegendentry{AMC-F};
		\addlegendentry{AMC-K};
		\addlegendentry{AMC-W};
		\addlegendentry{XNOR};
		\addlegendentry{ABC($1\times1$)};
		\addlegendentry{ABC($3\times3$)};
		\addlegendentry{ABC($5\times5$)};
		%\addlegendentry{ABC($7\times7$)};
		%\addlegendentry{KD-CE};
		\addlegendentry{KD-KL};
		\legend{};
    \end{axis}

\end{tikzpicture} 
       }
\subfloat[CW - ResNet56 Fixed $\epsilon=1$]{
       \pgfplotstableread[col sep=space]{%
gen	Vanilla KDCE KDKL AMCCH AMCF AMCK AMCW XNOR ABC1 ABC2 ABC3 ABC4
0   92.46	92.70   93.25	89.76	89.91   90.73  91.99	82.71   83.42   88.94   90.64   91.34
50	8.43    6.30    18.87   5.56    5.81    5.92    8.88    63.57   60.71   62.39   65.39   67.39
100	1.12    0.11    9.66    0.34    0.30    0.53    1.29    60.92   59.08   59.59   62.79   64.92
150	0.30    0.01    6.08    0.03    0.05    0.04    0.24    59.51   57.09   58.59   60.78   63.20
200	0.09    0.00    4.61    0.01    0.00    0.02    0.06    58.88   56.29   57.28   59.59   61.97
}\genresnettwentycifaretwelve

\begin{tikzpicture}
    \begin{axis}[
          height=0.22\textwidth,
          width=0.25\textwidth, % Scale the plot to \linewidth
          grid=major, 
          grid style={dashed,gray!30},
          xlabel= Attack Stress ($\sigma$), % Set the labels
          ylabel= Strain ($\varepsilon$),
          xmin=0,
          xmax=200,
          ymin=0,
          ymax=1,
          smooth,
          tension=0.2,
          yticklabel style = {font=\tiny, rotate=90, yshift=-0.3ex},
          xticklabel style = {font=\tiny},
		  ylabel style = {font=\tiny, yshift=-5.75ex},
		  xlabel style = {font=\tiny, yshift=2ex},
		  legend style={at={(1,0.5)},anchor=east, legend columns=1, fill=none ,draw = none, nodes={scale=0.38, transform shape}, column sep=5pt},
		  scale only axis,
		  every axis plot/.append style={line width=0.7pt, mark options={scale=1}},
          cycle list name=mycolorlist,
        ]
       \addplot table[x index = 0, y expr = {(92.46-\thisrow{Vanilla})/92.46}] {\genresnettwentycifaretwelve};
       
       \addplot table[x index = 0, y expr = {(89.76-\thisrow{AMCCH})/89.76}] {\genresnettwentycifaretwelve};  
       
       %\addplot table[x index = 0, y expr = {(89.91-\thisrow{AMCF})/89.91}] {\genresnettwentycifaretwelve};  
       
       \addplot table[x index = 0, y expr = {(90.73-\thisrow{AMCK})/90.73}] {\genresnettwentycifaretwelve};  
       
       \addplot table[x index = 0, y expr = {(91.99-\thisrow{AMCW})/91.99}] {\genresnettwentycifaretwelve};  
       
       \addplot table[x index = 0, y expr = {(82.71-\thisrow{XNOR})/82.71}] {\genresnettwentycifaretwelve};  
       
       \addplot table[x index = 0, y expr = {(83.42-\thisrow{ABC1})/83.42}] {\genresnettwentycifaretwelve};  
       
       \addplot table[x index = 0, y expr = {(88.94-\thisrow{ABC2})/88.94}] {\genresnettwentycifaretwelve};  
       
       \addplot table[x index = 0, y expr = {(90.64-\thisrow{ABC3})/90.64}] {\genresnettwentycifaretwelve};  
       %\pgfplotsset{cycle list shift=3}
       %\addplot table[x index = 0, y expr = {(91.34-\thisrow{ABC4})/91.34}] {\genresnettwentycifaretwelve};  
       
       %\addplot table[x index = 0, y expr = {(92.70-\thisrow{KDCE})/92.70}] {\genresnettwentycifaretwelve};  
       
       \addplot table[x index = 0, y expr = {(93.25-\thisrow{KDKL})/93.25}] {\genresnettwentycifaretwelve};  
       
       	\addlegendentry{Vanilla};
		\addlegendentry{AMC-CH};
		%\addlegendentry{AMC-F};
		\addlegendentry{AMC-K};
		\addlegendentry{AMC-W};
		\addlegendentry{XNOR};
		\addlegendentry{ABC($1\times1$)};
		\addlegendentry{ABC($3\times3$)};
		\addlegendentry{ABC($5\times5$)};
		%\addlegendentry{ABC($7\times7$)};
		%\addlegendentry{KD-CE};
		\addlegendentry{KD-KL};
		\legend{};
    \end{axis}

\end{tikzpicture} 
       }\\\vspace{-2ex}
\subfloat[GenAttack - ResNet20 - $\epsilon=8$ $|$ $p=16$]{
       \pgfplotstableread[col sep=space]{%
iter Vanilla KDCE KDKL AMCCH AMCF AMCK AMCW XNOR ABC1 ABC2 ABC3
0	93.88   93.68  94.24	92.93   93.08   93.04  93.54	83.24	86.29	92.48	92.82
1	10.35   12.20  12.12	10.53   10.23   10.95  10.76	10.58	10.39   9.85	10.27
10	9.17    10.79  11.50     9.81    9.03   11.20  10.14	10.58	9.99    9.48    9.93
20	7.71    11.14  10.84     8.34    8.01   11.12  10.71	9.94	10.16   8.42	9.01
50	4.64    10.38  10.97     5.25    3.94   10.81  10.05   9.29 	8.23    5.40	6.41
}\cwresnetcifar

\begin{tikzpicture}
    \begin{axis}[
          height=0.22\textwidth,
          width=0.25\textwidth, % Scale the plot to \linewidth
          grid=major, 
          grid style={dashed,gray!30},
          xlabel= Attack Stress ($\sigma$), % Set the labels
          ylabel= Strain ($\varepsilon$),
          xmin=0,
          xmax=50,
          ymin=0,
          ymax=1,
          %smooth,
          %enlargelimits=false,
          %tension=0.2,
          yticklabel style = {font=\tiny, rotate=90, yshift=-0.3ex},
          xticklabel style = {font=\tiny},
		  ylabel style = {font=\tiny, yshift=-5.75ex},
		  xlabel style = {font=\tiny, yshift=2ex},
		  legend style={at={(0.98,0.22)},anchor=east, legend columns=1, fill=none ,draw = none, nodes={scale=0.5, transform shape}, column sep=5pt},
		  scale only axis,
		  every axis plot/.append style={line width=0.7pt, mark options={scale=1}},
          cycle list name=mycolorlist,
        ]
       \addplot table[x index = 0, y expr = {(93.88-\thisrow{Vanilla})/93.88}] {\cwresnetcifar};
       
       \addplot table[x index = 0, y expr = {(92.93-\thisrow{AMCCH})/92.93}] {\cwresnetcifar};  
       
       %\addplot table[x index = 0, y expr = {(93.08-\thisrow{AMCF})/93.08}] {\cwresnetcifar};  
       
       \addplot table[x index = 0, y expr = {(93.04-\thisrow{AMCK})/93.04}] {\cwresnetcifar};  
       
       \addplot table[x index = 0, y expr = {(93.54-\thisrow{AMCW})/93.54}] {\cwresnetcifar};  
       
       \addplot table[x index = 0, y expr = {(83.24-\thisrow{XNOR})/83.24}] {\cwresnetcifar};  
       
       \addplot table[x index = 0, y expr = {(86.29-\thisrow{ABC1})/86.29}] {\cwresnetcifar};  
       
       \addplot table[x index = 0, y expr = {(92.48-\thisrow{ABC2})/92.48}] {\cwresnetcifar};  
       
       \addplot table[x index = 0, y expr = {(92.82-\thisrow{ABC3})/92.82}] {\cwresnetcifar};   
       %\pgfplotsset{cycle list shift=3}
       %\addplot table[x index = 0, y expr = {(93.68-\thisrow{KDCE})/93.68}] {\cwresnetcifar};  
       
       \addplot table[x index = 0, y expr = {(94.24-\thisrow{KDKL})/94.24}] {\cwresnetcifar};   
       
       	\addlegendentry{Vanilla};
		\addlegendentry{AMC-CH};
		%\addlegendentry{AMC-F};
		\addlegendentry{AMC-K};
		\addlegendentry{AMC-W};
		\addlegendentry{XNOR};
		\addlegendentry{ABC($1\times1$)};
		\addlegendentry{ABC($3\times3$)};
		\addlegendentry{ABC($5\times5$)};
        %\addlegendentry{KD-CE};
		\addlegendentry{KD-KL};
		\legend{}
    \end{axis}

\end{tikzpicture} 
       }
\subfloat[GenAttack - ResNet56 - $\epsilon=8$ $|$ $p=16$]{
       \pgfplotstableread[col sep=space]{%
gen Vanilla KDCE KDKL AMCCH AMCF AMCK AMCW XNOR ABC1 ABC2 ABC3
0	93.88   93.67   94.24   92.93   93.08   93.04   93.54   83.24	86.29	92.48	92.82
50	18.01   21.25   28.95   17.99   19.20   13.88   17.50   60.56   60.29   69.23   69.91
100	4.54    6.66    12.58   4.53    6.74    2.76    4.61    58.18   57.81   66.55   67.34
150	1.48    2.27    6.20    1.52    2.58    0.85    1.47    56.74   56.81   64.55   65.77
200	0.49    1.02    3.43    0.53    0.88    0.35    0.40    55.47   55.13   63.48   64.80
}\genresnetcifaretwelve

\begin{tikzpicture}
    \begin{axis}[
          height=0.18\textwidth,
          width=0.2\textwidth, % Scale the plot to \linewidth
          grid=major, 
          grid style={dashed,gray!30},
          xlabel= Attack Stress ($\sigma$), % Set the labels
          ylabel= Strain ($\varepsilon$),
          xmin=0,
          xmax=200,
          ymin=0,
          ymax=1,
          smooth,
          tension=0.2,
          yticklabel style = {font=\tiny, rotate=90, yshift=-0.3ex},
          xticklabel style = {font=\tiny},
		  ylabel style = {font=\tiny, yshift=-5.75ex},
		  xlabel style = {font=\tiny, yshift=2ex},
		  legend style={at={(0.98,0.5)},anchor=east, legend columns=1, fill=none ,draw = none, nodes={scale=0.5, transform shape}, column sep=5pt},
		  scale only axis,
		  every axis plot/.append style={line width=0.7pt, mark options={scale=1}},
          cycle list name=mycolorlist,
        ]
       \addplot table[x index = 0, y expr = {(93.88-\thisrow{Vanilla})/93.88}] {\genresnetcifaretwelve};
       
       \addplot table[x index = 0, y expr = {(92.93-\thisrow{AMCCH})/92.93}] {\genresnetcifaretwelve};  
       
       %\addplot table[x index = 0, y expr = {(93.08-\thisrow{AMCF})/93.08}] {\genresnetcifaretwelve};  
       
       \addplot table[x index = 0, y expr = {(93.04-\thisrow{AMCK})/93.04}] {\genresnetcifaretwelve};  
       
       \addplot table[x index = 0, y expr = {(93.54-\thisrow{AMCW})/93.54}] {\genresnetcifaretwelve}; 
       
       \addplot table[x index = 0, y expr = {(83.24-\thisrow{XNOR})/83.24}] {\genresnetcifaretwelve};  
       
       \addplot table[x index = 0, y expr = {(86.29-\thisrow{ABC1})/86.29}] {\genresnetcifaretwelve};  
       
       \addplot table[x index = 0, y expr = {(92.48-\thisrow{ABC2})/92.48}] {\genresnetcifaretwelve};  
       
       \addplot table[x index = 0, y expr = {(92.82-\thisrow{ABC3})/92.82}] {\genresnetcifaretwelve};  
       %\pgfplotsset{cycle list shift=3}
       %\addplot table[x index = 0, y expr = {(93.67-\thisrow{KDCE})/93.67}] {\genresnetcifaretwelve};  
       
       \addplot table[x index = 0, y expr = {(94.24-\thisrow{KDKL})/94.24}] {\genresnetcifaretwelve}; 
       
       	\addlegendentry{Vanilla};
		\addlegendentry{AMC-CH};
		%\addlegendentry{AMC-F};
		\addlegendentry{AMC-K};
		\addlegendentry{AMC-W};
		\addlegendentry{XNOR};
		\addlegendentry{ABC($1\times1$)};
		\addlegendentry{ABC($3\times3$)};
		\addlegendentry{ABC($5\times5$)};
		%\addlegendentry{KD-CE};
		\addlegendentry{KD-KL};
		\legend{};

    \end{axis}

\end{tikzpicture} 
       }
\caption{Stress-strain graphs for attacks (a-d) on compressed variants of ResNet20 (Top) and ResNet56 (Bottom).
%\alex{REMOVE: BatchNorm~\cite{bn} layers are in $\mathtt{statistics}$ mode.}
}
\vspace{-1ex}
\label{fig:Attack_graphs_supp}
\end{figure}

\section{Table of Comparisons}
    Tab.~\ref{tab:compressed_cnns} summarizes the compressed CNNs and their full-precision counterparts analyzed in this paper. It shows that the neural networks drastically vary in their memory demand and their compute complexity. Deep learning inference accelerators such as the NVIDIA-T4 GPU~\cite{nvidiat4} or Xilinx FPGAs with DSP48 blocks support SIMD-based bit-wise operations~\cite{orthruspe}. In particular, a single DSP48 block can perform two 16-bit fixed-point multiplications or 48 XNOR operations at once. 
% % Thus,24 XNOR operations can be normalized as one fixed point multiplication. 
The normalized compute complexity (NCC) is defined as the optimal utilization of MAC and XNOR operations in one compute unit. The DSP48 block serves as a reference implementation to compute NCC in Tab.~\ref{tab:compressed_cnns}.
%~\cite{nvidia_t4}
%~\cite{XilDSP48E1}
\begin{table*}[ht]
  \centering
  \resizebox{0.70\textwidth}{!}{
  \renewcommand{\arraystretch}{1.0}
  \begin{tabular}{ll|ccc} 
    \toprule
    \textbf{Dataset}&\textbf{Model} & \textbf{Acc. [\%]} & \textbf{Memory demand [MB]} & \textbf{NCC} [$10^6$]\\
    \midrule
    \midrule
    \multirow{23}{*}{\rotatebox{90}{\makecell{CIFAR-10}}} 
    &\textbf{ResNet20}~\cite{He2015DeepRL}           &92.46 \%   &1.07   &41\\
    %&\textbf{KD-CE}~\cite{contrastive_kd}           &92.70 \%   &1.07   &41\\
    &\textbf{KD-KL}~\cite{contrastive_kd}        &93.25 \%   &1.07   &41\\
    &\textbf{Ch.Prune}~\cite{AMC2019}           &89.76 \%   &0.70   &19\\
    %&\textbf{F.Prune}~\cite{AMC2019}            &89.91 \%   &0.82   &21\\
    &\textbf{K.Prune}~\cite{AMC2019}            &90.73 \%   &0.61   &20\\
    &\textbf{W.Prune}~\cite{AMC2019}            &91.98 \%   &0.59   &20\\
    &\textbf{XNOR}~\cite{rastegari2016}              &82.71 \%   &0.04   &1.3\\
    &\textbf{ABC(1$\times$1)}~\cite{lin2017}         &83.42 \%   &0.04   &1.3\\
    &\textbf{ABC(3$\times$3)}~\cite{lin2017}         &88.94 \%   &0.12   &8.0\\
    &\textbf{ABC(5$\times$5)}~\cite{lin2017}         &90.64 \%   &0.20   &21.3\\
    %&\textbf{ABC(7$\times$7)}~\cite{lin2017}         &91.34 \%   &0.28   &41.3\\
    \cmidrule{2-5}
    &\textbf{ResNet56}~\cite{He2015DeepRL}           &93.88 \%   &3.40   &125\\
    %&\textbf{KD-CE}~\cite{contrastive_kd}           &93.68 \%   &3.40   &125\\
    &\textbf{KD-KL}~\cite{contrastive_kd}        &94.24 \%   &3.40   &125\\
    &\textbf{Ch.Prune}~\cite{AMC2019}           &92.86 \%   &2.50   &62\\
    %&\textbf{F.Prune}~\cite{AMC2019}            &93.09 \%   &2.29   &64\\
    &\textbf{K.Prune}~\cite{AMC2019}            &93.04 \%   &2.19   &63\\
    &\textbf{W.Prune}~\cite{AMC2019}            &93.54 \%   &2.02   &62\\
    &\textbf{XNOR}~\cite{rastegari2016}              &83.24 \%   &0.11   &3.0\\
    &\textbf{ABC(1$\times$1)}~\cite{lin2017}         &86.29 \%   &0.11   &3.0\\
    &\textbf{ABC(3$\times$3)}~\cite{lin2017}         &92.48 \%   &0.33   &24\\
    &\textbf{ABC(5$\times$5)}~\cite{lin2017}         &92.82 \%   &0.55   &66\\
    \midrule
    \multirow{6}{*}{\rotatebox{90}{\makecell{ImageNet}}}
    &\textbf{ResNet50}~\cite{He2015DeepRL}           &75.43 \%   &102.01 &10216\\
    &\textbf{ResNet18}~\cite{He2015DeepRL}           &69.00 \%   &46.72  &1814\\
    &\textbf{ResNet18-Ch.Prune}~\cite{AMC2019}  &67.62 \%   &34.52  &884\\
    &\textbf{ResNet18-XNOR}~\cite{rastegari2016}     &49.10 \%   &4.14   &173\\
    &\textbf{ResNet18-ABC(1$\times$1)}~\cite{lin2017}&51.07 \%   &3.48   &153\\
    &\textbf{ResNet18-ABC(3$\times$3)}~\cite{lin2017}&59.83 \%   &6.28   &417\\
    \bottomrule 
    \end{tabular}}
    \vspace{1em}
    \caption{Accuracy Top1 [\%], Memory demand [MB] and the normalized compute complexity (NCC) of compressed CNNs and their full-precision counterparts.}
  \label{tab:compressed_cnns}
\end{table*}

    \subsection{Adversarial Attacks on FGSM}
    The experiments examined on FGSM are summarized in Tab.~\ref{tab:FGSMResnet20} and Tab.~\ref{tab:FGSMResnet56} for compressed variants of ResNet20 and ResNet56. In Tab.~\ref{tab:FGSMImageNet} the results for CNNs trained on ImageNet are given. 
%The accuracy is given for the BatchNorm's $\mathtt{mini}$-$\mathtt{batch}$ and the $\mathtt{statistics}$ mode after applying the attack.
FGSM is applied for the adversarial manipulation budget $\epsilon\in \{2,4,8,16\}$.
    
    \subsection{Adversarial Attacks on PGD}
    The adversarial attacks examined w.r.t. PGD are summarized in Tab.~\ref{tab:PGDResnet20_eval}, \ref{tab:PGDResnet56_eval} and \ref{tab:PGDImageNet_eval} for ResNet20, ResNet56 and ImageNet-trained CNNs. 
%Tab.~\ref{tab:PGDResnet20_train}, \ref{tab:PGDResnet56_train} and \ref{tab:PGDImageNet_train} details the results for the BatchNorm's $\mathtt{mini}$-$\mathtt{batch}$ mode. 
PGD is applied for $\epsilon\in \{0.1,0.5,1.0,2.0\}$ multiple iterations $i\in\{2,3,4,5\}$.
    
    \subsection{Adversarial Attacks on Carlini \& Wagner}
    The adversarial attacks examined w.r.t. C\&W on ResNet20, 56 and its compressed variants are summarized in Tab.~\ref{tab:CWResnet20_eval} and \ref{tab:CWResnet56_eval} respectively.
%For the BatchNorm's $\mathtt{mini}$-$\mathtt{batch}$ mode, the experiments are given in Tab.~\ref{tab:CWResnet20_train} and \ref{tab:CWResnet56_train}.
The regularization factor is $c\in\{0.01,0.1,1.0,5.0,10.0\}$ and the attack is run for $i\in\{1,10,20,50\}$ iterations for the target class $t=deer$ of the CIFAR-10 dataset.
    
    \subsection{Adversarial Attacks on DeepFool}
    The adversarial attacks examined w.r.t. DeepFool on compressed versions of ResNet20 and ResNet56 are summarized in Tab.~\ref{tab:DeepFoolResnet20_full} and \ref{tab:DeepFoolResnet56_full} respectively for $i=\{1,5,10,20\}$ iterations.
    
    % %  __             __        __   __      
    % % |__) |     /\  /  ` |__/ |__) /  \ \_/ 
    % % |__) |___ /~~\ \__, |  \ |__) \__/ / \
    \subsection{Adversarial Attacks on LocalSearch}
    Tab.~\ref{tab:LocalSearchResNet20_eval}, \ref{tab:LocalSearchResNet56_eval} and \ref{tab:LocalSearchImageNet_eval} show the accuracy of compressed versions of ResNet20, ResNet56 and ResNet18 when attacked by LocalSearch.
%for the BatchNorm's $\mathtt{statistics}$ mode.
%For the BatchNorm's $\mathtt{mini}$-$\mathtt{batch}$ mode, the experiments are given in Tab.~\ref{tab:LocalSearchResNet20_train}, \ref{tab:LocalSearchResNet56_train} and \ref{tab:LocalSearchImageNet_train}.
The attack is run for $i\in\{50,100,150,200\}$ iterations respectively.

    \subsection{Adversarial Attacks on GenAttack}
    GenAttack results of different CNN models trained on CIFAR-10 and ImageNet datasets are presented in Tab.~\ref{tab:GenAttack_ResNet20_eval}, \ref{tab:GenAttackResNet56_eval} and \ref{tab:GenAttackImageNet_eval}. For the CNN models trained on CIFAR-10 dataset, we choose the number of iterations $i$ as 200.
%For the BatchNorm's $\mathtt{mini}$-$\mathtt{batch}$ mode, the experiments are given in Tab.~\ref{tab:GenAttack_ResNet20_train}, \ref{tab:GenAttack_ResNet20_train} and \ref{tab:GenAttackImageNet_train}.
For GenAttacks on CIFAR-10 experiments, the mutation probability $\rho$ and step size $\alpha$ are set to $0.05$ and $1.0$ respectively as demonstrated in ~\cite{GenAttack}. 
Elitism is applied for the highest fitness member. On ImageNet experiments $\rho$ and $\alpha$ are adaptive based on the dataset configuration and the population size is 6, as in~\cite{GenAttack}. Results are shown for $\epsilon=8$ and $12$ with both population sizes $N=6$ and 16.

\begin{table*}[hbt!]
  \centering
  \resizebox{0.5\textwidth}{!}{
  \renewcommand{\arraystretch}{1.0}
\begin{tabular}{l|l|cccc} 
    \toprule
    \textbf{FGSM} &\textbf{Nat. Acc} & $\epsilon=2$ & $\epsilon=4$ & $\epsilon=8$ & $\epsilon=16$\\
    \midrule
    \midrule
    \textbf{Vanilla}~\cite{He2015DeepRL}    
                                  &92.46 \%             &27.21  &16.96  &11.79  &10.04\\
    % \midrule
    % \textbf{KD-CE}~\cite{contrastive_kd}   &$\mathtt{mini}$-$\mathtt{batch}$   &26.80  &16.04  &11.53  &11.08\\
    % (92.70 \%)                              &$\mathtt{statistics}$              &26.56  &16.36  &11.73  &9.95\\
    % \midrule
    \textbf{KD-KL}~\cite{contrastive_kd}
                                 &93.25 \%              &53.01  &43.30  &32.54  &18.69\\
    % \midrule
    \textbf{Ch.Prune}~\cite{AMC2019}  
                                  &89.76 \%             &22.31  &12.63  &9.29   &9.50\\
    % \midrule
    % \textbf{F.Prune}~\cite{AMC2019}    &$\mathtt{mini}$-$\mathtt{batch}$   &22.83  &12.47  &8.42   &8.35\\
    % (89.91 \%)                              &$\mathtt{statistics}$              &21.53  &11.69  &8.45   &8.14 \\
    % \midrule
    \textbf{K.Prune}~\cite{AMC2019}    
                                  &90.73 \%              &23.21  &13.10  &8.99   &8.72\\
    % \midrule
    \textbf{W.Prune}~\cite{AMC2019}    
                                &91.99 \%             &26.68  &15.46  &10.49  &8.77\\
    % \midrule
    \textbf{XNOR}~\cite{rastegari2016}      
                                  &82.71 \%             &42.84 &21.71  &8.40    &4.85\\
    % \midrule
    \textbf{ABC(1$\times$1)}~\cite{lin2017} 
                                  &83.42 \%            &38.78  &20.32  &10.93  &6.80\\
\bottomrule                                
    \end{tabular}}
  \vspace{1em}
  \caption{Accuracy (Top1) [\%] of compressed variants of ResNet20 after FGSM adversarial attacks for CIFAR-10.}
  \label{tab:FGSMResnet20}
\end{table*}

    %\nael{XNOR performs best - has lowest baseline Acc, but highest after attack. However with increasing $\epsilon$, it degrades quickly. All BNNs degrade quickly compared to AMC and vanilla with increased $\epsilon$}
%log files: /storage/public/Attacks/cifar10/FGSM/ResNet56

\begin{table*}[hbt!]
  \centering
  \resizebox{0.5\textwidth}{!}{
  \renewcommand{\arraystretch}{1.0}
  \begin{tabular}{l|l|cccc} 
    \toprule
    \textbf{FGSM} &\textbf{Nat.Acc} & $\epsilon=2$ & $\epsilon=4$ & $\epsilon=8$ & $\epsilon=16$\\
    \midrule
    \midrule
    \textbf{Vanilla}~\cite{He2015DeepRL}    &93.88 \%
    %&47.03  &36.46  &29.32  &25.31\\%46.67&36.55&29.40&25.34\\
                              %&$\mathtt{statistics}$              
                              &48.63  &37.67  &27.18&17.45\\%48.68&37.71&27.15&17.59\\
    %\midrule
    % \textbf{KD-CE}~\cite{contrastive_kd}   &$\mathtt{mini}$-$\mathtt{batch}$   &51.11  &41.97  &36.01  &31.59\\
    % (93.68 \%)                              &$\mathtt{statistics}$              &51.92  &42.02  &32.92  &19.85\\
    %\midrule
    \textbf{KD-KL}~\cite{contrastive_kd}
    %&$\mathtt{mini}$-$\mathtt{batch}$   &59.03  &51.56  &46.47  &42.22\\
                                 &94.24 \%              &60.24  &52.86  &45.36  &27.65\\
    %\midrule
    \textbf{Ch.Prune}~\cite{AMC2019}   &92.93 \%   
    %&46.53  &36.96  &30.61  &26.49\\%46.52 &37.07&30.88&26.61\\
                                 %&$\mathtt{statistics}$              
                                 &48.27  &38.47  &28.21  &18.42\\%47.95 &38.55&28.28&18.48\\
    %\midrule
    % \textbf{F.Prune}~\cite{AMC2019}      &$\mathtt{mini}$-$\mathtt{batch}$ &46.11  &36.17  &29.21  &25.10\\
    % (93.08 \%)                              &$\mathtt{statistics}$              &48.12  &37.72  &26.99  &17.15\\
    %\midrule
    \textbf{K.Prune}~\cite{AMC2019}    &93.04 \%   
    %&41.48  &30.13  &22.81  &19.07\\
                                 %&$\mathtt{statistics}$              
                                 &42.65  &30.08  &20.31  &13.67\\
    %\midrule
    \textbf{W.Prune}~\cite{AMC2019}    &93.54 \%   
    %&45.85  &34.94  &27.75  &23.44\\
                                 %&$\mathtt{statistics}$              
                                 &47.75  &35.55  &24.51  &15.68\\
\bottomrule                                
    \end{tabular}}
    \vspace{1em}
    \caption{Accuracy (Top1) [\%] of compressed variants of ResNet56 after FGSM adversarial attacks for CIFAR-10.}
    \label{tab:FGSMResnet56}
\end{table*}

    \begin{table*}[t]
  \centering
  \resizebox{0.6\textwidth}{!}{
  \renewcommand{\arraystretch}{1.0}
  \begin{tabular}{l|l|cccc} 
    \toprule
    \textbf{FGSM} &\textbf{Nat.Acc} & $\epsilon=2$ & $\epsilon=4$ & $\epsilon=8$ & $\epsilon=16$\\
    \midrule
    \midrule
    \textbf{ResNet50}~\cite{He2015DeepRL}    
    %&$\mathtt{mini}$-$\mathtt{batch}$  &23.02&15.87&11.27&8.23\\
                                  
    &75.43 \%             &22.18&16.24&12.08&7.46\\    
    %\midrule
    \textbf{ResNet18}~\cite{He2015DeepRL}    
    %&$\mathtt{mini}$-$\mathtt{batch}$  &11.99  &6.82   &4.49   &2.74\\
    &69.00 \%             &12.82  &8.16   &5.19   &2.95\\
    %\midrule
    \textbf{ResNet18-Ch.Prune}~\cite{AMC2019}    
    %&$\mathtt{mini}$-$\mathtt{batch}$  &14.95  &9.46   &5.52   &3.43\\ 
                                 
    &67.62 \%             &11.18  &6.64   &3.99   &2.34\\
    %\midrule
    \textbf{ResNet18-XNOR}~\cite{rastegari2016}       %&$\mathtt{mini}$-$\mathtt{batch}$  &11.54  &6.43   &3.19   &1.33\\
                                   
    &49.10 \%         &7.57   &4.54   &2.19   &0.93\\
    %\midrule
    \textbf{ResNet18-ABC(1$\times$1)}~\cite{lin2017}  %&$\mathtt{mini}$-$\mathtt{batch}$  &10.98&5.00&2.1 7&1.02\\
    &    51.07 \%                &9.11&4.65&2.30&1.13\\
    \textbf{ResNet18-ABC(3$\times$3)}~\cite{lin2017}  %&$\mathtt{mini}$-$\mathtt{batch}$  &12.12&5.61&2.56&1.36\\
    &59.83 \%             &11.33&5.73&2.65&1.43\\
    \bottomrule                                
    \end{tabular}}
  \vspace{1em}
  \caption{Accuracy (Top1) [\%] of CNNs after FGSM adversarial attacks for ImageNet.}
  \label{tab:FGSMImageNet}
\end{table*}

    %STATISTICS MODE:
%/storage/public/Attacks/cifar10/PGD/ResNet20
\begin{table*}[!]
  \centering
  \resizebox{0.6\textwidth}{!}{
  \renewcommand{\arraystretch}{0.7}
  \begin{tabular}{p{4cm}|c|cccc} 
    \toprule
    \textbf{PGD} &$\epsilon$& $i=2$ &$i=3$ &$i=4$ &$i=5$ \\
    \midrule    
    \midrule
    \textbf{Vanilla}~\cite{He2015DeepRL}    &0.1     &81.58 &74.69  &67.52  &59.55\\%81.57 &74.73  &67.6   &59.59\\
    (92.46 \%)                              &0.5     &35.30 &15.61  &6.40   &2.20\\%35.09 &18.63  &6.39   &2.29\\
                                            &1.0     &11.81 &1.92   &0.39   &0.09\\%11.90 &2.00   &0.37   &0.06\\
                                            &2.0     &1.84  &0.09   &0.00   &0.00\\%1.82  &0.09   &0.00   &0.00\\
    \midrule
    %\textbf{KD-CE}~\cite{contrastive_kd}   &0.1      &81.72	&74.24	&66.33	&58.52\\%81.69 &74.36  &66.34  &58.54\\
    %(92.70 \%)                              &0.5     &33.84	&14.12	&5.10	&1.97\\%33.94 &14.27  &5.21   &2.04\\
    %                                        &1.0     &10.54	&1.97	&0.34	&0.06\\%10.61 &1.99   &0.36   &0.05\\
    %                                        &2.0     &1.83	&0.11	&0.01	&0.00\\%1.82  &0.12   &0.01   &0.00\\
    %\midrule
    \textbf{KD-KL}~\cite{contrastive_kd}   &0.1      &85.11	&80.39	&74.66	&69.42\\%81.73 &74.31  &66.28  &58.67\\
    (93.25 \%)                              &0.5     &55.36	&39.40	&27.78	&19.60\\%55.30 &39.53  &27.85  &19.77\\
                                            &1.0     &36.92	&20.06	&10.44	&5.57\\
\bottomrule                                
    \end{tabular}}
    \vspace{1em}
    \caption{Accuracy [\%] of compressed variants of ResNet20 after PGD adversarial attacks for CIFAR-10. 
    %BatchNorm~\cite{bn} layers are in $\mathtt{statistics}$ mode.
    }
  \label{tab:PGDResnet20_eval}
\end{table*}

    %MIMI-BATCH MODE:
%/storage/public/Attacks/cifar10/PGD/ResNet20

\begin{table}[!]
  \centering
  \resizebox{0.7\textwidth}{!}{
  \renewcommand{\arraystretch}{0.9}
  \begin{tabular}{p{3.8cm}|c|cccc} 
    \toprule
    \textbf{PGD}&$\epsilon$& $i=2$ &$i=3$ &$i=4$ &$i=5$ \\
    \midrule
    \midrule
    \textbf{Vanilla}~\cite{He2015DeepRL}    &0.1     &82.27 &76.01  &69.57  &62.29\\%82.30 &76.32  &69.79  &62.35\\
    (92.46 \%)                              &0.5     &37.39 &18.46  &8.79   &4.10\\%37.65 &18.63  &8.91   &3.97\\
                                            &1.0     &14.15 &3.53   &0.85   &0.25\\%14.18 &3.37   &0.84   &0.26\\
                                            &2.0     &3.36  &0.38   &0.05   &0.01\\%3.37  &0.41   &0.05   &0.01\\
    \midrule
    \textbf{KD-CE}~\cite{contrastive_kd}   &0.1     &82.33	&76.61	&69.50	&62.01\\%82.52 &76.53   &69.51  &62.01\\
    (92.70 \%)                              &0.5     &37.14	&17.96	&8.26	&3.66\\%37.10 &18.05  &8.29   &3.88\\
                                            &1.0     &13.54	&3.49	&0.97	&0.22\\
\bottomrule                                
    \end{tabular}}
    \vspace{1em}
    \caption{Accuracy [\%] of compressed variants of ResNet20 after PGD adversarial attacks for CIFAR-10. BatchNorm~\cite{bn} layers are in $\mathtt{mini}$-$\mathtt{batch}$ mode.}
  \label{tab:PGDResnet20_train}
\end{table}
    
    %STATISTICS MODE:
%logs in /storage/public/Attacks/cifar10/PGD/ResNet56
\begin{table*}[!]
  \centering
  \resizebox{0.6\textwidth}{!}{
  \renewcommand{\arraystretch}{0.9}
  \begin{tabular}{p{3.8cm}|c|cccc} 
    \toprule
    \textbf{PGD}&$\epsilon$    & $i=2$ &$i=3$ &$i=4$ &$i=5$ \\
    \midrule    
    \midrule
    \textbf{Vanilla}~\cite{He2015DeepRL}    &0.1     &85.71	&80.44	&74.58	&68.17\\%85.73 &80.33  &74.54  &68.19\\
    (93.88 \%)                              &0.5     &51.89	&33.65	&20.92	&12.19\\%51.93 &33.60  &20.80  &12.32\\
                                            &1.0     &30.53	&13.19	&5.23	&1.91\\%30.38 &12.80  &5.38   &1.99\\
                                            &2.0     &13.92	&3.21	&0.71	&0.13\\%13.95 &3.35   &0.75   &0.17\\
    \midrule
    %\textbf{KD-CE}~\cite{contrastive_kd}   &0.1     &89.99	&89.96	&89.97	&89.92\\
    %(93.68\%)                               &0.5     &73.88	&72.35	&73.13	&72.40\\
    %                                        &1.0     &59.16	&53.40	&55.41	&53.21\\
    %                                        &2.0     &36.28	&25.14	&19.61	&17.43\\
    %\midrule
    \textbf{KD-KL}~\cite{contrastive_kd}&0.1     &90.88 &90.78  &90.77  &90.76\\
    (94.24 \%)                              &0.5     &76.83 &74.23  &73.49  &73.11\\
                                            &1.0     &63.51 &55.81  &62.62  &50.82\\
\bottomrule                                
    \end{tabular}}
    \vspace{1em}
    \caption{Accuracy [\%] of compressed variants of ResNet56 after PGD adversarial attacks for CIFAR-10. 
    %BatchNorm~\cite{bn} layers are in $\mathtt{statistics}$ mode.
    }
  
  \label{tab:PGDResnet56_eval}
\end{table*}
    %MIMI-BATCH MODE:

%logs in /storage/public/Attacks/cifar10/PGD/ResNet56
\begin{table}[!]
  \centering
  
  \resizebox{0.7\textwidth}{!}{
  \renewcommand{\arraystretch}{0.9}
  \begin{tabular}{p{3.8cm}|c|cccc} 
    \toprule
    \textbf{PGD}&$\epsilon$    & $i=2$ &$i=3$ &$i=4$ &$i=5$ \\
    \midrule
    \midrule
    \textbf{Vanilla}~\cite{He2015DeepRL}    &0.1     &85.27	&80.42	&74.76	&68.80\\%85.28 &80.36  &74.71  &68.81\\
    (93.88 \%)                              &0.5     &52.14	&35.12	&22.80	&14.46\\%52.20 &35.20  &22.77  &14.48\\
                                            &1.0     &32.26	&15.47	&7.35	&3.49\\%32.14 &15.23  &7.42   &3.52\\
                                            &2.0     &17.37	&5.53	&1.77	&0.65\\%17.18 &5.71   &1.90   &0.67\\
    
    \midrule
    \textbf{KD-CE}~\cite{contrastive_kd}   &0.1     &89.27	&89.26	&89.30	&89.28\\
    (93.68 \%)                              &0.5     &73.98	&72.34	&73.17	&72.56\\
                                            &1.0     &59.70	&54.54	&57.13	&54.71\\
\bottomrule                                
    \end{tabular}}
    \vspace{1em}
    \caption{Accuracy [\%] of compressed variants of ResNet56 after PGD adversarial attacks for CIFAR-10. BatchNorm~\cite{bn} layers are in $\mathtt{mini}$-$\mathtt{batch}$ mode.}
  \label{tab:PGDResnet56_train}
\end{table}

    %STATISTICS MODE:
\begin{table*}[t]
    \centering
    \resizebox{0.6\textwidth}{!}{
    \renewcommand{\arraystretch}{0.6}
    \begin{tabular}{p{5cm}|c|cccc} 
    \toprule
    \textbf{PGD} &$\epsilon$    & $i=2$ &$i=3$ &$i=4$ &$i=5$ \\
    \midrule    
    \midrule
    \textbf{ResNet50}~\cite{He2015DeepRL}    &0.1     &25.77&16.07 & 9.83 & 5.91 \\  
    (75.43 \%)                      &0.5     &3.35&0.94&0.43& 0.27\\
    \midrule
    \textbf{ResNet18}~\cite{He2015DeepRL}    &0.1     &17.86 &10.32  &5.58   &3.11\\
    (69.00 \%)                      &0.5     &1.33  &0.17   &0.04   &0.01\\
    \midrule
    \textbf{ResNet18-Ch.Prune}~\cite{AMC2019}    &0.1     &17.02 &10.23  &5.92   &3.50\\
    (67.62 \%)                      &0.5     &1.40  &0.27   &0.06   &0.02\\
    \midrule
    \textbf{ResNet18-XNOR}~\cite{rastegari2016}       &0.1     &13.16 &11.46  &10.06  &8.84\\
    (49.10 \%)                      &0.5     &5.67  &3.07   &1.57   &0.78\\
    \midrule                                       
    \textbf{ResNet18-ABC(1$\times$1)}~\cite{lin2017}  &0.1     &18.35&16.22&14.20&12.37\\               
    (51.91)                         &0.5     &7.60&3.64&1.75&0.82\\
    \midrule                                       
    \textbf{ResNet18-ABC(3$\times$3)}~\cite{lin2017}  &0.1     &23.90&20.81&17.80&15.07\\                
    (59.83)                         &0.5     &8.31&3.70&1.59&0.66\\
    \bottomrule                                
    \end{tabular}}
    \vspace{1em}
    \caption{Accuracy [\%] of CNNs after PGD adversarial attacks for ImageNet.}
    
    \label{tab:PGDImageNet_eval}
\end{table*}
    %MIMI-BATCH MODE:
\begin{table}[!]
  \centering
  \resizebox{0.7\textwidth}{!}{
  \begin{tabular}{p{3.7cm}|c|cccc} 
    \toprule
    \textbf{PGD}                    &$\epsilon$    & $i=2$ &$i=3$ &$i=4$ &$i=5$ \\
    \midrule  
    \midrule
    \textbf{ResNet50}~\cite{He2015DeepRL}    &0.1     &40.38& 33.31& 26.87& 21.19\\
    (75.43 \%)                      &0.5     &12.10&4.38&1.64 & 0.75 \\                      
    \midrule
    \textbf{ResNet18}~\cite{He2015DeepRL}    &0.1     &26.13 &20.59  &15.60  &11.65\\
    (69.00 \%)                      &0.5     &5.68  &1.54   &0.40   &0.13\\
    \midrule
    \textbf{Ch.Prune}~\cite{AMC2019}    &0.1     &27.27 &23.91  &20.69  &17.75\\
    (67.62 \%)                      &0.5     &11.72 &6.13   &3.19   &1.67\\
    \midrule
    \textbf{XNOR}~\cite{rastegari2016}       &0.1     &23.57 &21.95  &20.31  &18.99\\
    (47.24 \%)                      &0.5     &15.08  &10.45   &6.79   &4.42\\
    \midrule                                       
    \textbf{ABC(1$\times$1)}~\cite{lin2017}  &0.1     &25.78&23.71&23.71&19.50\\
    (51.91 \%)                      &0.5     &13.28 &7.41 &3.69&1.78 \\
    \midrule                                       
    \textbf{ABC(3$\times$3)}~\cite{lin2017}  &0.1     &30.55&27.73&24.48&21.41\\
    (59.83 \%)                      &0.5     &13.62 &6.54 &2.87&1.16 \\
    \bottomrule                                
    \end{tabular}}
    \vspace{1em}
    \caption{Accuracy [\%] of CNNs after PGD adversarial attacks for ImageNet. BatchNorm~\cite{bn} layers are in $\mathtt{mini}$-$\mathtt{batch}$ mode.}
  \label{tab:PGDImageNet_train}
\end{table}

    %STATISTIC MODE
\begin{table*}[!]
  \centering
  
  \resizebox{0.7\textwidth}{!}{
  \renewcommand{\arraystretch}{0.9}
  \begin{tabular}{p{3.8cm}|c|cccc} 
    \toprule
    \multirow{2}{*}{\textbf{C\&W}}&\multirow{2}{*}{$c$}    & $i=1$ &$i=10$ &$i=20$ &$i=50$ \\
            &&\textbf{OA/TA}    & \textbf{OA/TA}    & \textbf{OA/TA}    & \textbf{OA/TA} \\
    \midrule
    \midrule
    \textbf{Vanilla}~\cite{He2015DeepRL}    &0.01    &22.70/2.21	&2.80/76.14	    &0.40/96.31	    &0.20/99.00\\
    (92.46 \%)                              &0.1     &13.25/1.76	&5.74/53.72	    &1.82/80.87	    &0.32/96.59\\
                                            &1.0     &11.67/0.50	&8.02/29.05	    &4.58/56.31	    &1.28/85.65\\
                                            &5.0     &10.80/0.32	&9.38/13.96	    &6.41/35.24	    &3.10/69.74\\
                                            &10.0    &10.43/0.20	&9.88/6.05	    &7.84/20.65	    &4.56/53.73\\
    \midrule
    %\textbf{KD-CE }~\cite{contrastive_kd}  &0.01    &20.17/1.23	&5.01/60.22	    &1.33/90.50	    &0.32/98.19\\
    %(92.70 \%)                              &0.1     &11.18/0.24	&7.18/28.99	    &3.67/66.36	    &0.59/94.50\\
    %                                        &1.0     &10.63/0.10	&9.05/13.00	    &6.22/38.71	    &2.13/79.31\\
    %                                        &5.0     &10.05/0.06	&9.94/6.29	    &8.00/21.39	    &3.87/60.08\\
    %                                        &10.0    &9.74/0.03	    &9.77/3.21	    &8.64/11.30	    &5.53/42.74\\
    %\midrule
    \textbf{KD-KL}~\cite{contrastive_kd}   &0.01    &23.39/0.59	&9.72/28.91	    &6.27/55.99	    &1.81/86.08\\
\end{tabular}}
    \vspace{1em}
    \caption{Accuracy [\%] of compressed variants of ResNet20 after C\&W adversarial attacks for CIFAR-10.}
  \label{tab:CWResnet20_eval}
\end{table*}
    %MINI-BATCH MODE 
\begin{table}[ht]
  \centering
  
  \resizebox{0.7\textwidth}{!}{
  \renewcommand{\arraystretch}{0.9}
  \begin{tabular}{p{3.8cm}|c|cccc} 
    \toprule
    \multirow{2}{*}{\textbf{C\&W}}&\multirow{2}{*}{$c$}    & $i=1$ &$i=10$ &$i=20$ &$i=50$ \\
            &&\textbf{OA/TA}    & \textbf{OA/TA}    & \textbf{OA/TA}    & \textbf{OA/TA} \\
    \midrule
    \midrule
    \textbf{Vanilla}~\cite{He2015DeepRL}    &0.01    &30.73/7.31	&19.89/12.98	&18.14/14.31	&20.02/13.02\\
    (92.46 \%)                              &0.1     &16.36/8.53	&12.80/11.98	&11.80/13.64	&11.61/13.70\\
                                            &1.0     &13.82/9.69	&11.95/12.91	&11.30/14.57	&11.08/15.40\\
                                            &5.0     &11.94/9.85	&10.46/12.81	&9.93/15.11	    &9.80/16.82\\
                                            &10.0    &11.69/10.97	&10.62/12.81	&9.99/14.94	    &9.56/17.66\\
    \midrule
    \textbf{KD-CE }~\cite{contrastive_kd}  &0.01    &32.36/8.03	&22.34/14.19	&20.65/15.90	&22.21/14.70\\
\end{tabular}}
    \vspace{1em}
    \caption{Accuracy [\%] of compressed variants of ResNet20 after C\&W adversarial attacks for CIFAR-10. BatchNorm~\cite{bn} layers are in $\mathtt{mini}$-$\mathtt{batch}$ mode.}
  \label{tab:CWResnet20_train}
\end{table}

    %STATISTICS MODE 
\begin{table*}[!h]
  \centering
  \resizebox{0.7\textwidth}{!}{
  \begin{tabular}{p{3.8cm}|c|cccc} 
    \toprule
    \multirow{2}{*}{\textbf{C\&W}}&\multirow{2}{*}{$c$}    & $i=1$ &$i=10$ &$i=20$ &$i=50$ \\
            &&\textbf{OA/TA}    & \textbf{OA/TA}    & \textbf{OA/TA}    & \textbf{OA/TA} \\
    \midrule
    \midrule
    \textbf{Vanilla}~\cite{He2015DeepRL}    &0.01    &21.71/0.77	&7.90/43.84	    &2.98/76.34	    &0.79/95.08\\
    (93.88 \%)                              &0.1     &11.01/0.13	&8.22/26.29	    &5.05/54.25	    &1.97/82.49\\
                                            &1.0     &10.35/0.08	&9.17/10.20	    &7.71/26.41	    &4.64/55.07\\
                                            &5.0     &10.49/0.00	&10.36/4.37	    &9.25/12.48	    &6.72/34.38\\
                                            &10.0    &10.37/0.01	&10.09/1.44	    &9.84/5.74	    &8.30/20.77\\
    \midrule
    %\textbf{KD-CE}~\cite{contrastive_kd}   &0.01    &25.43/0.99	&14.36/27.03	&12.11/57.54	&10.41/87.60\\
    %(93.68 \%)                              &0.1     &14.22/0.00	&11.91/8.31	    &11.26/29.44	&10.12/66.67\\
    %                                        &1.0     &12.20/0.00	&10.79/2.17	    &11.14/10.52    &10.38/39.44\\
    %                                        &5.0     &10.82/0.00	&10.59/0.50	    &10.32/3.95	    &10.02/22.78\\
    %                                        &10.0    &10.39/0.00	&10.53/0.18	    &10.57/1.56 	&10.01/12.39\\
    %\midrule
    \textbf{KD-KL}~\cite{contrastive_kd}&0.01    &22.66/0.68    &14.02/10.28    &13.15/22.84    &11.73/50.81\\
\end{tabular}}
    \vspace{1em}
    \caption{Accuracy [\%] of compressed variants of ResNet56 after C\&W adversarial attacks for CIFAR-10. 
    %BatchNorm~\cite{bn} layers are in $\mathtt{statistics}$ mode.
    }
  \label{tab:CWResnet56_eval}
\end{table*}
    %MINI-BATCH MODE 
\begin{table}[!h]
  \centering
  
  \resizebox{0.7\textwidth}{!}{
  \begin{tabular}{p{3.8cm}|c|cccc} 
    \toprule
    \multirow{2}{*}{\textbf{C\&W}}&\multirow{2}{*}{$c$}    & $i=1$ &$i=10$ &$i=20$ &$i=50$ \\
            &&\textbf{OA/TA}    & \textbf{OA/TA}    & \textbf{OA/TA}    & \textbf{OA/TA} \\
    \midrule
    \midrule
    \textbf{Vanilla}~\cite{He2015DeepRL}    &0.01    &30.24/7.45	&21.66/13.80	&19.20/15.60	&20.55/14.27\\
    (93.88 \%)                              &0.1     &16.02/9.28	&14.01/12.59	&12.48/14.91	&12.04/16.15\\
                                            &1.0     &14.06/9.22	&12.73/11.32	&11.65/13.85	&10.61/16.56\\
                                            &5.0     &12.70/9.14	&11.53/10.50	&10.77/12.15	&10.02/15.01\\
                                            &10.0    &10.97/9.33	&10.76/10.17	&10.11/11.13	&9.35/14.00\\
    \midrule
    \textbf{KD-CE}~\cite{contrastive_kd}   &0.01    &32.20/10.06	&24.63/14.61	&22.01/17.05	&22.99/16.34\\
\end{tabular}}
    \vspace{1em}
    \caption{Accuracy [\%] of compressed variants of ResNet56 after C\&W adversarial attacks for CIFAR-10. BatchNorm~\cite{bn} layers are in $\mathtt{mini}$-$\mathtt{batch}$ mode.}
  \label{tab:CWResnet56_train}
\end{table}

    \begin{table*}[!]
  \centering
  \resizebox{0.6\textwidth}{!}{
  \renewcommand{\arraystretch}{0.9}
  \begin{tabular}{l|l|cccc} 
    \toprule
    \textbf{DeepFool} &\textbf{Nat.Acc} &$i=1$  &$i=5$ &$i=10$ &$i=20$\\
    \midrule
    \midrule
    \textbf{Vanilla}~\cite{He2015DeepRL}    
    %&$\mathtt{mini}$-$\mathtt{batch}$   &38.09	&6.90	&6.27	&6.08\\%38.14  &7.25   &6.50   &6.23 \\
    &92.46 \%             &37.75	&5.15	&5.11	&5.09\\%37.75  &5.15   &5.12   &5.09 \\
    %\midrule
    %\textbf{KD-CE}~\cite{contrastive_kd}   &$\mathtt{mini}$-$\mathtt{batch}$   &39.00	&7.50	&6.89	&6.52\\%39.23  &7.91   &7.17   &6.86 \\
    %(92.70 \%)                              &$\mathtt{statistics}$              &38.10	&4.78	&4.75	&4.80\\%38.10  &4.78   &4.75   &4.79 \\
    %\midrule
    \textbf{KD-KL}~\cite{contrastive_kd}
    %&$\mathtt{mini}$-$\mathtt{batch}$   &60.42	&11.84	&9.36	&8.73\\%59.88  &10.99  &9.02   &8.23 \\
    &93.25 \%              &61.28	&6.35	&4.64	&4.53\\%60.72  &6.24   &4.79   &4.72 \\
    %\midrule
    \textbf{Ch.Prune}~\cite{AMC2019}   
    %&$\mathtt{mini}$-$\mathtt{batch}$   &33.01	&7.59	&7.31	&7.14\\%33.06  &7.59   &7.31   &7.17 \\
    &89.76 \%           &33.18	&6.79	&6.74	&6.66\\%33.18  &6.79   &6.74   &6.68 \\
    %\midrule
    %\textbf{F.Prune}~\cite{AMC2019}    
    %&$\mathtt{mini}$-$\mathtt{batch}$   &34.00	&8.54	&7.90	&7.50\\
    %&(89.91 \%)            &32.23	&6.39	&6.27	&6.25\\
    %\midrule
    \textbf{K.Prune}~\cite{AMC2019}    
    %&$\mathtt{mini}$-$\mathtt{batch}$   &34.56  &7.57   &7.13   &7.06\\
    &90.73 \%              &34.50  &6.19   &6.16   &6.15\\
    %\midrule
    \textbf{W.Prune}~\cite{AMC2019}    
    %&$\mathtt{mini}$-$\mathtt{batch}$   &47.94	&25.12	&23.86	&22.70\\
    &91.99 \%             &32.58	&7.75	&7.75	&7.73\\
    %\midrule
    \textbf{XNOR}~\cite{rastegari2016}      
    %&$\mathtt{mini}$-$\mathtt{batch}$   &63.75	&20.66	&14.23	&12.16\\%64.31  &20.90  &14.50  &12.07 \\
    &82.71 \%              &62.59	&15.51	&11.38	&10.40\\
\bottomrule                                
    \end{tabular}}
    \vspace{1em}
    \caption{Accuracy (Top1) [\%] of ResNet20 like CNNs after DeepFool adversarial attacks for CIFAR-10.}
  \label{tab:DeepFoolResnet20_full}
\end{table*}

    \begin{table*}[ht]
  \centering
  \resizebox{0.6\textwidth}{!}{
  \renewcommand{\arraystretch}{0.9}
  \begin{tabular}{l|l|cccc} 
    \toprule
    \textbf{DeepFool} &\textbf{Nat.Acc} &$i=1$  &$i=5$ &$i=10$ &$i=20$\\
    \midrule
    \midrule
    \textbf{Vanilla}~\cite{He2015DeepRL}    
    %&$\mathtt{mini}$-$\mathtt{batch}$   &56.49	&11.08	&9.31	&8.40\\%56.54  &11.08  &9.33   &8.36 \\ 
    &93.88 \%              &58.27	&4.59	&4.18	&4.09\\%58.27  &4.59   &4.18   &4.11 \\  
    %\midrule
    %\textbf{KD-CE}~\cite{contrastive_kd}   &$\mathtt{mini}$-$\mathtt{batch}$   &59.48	&10.67	&7.93	&7.09\\
    %(93.68 \%)                              &$\mathtt{statistics}$              %&60.70	&5.69	&4.27	&4.29\\
    %\midrule
    \textbf{KD-KL}~\cite{contrastive_kd}
    %&$\mathtt{mini}$-$\mathtt{batch}$   &65.64	&11.36	&6.27	&5.43\\
    &94.24 \%             &67.03	&8.46	&4.16	&4.01\\
    %\midrule
    \textbf{Ch.Prune}~\cite{AMC2019}   
    %&$\mathtt{mini}$-$\mathtt{batch}$   &55.64	&10.49	&8.60	&7.95\\%75.43  &39.63  &22.50  &15.43 \\
    &(92.86 \%)               &57.50	&5.60	&4.82	&4.78\\%57.50  &5.60   &4.82   &4.78 \\
    %\midrule
    %\textbf{F.Prune}~\cite{AMC2019}    
    %&$\mathtt{mini}$-$\mathtt{batch}$   &55.35	&9.99	&8.20	&7.66\\
    %&(93.09 \%)            &57.68	&5.31	&4.72	&4.66\\
    %\midrule
    \textbf{K.Prune}~\cite{AMC2019}    
    %&$\mathtt{mini}$-$\mathtt{batch}$   &59.48  &10.75  &7.98   &7.15\\
    &93.04 \%               &60.70  &5.71   &4.29   &4.29\\
    %\midrule
    \textbf{W.Prune}~\cite{AMC2019}    
    %&$\mathtt{mini}$-$\mathtt{batch}$   &55.77  &10.53  &9.35   &8.56\\
    &(93.54 \%)            &57.78  &4.62   &4.36   &4.28\\
    %\midrule    
    \textbf{XNOR}~\cite{rastegari2016}      
    %&$\mathtt{mini}$-$\mathtt{batch}$   &75.22  &38.94  &22.62  &15.27\\%55.15  &14.06  &11.02  &9.81 \\
    &(83.24 \%)                 &74.66  &34.14  &16.94  &11.79\\
\bottomrule                                
    \end{tabular}}
    \vspace{1em}
    \caption{Accuracy (Top1) [\%] of ResNet56 like CNNs after DeepFool adversarial attacks for CIFAR-10.}
  \label{tab:DeepFoolResnet56_full}
\end{table*}

    % -------------------------------------------------- EVAL MODE ---------------------------------------------------------- %

\begin{table*}[!]
\centering
\resizebox{0.7\textwidth}{!}{
\renewcommand{\arraystretch}{1.0}
\begin{tabular}{l|c|cccc}
\toprule
\multirow{2}{*}{\textbf{LocalSearch}}  & \multirow{2}{*}{$\epsilon$}        & $i=50$            & $i=100$           & $i=150$           & $i=200$              \\ 
                                        &                                   & \textbf{OA/TA}    & \textbf{OA/TA}    & \textbf{OA/TA}    & \textbf{OA/TA}       \\
\midrule
\midrule
\textbf{Vanilla}~\cite{He2015DeepRL}    & 8.0                               & 63.48/12.71       & 56.61/14.06       & 50.06/15.59       & 44.55/17.15          \\
(92.46 \%)                              & 16.0                              & 46.70/30.55       & 32.26/43.19       & 24.27/51.62       & 18.99/58.90          \\    
                                        & 32.0                              & 11.58/81.73       &  1.79/97.13       &  0.52/99.18       &  0.28/99.59           \\
\midrule                        
\textbf{KD-CE}~\cite{contrastive_kd}   & 8.0                               & 58.89/17.95       & 49.15/20.46       & 42.90/20.79       & 38.08/21.09          \\
(92.70 \%)                              & 16.0                              & 43.37/35.05       & 27.49/50.61       & 20.44/58.25       & 15.74/63.02          \\ 
                                        & 32.0                              & 10.40/84.36       &  1.38/18.08       &  0.49/99.33       &  0.22/99.59           \\
\end{tabular}}
\vspace{2ex}
\caption{Accuracy (Top1) [\%] of ResNet20 like CNN after LocalSearch adversarial attacks for CIFAR-10.}
\label{tab:LocalSearchResNet20_eval}
\end{table*}
    
% ----------------------------------------- %
%             Local Search                  %
% ----------------------------------------- %

% ----------------------------------------- %
%               ResNet 20                   %
% ----------------------------------------- %

% ----------------------------------------- %
%             DS: Cifar-10                  %
% ----------------------------------------- %

% -------------------------------------------------- TRAIN MODE ---------------------------------------------------------- %

\begin{table}[ht]
\centering
\resizebox{0.7\textwidth}{!}{
\renewcommand{\arraystretch}{0.9}
\begin{tabular}{l|c|cccc}
\toprule
\multirow{2}{*}{\textbf{LocalSearch}}  & \multirow{2}{*}{$\epsilon$}        & $i=50$            & $i=100$           & $i=150$           & $i=200$              \\ 
                                        &                                   & \textbf{OA/TA}    & \textbf{OA/TA}    & \textbf{OA/TA}    & \textbf{OA/TA}       \\ 
\midrule
\midrule
\textbf{Vanilla}~\cite{He2015DeepRL}    & 8.0                           & 69.78/6.23        & 64.86/6.67        & 60.36/5.95        & 58.21/7.02           \\
(92.46 \%)                              & 16.0                          & 59.57/11.32       & 51.64/12.22       & 47.60/12.77       & 44.80/13.68          \\
                                        & 32.0                          & 43.62/19.81       & 41.25/18.38       & 40.23/17.19       & 39.45/18.14          \\
\midrule                        
\textbf{KD-CE}~\cite{contrastive_kd}   & 8.0                           & 68.46/6.97        & 63.34/6.78        & 59.27/6.67        & 57.48/6.77           \\
(92.70 \%)                              & 16.0                          & 58.76/35.05       & 51.66/11.83       & 47.40/11.94       & 44.93/12.36          \\ 
                                        & 32.0                          & 44.83/19.78       & 41.09/18.08       & 40.69/17.18       & 40.49/16.95          \\
\midrule                        
\textbf{KD-KL}~\cite{contrastive_kd}   & 8.0                           & 71.01/6.869       & 56.80/13.56       & 51.61/14.38       & 46.95/14.64          \\
(93.25 \%)                              & 16.0                          & 62.54/11.19       & 55.49/13.25       & 51.60/14.32       & 47.71/15.46          \\ 
                                        & 32.0                          & 47.62/20.99       & 43.43/21.38       & 42.47/20.83       & 41.39/20.62          \\
\midrule        
\textbf{Ch.Prune}~\cite{AMC2019}   & 8.0                           & 67.08/3.64        & 62.28/3.80        & 57.91/4.27        & 54.31/4.78           \\
(89.76 \%)                              & 16.0                          & 56.08/10.49       & 49.49/11.84       & 46.15/12.10       & 43.68/12.36          \\ 
                                        & 32.0                          & 44.64/20.01       & 40.52/21.88       & 39.33/21.68       & 37.93/21.53          \\
\midrule
\textbf{F.Prune}~\cite{AMC2019}    & 8.0                           & 67.70/3.55        & 62.70/3.64        & 58.25/4.31        & 55.36/4.86           \\
(89.92 \%)                              & 16.0                          & 58.07/9.57        & 52.26/10.45       & 48.09/11.20       & 45.34/11.68          \\ 
                                        & 32.0                          & 46.11/19.08       & 41.75/20.70       & 39.79/20.63       & 38.39/20.86          \\
\midrule        
\textbf{K.Prune}~\cite{AMC2019}    & 8.0                           & 69.88/3.99        & 64.15/4.02        & 60.36/4.44        & 56.30/4.95           \\
(90.74 \%)                              & 16.0                          & 58.29/9.93        & 51.81/11.72       & 47.68/12.43       & 45.07/12.82          \\ 
                                        & 32.0                          & 46.63/19.04       & 43.27/20.57       & 40.67/20.55       & 39.78/20.73          \\
\midrule
\textbf{W.Prune}~\cite{AMC2019}    & 8.0                           & 71.14/3.25        & 66.23/3.36        & 61.12/4.15        & 57.49/4.60           \\
(91.98 \%)                              & 16.0                          & 59.57/9.79        & 52.28/11.51       & 48.58/11.84       & 45.42/12.19          \\ 
                                        & 32.0                          & 47.92/18.76       & 43.14/20.20       & 41.09/21.03       & 39.77/21.00          \\
\midrule                        
\textbf{XNOR}~\cite{rastegari2016}      & 8.0                           & 68.81/5.00        & 64.28/5.24        & 60.20/6.02        & 57.81/6.58           \\
(82.71 \%)                              & 16.0                          & 68.34/5.36        & 62.18/6.31        & 58.39/6.85        & 56.17/7.33           \\ 
                                        & 32.0                          & 62.29/8.95        & 54.63/10.31       & 50.97/11.30       & 49.71/11.85          \\
\midrule        
\textbf{ABC(1$\times$1)}~\cite{lin2017} & 8.0                           & 68.91/5.05        & 62.76/5.76        & 59.12/6.34        & 56.12/6.62           \\
(83.42 \%)                              & 16.0                          & 67.48/5.53        & 60.68/6.76        & 56.63/7.38        & 54.62/7.67           \\ 
                                        & 32.0                          & 62.60/8.17        & 54.79/10.07       & 50.85/10.57       & 48.57/11.03          \\
\midrule                                        
\textbf{ABC(3$\times$3)}~\cite{lin2017} & 8.0                           & 72.85/4.44        & 68.20/4.45        & 64.94/4.69        & 62.37/5.45           \\
(88.94 \%)                              & 16.0                          & 72.09/4.66        & 65.99/5.34        & 62.12/5.95        & 58.86/6.38           \\ 
                                        & 32.0                          & 65.53/6.52        & 57.57/8.20        & 54.13/8.70        & 51.26/9.19           \\
\midrule                                        
\textbf{ABC(5$\times$5)}~\cite{lin2017} & 8.0                           & 74.54/2.48        & 69.94/2.66        & 66.28/3.02        & 63.22/3.20           \\
(90.64 \%)                              & 16.0                          & 73.05/3.55        & 66.81/4.09        & 62.09/4.54        & 60.00/4.32           \\ 
                                        & 32.0                          & 64.83/7.38        & 56.75/9.29        & 52.74/10.03       & 51.47/9.97           \\
\midrule                                        
\textbf{ABC(7$\times$7)}~\cite{lin2017} & 8.0                           & 75.70/2.75        & 71.53/2.61        & 67.28/2.98        & 63.99/2.96           \\
(91.34 \%)                              & 16.0                          & 73.00/3.89        & 66.55/4.42        & 61.86/4.81        & 60.59/4.57           \\ 
                                        & 32.0                          & 62.67/8.41        & 55.12/10.75       & 51.67/11.47       & 49.96/11.37          \\
\bottomrule
\multicolumn{6}{r}{OA/TA = Accuracy to original label / Accuracy to target label.}\\                                        
\end{tabular}}
\vspace{2ex}
\caption{Accuracy (Top1) [\%] of ResNet20 like CNN after LocalSearch adversarial attacks for CIFAR-10. BatchNorm~\cite{bn} layers are in $\mathtt{mini}$-$\mathtt{batch}$ mode.}
\label{tab:LocalSearchResNet20_train}
\end{table}

% -------------------------------------------------- EVAL MODE ---------------------------------------------------------- %

\begin{table*}[ht]
\centering
\resizebox{0.7\textwidth}{!}{
\renewcommand{\arraystretch}{0.9}
\begin{tabular}{l|c|cccc}
\toprule
\multirow{2}{*}{\textbf{LocalSearch}}  & \multirow{2}{*}{$\epsilon$}        & $i=50$            & $i=100$           & $i=150$           & $i=200$              \\ 
                                        &                                   & \textbf{OA/TA}    & \textbf{OA/TA}    & \textbf{OA/TA}    & \textbf{OA/TA}       \\
\midrule
\midrule
\textbf{Vanilla}\cite{He2015DeepRL}     & 8.0           & 71.41/2.68             & 62.93/2.02            & 56.14/1.63            & 49.89/1.47           \\
(93.88 \%)                              & 16.0          & 54.94/18.49            & 39.88/25.33           & 31.70/28.50           & 26.27/30.97          \\    
                                        & 32.0          & 22.82/64.36            & 8.27/85.10            & 4.17/91.68            & 2.67/94.62           \\
\midrule                        
\textbf{KD-CE}~\cite{contrastive_kd}   & 8.0           & 67.39/12.89            & 58.73/14.40           & 52.20/14.24           & 46.42/14.10          \\
(93.68 \%)                                                       & 16.0          & 52.89/26.06            & 37.23/38.77           & 28.62/44.56           & 22.73/48.25          \\ 
                                        & 32.0          & 22.63/66.31            &  5.30/91.33           &  1.75/97.12           &  0.77/98.79          \\
\midrule                        
\textbf{KD-KL}~\cite{contrastive_kd}   & 8.0           & 68.43/13.95            & 59.66/17.78           & 52.45/20.02           & 46.34/22.14          \\
(94.24 \%)                                                         & 16.0          & 56.48/23.65            & 39.83/37.43           & 29.48/46.83           & 22.38/54.07          \\ 
                                        & 32.0          & 20.95/54.27            & 10.61/81.64           &  4.34/92.12           &  2.20/96.19          \\
\midrule        
\textbf{Ch.Prune} \cite{AMC2019}   & 8.0           & 68.57/2.79             & 59.32/1.86            & 52.81/1.43            & 47.07/1.27           \\
(92.93 \%)                                                         & 16.0          & 53.65/19.21            & 39.34/25.34           & 30.17/27.96           & 24.56/29.85          \\ 
                                        & 32.0          & 23.99/62.58            &  9.57/82.36           &  5.06/90.00           &  2.98/93.35          \\
\midrule                                        
%\textbf{F.Prune} \cite{AMC2019}    & 8.0           & 67.88/3.31             & 57.84/2.34            & 50.13/2.16            & 43.89/1.97           \\
%(93.08 \%)                              & 16.0          & 53.18/20.14            & 37.95/27.75           & 28.66/30.94           & 23.49/32.40          \\ 
%                                        & 32.0          & 21.93/65.52            &  7.64/85.59           &  3.90/91.83           &  2.48/94.60          \\
%\midrule                                        
\textbf{K.Prune} \cite{AMC2019}    & 8.0           & 65.06/11.39            & 56.78/12.74           & 50.43/14.25           & 44.66/16.28          \\
(93.03 \%)                              & 16.0          & 49.20/25.88            & 33.73/39.35           & 24.27/49.04           & 19.04/55.55          \\ 
                                        & 32.0          & 16.56/71.86            &  4.03/92.89           &  1.45/97.54           &  0.77/98.80          \\
\midrule                                        
\textbf{W.Prune} \cite{AMC2019}    & 8.0           & 67.26/10.09            & 58.88/10.46           & 53.27/10.83           & 48.34/11.60          \\
(93.54 \%)                              & 16.0          & 52.24/23.40            & 36.00/34.70           & 28.12/41.25           & 22.25/46.79          \\ 
                                        & 32.0          & 19.21/68.06            &  5.30/90.27           &  2.12/96.14           &  0.92/98.24          \\
\midrule                        
\textbf{XNOR}\cite{rastegari2016}       & 8.0           & 64.88/4.61             & 56.11/3.94            & 49.16/4.14            & 44.42/3.30           \\
(83.24 \%)                              & 16.0          & 63.70/5.32             & 53.50/6.16            & 47.22/6.00            & 42.52/5.88           \\ 
                                        & 32.0          & 56.39/12.42            & 44.30/16.97           & 38.19/18.40           & 33.19/18.26          \\
\midrule        
\textbf{ABC(1$\times$1)}\cite{lin2017}  & 8.0           & 64.73/5.09             & 50.24/3.68            & 38.60/2.14            & 29.83/1.01           \\
(86.29 \%)                              & 16.0          & 62.73/6.42             & 48.23/5.81            & 36.27/4.09            & 27.71/2.26           \\ 
                                        & 32.0          & 55.97/12.92            & 39.33/14.22           & 29.35/10.97           & 22.79/7.24           \\
\midrule                                        
\textbf{ABC(3$\times$3)}\cite{lin2017}  & 8.0           & 73.79/3.58             & 62.86/3.06            & 53.35/2.52            & 45.80/1.40           \\
(92.48 \%)                              & 16.0          & 71.45/4.99             & 58.32/5.71            & 48.95/4.84            & 41.92/3.45           \\ 
                                        & 32.0          & 61.34/13.28            & 43.27/19.33           & 34.05/20.11           & 28.52/17.18          \\
\midrule                                        
\textbf{ABC(5$\times$5)}\cite{lin2017}  & 8.0           & 75.16/2.46             & 64.54/2.25            & 56.75/1.53            & 49.92/1.04           \\
(92.82 \%)                              & 16.0          & 71.78/4.52             & 60.32/5.32            & 51.47/4.23            & 44.45/3.21           \\ 
                                        & 32.0          & 60.77/14.02            & 44.40/20.55           & 35.42/20.55           & 30.88/18.74          \\
\bottomrule                                        
\multicolumn{6}{r}{OA/TA = Accuracy to original label / Accuracy to target label.}\\ 
\end{tabular}}
\vspace{2ex}
\caption{Accuracy (Top1) [\%] of ResNet56 like CNN after LocalSearch adversarial attacks for CIFAR-10. 
%BatchNorm~\cite{bn} layers are in $\mathtt{statistics}$ mode.
}
\label{tab:LocalSearchResNet56_eval}
\end{table*}

% ----------------------------------------- %
%             Local Search                  %
% ----------------------------------------- %

% ----------------------------------------- %
%              ResNet 56                    %
% ----------------------------------------- %

% ----------------------------------------- %
%              DS: Cifar-10                 %
% ----------------------------------------- %

% -------------------------------------------------- TRAIN MODE ---------------------------------------------------------- %

\begin{table}[ht]
\centering
\resizebox{0.7\textwidth}{!}{
\renewcommand{\arraystretch}{0.9}
\begin{tabular}{l|c|cccc}
\toprule
\multirow{2}{*}{\textbf{LocalSearch}}   & \multirow{2}{*}{$\epsilon$}           & $i=50$                & $i=100$               & $i=150$           & $i=200$              \\ 
                                        &                                       & \textbf{OA/TA}        & \textbf{OA/TA}        & \textbf{OA/TA}    & \textbf{OA/TA}       \\
\midrule
\midrule
\textbf{Vanilla}\cite{He2015DeepRL}     & 8.0           & 73.96/2.99             & 68.56/3.07            & 64.76/3.20            & 61.27/3.47           \\
(93.88 \%)                              & 16.0          & 62.45/9.52             & 55.05/10.91           & 51.09/11.81           & 47.50/12.96          \\
                                        & 32.0          & 44.25/19.75            & 39.08/22.40           & 37.41/23.00           & 36.21/23.27          \\
\midrule                        
\textbf{KD-CE}~\cite{contrastive_kd}   & 8.0           & 72.95/6.93             & 68.69/6.87            & 65.12/6.85            & 60.87/7.45           \\
(93.68 \%)                              & 16.0          & 62.75/11.99            & 55.98/13.64           & 51.37/14.27           & 47.76/14.44          \\ 
                                        & 32.0          & 45.08/22.18            & 39.05/21.78           & 37.52/21.34           & 37.02/20.48          \\
\midrule                        
\textbf{KD-KL}~\cite{contrastive_kd}   & 8.0           & 74.40/5.97             & 69.82/6.26            & 65.80/6.95            & 62.58/7.22           \\
(94.24 \%)                              & 16.0          & 64.90/11.22            & 57.49/13.16           & 52.68/13.99           & 49.24/14.60          \\ 
                                        & 32.0          & 48.25/20.95            & 41.67/21.62           & 40.29/20.79           & 39.71/20.93          \\
\midrule        
\textbf{Ch.Prune} \cite{AMC2019}   & 8.0           & 72.11/2.78             & 66.29/3.00            & 62.28/3.25            & 58.73/3.74           \\
(92.93 \%)                              & 16.0          & 60.35/9.48             & 53.31/10.97           & 48.97/11.87           & 45.83/12.30          \\ 
                                        & 32.0          & 42.79/18.74            & 37.60/21.29           & 35.44/21.96           & 34.96/21.80          \\
\midrule                                        
\textbf{F.Prune} \cite{AMC2019}    & 8.0           & 72.42/3.07             & 57.84/2.34            & 50.13/2.16            & 43.89/1.97           \\
(93.08 \%)                              & 16.0          & 61.89/9.72             & 54.50/11.28           & 49.45/12.34           & 46.27/13.04          \\ 
                                        & 32.0          & 43.40/19.28            & 38.07/22.29           & 36.12/22.82           & 35.26/22.59          \\
\midrule                                        
\textbf{K.Prune} \cite{AMC2019}    & 8.0           & 70.33/6.59             & 65.56/6.56            & 61.59/6.96            & 58.22/7.20           \\
(93.03 \%)                              & 16.0          & 58.79/11.16            & 51.68/12.71           & 47.63/13.69           & 44.82/14.19          \\ 
                                        & 32.0          & 42.39/20.27            & 38.40/20.92           & 36.40/19.99           & 36.27/19.74          \\
\midrule                                        
\textbf{W.Prune} \cite{AMC2019}    & 8.0           & 71.56/6.25             & 66.56/6.29            & 62.88/6.35            & 60.44/6.46           \\
(93.54 \%)                              & 16.0          & 60.50/10.90            & 53.62/12.60           & 50.00/12.77           & 46.08/13.21          \\ 
                                        & 32.0          & 42.07/21.49            & 37.58/20.91           & 36.80/20.39           & 36.26/20.25          \\
\midrule
\textbf{XNOR}\cite{rastegari2016}       & 8.0           & 69.49/4.12             & 64.85/4.44            & 61.04/4.68            & 59.01/4.96           \\
(83.24 \%)                              & 16.0          & 68.72/4.50             & 64.07/5.36            & 59.27/5.75            & 56.26/6.18           \\ 
                                        & 32.0          & 56.39/12.42            & 55.35/10.31           & 51.11/11.50           & 50.12/11.60          \\
\midrule        
\textbf{ABC(1$\times$1)}\cite{lin2017}  & 8.0           & 69.76/4.04             & 64.42/4.17            & 60.77/4.04            & 57.87/4.50           \\
(86.29 \%)                              & 16.0          & 68.01/4.62             & 61.97/4.94            & 58.18/5.22            & 55.11/5.81           \\ 
                                        & 32.0          & 62.79/7.54             & 55.62/9.21            & 52.19/9.57            & 50.94/9.44           \\
\midrule                                        
\textbf{ABC(3$\times$3)}\cite{lin2017}  & 8.0           & 76.48/2.76             & 70.92/2.72            & 67.41/3.15            & 64.37/3.35           \\
(92.48 \%)                              & 16.0          & 74.71/3.69             & 67.79/4.35            & 64.01/4.47            & 61.12/4.83           \\ 
                                        & 32.0          & 65.46/8.05             & 57.34/9.68            & 54.46/9.96            & 52.10/10.54          \\
\midrule                                        
\textbf{ABC(5$\times$5)}\cite{lin2017}  & 8.0           & 77.42/2.57             & 72.62/2.74            & 68.07/2.96            & 66.04/3.34           \\
(92.82 \%)                              & 16.0          & 75.82/3.57             & 69.66/3.96            & 64.44/4.35            & 61.73/4.90           \\ 
                                        & 32.0          & 65.30/8.76             & 57.46/10.68           & 54.74/11.09           & 52.29/11.01          \\
                                        
\bottomrule                                        
\multicolumn{6}{r}{OA/TA = Accuracy to original label / Accuracy to target label.}\\ 
\end{tabular}}
\vspace{2ex}
\caption{Accuracy (Top1) [\%] of ResNet56 like CNN after LocalSearch adversarial attacks for CIFAR-10. BatchNorm~\cite{bn} layers are in $\mathtt{mini}$-$\mathtt{batch}$ mode.}
\label{tab:LocalSearchResNet56_train}
\end{table}

% -------------------------------------------------- EVAL MODE ---------------------------------------------------------- %

\begin{table*}[!]
\centering
\resizebox{0.7\textwidth}{!}{
\renewcommand{\arraystretch}{1.0}
\begin{tabular}{l|ccccc}
\toprule
\multirow{2}{*}{\textbf{LocalSearch}}&  \textbf{Nat.Acc}             & $i=50$            & $i=100$           & $i=150$           & $i=200$              \\ 
                                                    &     & \textbf{OA/TA}    & \textbf{OA/TA}    & \textbf{OA/TA}   & \textbf{OA/TA}      \\
\midrule
\midrule
\textbf{ResNet50}~\cite{He2015DeepRL} &75.43 \%    & 35.44/0.08        & 31.95/0.12        & 29.87/0.16        & 27.72/0.16         \\
%\midrule
\textbf{ResNet18}~\cite{He2015DeepRL} &69.00 \%    & 26.82/0.20        & 24.10/0.22        & 22.42/0.26        & 20.91/0.24   \\
%\midrule
\textbf{ResNet18-Ch.Prune~\cite{AMC2019}} &67.62 \% & 20.61/0.24     & 18.17/0.30    & 16.97/0.36   & 15.34/0.38 \\
%\midrule
\textbf{ResNet18-XNOR}~\cite{rastegari2016}& 49.10 \%      & 11.24/0.24            & 9.03/0.26          & 7.55/0.30                & 6.71/0.34            \\
%\midrule
\textbf{ResNet18-ABC(1$\times$1)}~\cite{lin2017} &51.07 \% & 14.30/0.26     & 11.86/0.28   & 10.48/0.34        & 8.57/0.36 \\
%\midrule
\textbf{ResNet18-ABC(3$\times$3)}~\cite{lin2017}& 59.83 \% & 20.10/0.30     & 18.20/0.40   & 15.60/0.50       & 14.00/0.50           \\
\bottomrule                                        
\multicolumn{5}{r}{OA/TA = Accuracy to original label / Accuracy to target label.}\\ 
\end{tabular}}
\vspace{2ex}
\caption{Accuracy (Top1) [\%] of CNN models after LocalSearch adversarial attacks for ImageNet dataset. $\epsilon=8$.}
\label{tab:LocalSearchImageNet_eval}
\end{table*}

% -------------------------------------------------- EVAL MODE ---------------------------------------------------------- %

\begin{table}[!]
\centering
\resizebox{0.7\textwidth}{!}{
\renewcommand{\arraystretch}{1.0}
\begin{tabular}{l|cccc}
\toprule
\multirow{2}{*}{\textbf{LocalSearch}}               & $i=50$            & $i=100$           & $i=150$           & $i=200$              \\ 
                                                    & \textbf{OA/TA}    & \textbf{OA/TA}    & \textbf{OA/TA}    & \textbf{OA/TA}       \\
\midrule
\midrule
\textbf{ResNet50}~\cite{He2015DeepRL} (75.43 \%)    & 40.48/0.10        & 37.82/0.10        & 36.18/0.14        & 34.90/0.12        \\
\midrule
\textbf{ResNet18}~\cite{He2015DeepRL} (69.00 \%)    & 15.81/0.06        & 14.84/0.02        & 14.36/0.02        & 14.00/0.02           \\
\midrule
\textbf{ResNet18-Ch.Prune~\cite{AMC2019}} (67.62 \%) & 26.18/0.08        & 24.10/0.08        & 23.12/0.10        & 22.74/0.12        \\
\midrule
\textbf{ResNet18-XNOR}~\cite{rastegari2016} (49.10 \%)     & 17.13/0.12        & 16.15/0.14        & 14.44/0.18        & 14.26/0.16        \\
\midrule
\textbf{ResNet18-ABC(1$\times$1)}~\cite{lin2017} (51.07 \%) & 17.49/0.12        & 16.61/0.14        & 15.10/0.08        & 14.72/0.08        \\
\midrule
\textbf{ResNet18-ABC(3$\times$3)}~\cite{lin2017} (59.83 \%) & 22.50/0.10        & 20.10/0.00        & 19.00/0.10        & 18.00/0.20        \\             
\multicolumn{5}{r}{OA/TA = Accuracy to original label / Accuracy to target label.}\\ 
\end{tabular}}
\vspace{2ex}
\caption{Accuracy (Top1) [\%] of CNN models after LocalSearch adversarial attacks for ImageNet dataset. $\epsilon=8$. BatchNorm~\cite{bn} layers are in $\mathtt{mini}$-$\mathtt{batch}$ mode.}
\label{tab:LocalSearchImageNet_train}
\end{table}

    % -------------------------------------------------- EVAL MODE ---------------------------------------------------------- %
\begin{table*}[ht]
\centering
\resizebox{0.7\textwidth}{!}{
\begin{tabular}{l|c|c|cccc}
\toprule
\multirow{2}{*}{\textbf{GenAttack}}     &\multirow{2}{*}{$\epsilon$}  & Pop      & $i=50$             & $i=100$           & $i=150$           & $i=200$        \\ 
                                        &                       & size     & \textbf{OA/TA}     & \textbf{OA/TA}    & \textbf{OA/TA}    & \textbf{OA/TA} \\
\midrule
\midrule
\textbf{Vanilla}\cite{He2015DeepRL}     & 8.0                   & 6        & 41.26/35.80        & 16.20/70.62       &  6.21/87.33       &  2.98/93.78 \\
(92.46 \%)                              &                       & 16       & 19.77/67.33        & 3.28/93.929       &  0.64/98.66       &  0.19/99.61 \\
                                        & 12.0                  & 6        & 24.51/50.77        &  7.44/82.31       &  2.69/93.30       &  1.09/96.91 \\        
                                        &                       & 16       & 8.43/82.05         &  1.12/97.44       &  0.30/99.47       &  0.09/99.88 \\
\midrule          
%\textbf{KD-CE}\cite{contrastive_kd}    & 8.0                   & 6        & 42.89/35.25        & 14.91/74.08       &  4.42/91.72       &  1.16/97.57  \\
%(92.70 \%)                              &                       & 16       & 17.66/71.25        & 1.24/97.52        &  0.12/99.76       &  0.00/100.00 \\
%                                        & 12.0                  & 6        & 25.22/51.82        & 5.63/87.78        &  1.15/97.25       &  0.68/96.56  \\
%                                        &                       & 16       & 6.30/87.26         & 0.11/99.62        &  0.01/99.95       &  0.00/100.00 \\
%\midrule                                
\textbf{KD-KL}\cite{contrastive_kd}    & 8.0                   & 6        & 48.50/18.71        & 29.21/37.72       & 19.05/52.50       & 14.23/62.63  \\
(93.25 \%)                              &                       & 16       & 31.86/36.82        & 15.48/63.23       &  9.76/76.07       &  6.91/83.38  \\
                                        & 12.0                  & 6        & 31.48/24.54        & 18.34/44.95       & 12.97/57.97       &  9.78/66.71  \\
                                        &                       & 16       & 18.87/46.52        &  9.66/70.43       &  6.08/80.71       &  4.61/86.14  \\
\midrule                                
\textbf{Ch.Prune} \cite{AMC2019}   & 8.0                   & 6        & 37.88/41.03        & 12.35/78.30       & 3.95/92.12        &  1.46/97.00  \\
(89.76 \%)                              &                       & 16       & 14.45/75.62        &  1.54/96.94       & 0.18/99.56        &  0.03/99.89  \\
                                        & 12.0                  & 6        & 21.60/56.24        &  5.17/87.73       & 1.19/96.45        &  0.40/98.81  \\
                                        &                       & 16       & 5.56/88.19         &  0.34/98.99       & 0.03/99.88        &  0.01/99.98  \\
\midrule
%\textbf{F.Prune} \cite{AMC2019}    & 8.0                   & 6        & 37.43/39.36        & 12.57/74.80       & 3.74/91.09        &  1.41/96.47  \\
%(89.92 \%)                              &                       & 16       & 15.52/72.97        &  1.53/96.63       & 0.18/99.66        &  0.05/99.91  \\
%                                        & 12.0                  & 6        & 22.06/51.78        &  5.32/84.13       & 1.73/94.57        &  0.45/98.21  \\
%                                        &                       & 16       & 5.81/84.97         &  0.30/98.95       & 0.05/99.85        &  0.00/99.98  \\
%\midrule                                        
\textbf{K.Prune}~\cite{AMC2019}    & 8.0                   & 6        & 38.99/37.23        & 13.00/73.47       &  4.72/89.59       &  1.82/95.56\\
(90.73 \%)                              &                       & 16       & 16.61/69.82        &  1.94/95.53       &  0.36/99.20       &  0.05/99.85\\
                                        & 12.0                  & 6        & 21.19/52.53        &  5.09/85.16       &  1.63/95.06       &  0.61/98.25\\
                                        &                       & 16       &  5.92/85.05        &  0.53/98.47       &  0.04/99.78       &  0.02/99.48\\ 
\midrule
\textbf{W.Prune}~\cite{AMC2019}    & 8.0                   & 6        & 41.79/35.02        & 17.07/68.59       &  7.28/18.76       &  2.95/93.16\\
(91.98 \%)                              &                       & 16       & 20.41/65.74        &  3.24/92.87       &  0.66/98.30       &  0.31/99.34\\
                                        & 12.0                  & 6        & 25.61/47.95        &  7.85/80.04       &  3.08/91.54       &  1.36/96.12\\
                                        &                       & 16       &  8.88/80.30        &  1.29/96.81       &  0.24/99.21       &  0.06/99.81\\
\midrule                                
\textbf{XNOR}\cite{rastegari2016}       & 8.0                   & 6        & 72.00/9.99         & 70.85/11.69       & 69.97/12.70       &  69.23/13.41 \\
(82.71 \%)                              &                       & 16       & 69.26/13.25        & 67.67/15.07       & 66.19/16.45       &  65.72/17.37 \\
                                        & 12.0                  & 6        & 66.94/11.64        & 65.05/13.78       & 63.71/14.99       &  62.71/16.06 \\
                                        &                       & 16       & 63.57/15.94        & 60.92/18.20       & 59.51/19.68       &  58.88/20.60 \\
\midrule                                
\textbf{ABC(1$\times$1)}\cite{lin2017}  & 8.0                   & 6        & 71.06/10.53        & 69.30/12.19       & 68.23/13.17       & 67.71/13.95  \\
(83.42 \%)                              &                       & 16       & 67.67/13.71        & 66.18/15.80       & 64.60/17.14       & 64.13/17.89  \\
                                        & 12.0                  & 6        & 64.23/11.51        & 62.65/13.91       & 61.46/15.07       & 59.95/16.48  \\
                                        &                       & 16       & 60.71/15.98        & 59.08/18.46       & 57.09/20.59       & 56.29/20.93  \\
\midrule                                
\textbf{ABC(3$\times$3)}\cite{lin2017}  & 8.0                   & 6        & 74.69/8.99         & 73.00/10.51       & 72.25/11.90       & 71.02/12.76  \\
(88.94 \%)                              &                       & 16       & 71.52/12.44        & 69.79/14.44       & 68.69/15.42       & 67.59/16.54  \\
                                        & 12.0                  & 6        & 66.47/10.36        & 64.10/12.59       & 63.37/14.12       & 62.24/14.70  \\
                                        &                       & 16       & 62.39/15.19        & 59.59/17.45       & 58.59/19.29       & 57.28/20.49  \\
\midrule
\textbf{ABC(5$\times$5)}\cite{lin2017}  & 8.0                   & 6        & 77.91/7.87         & 76.20/75.41       & 75.41/10.05       & 74.57/10.78  \\
(90.64 \%)                              &                       & 16       & 75.24/10.43        & 73.47/12.10       & 72.10/13.10       & 71.19/14.09  \\
                                        & 12.0                  & 6        & 69.48/9.13         & 67.57/11.29       & 65.91/12.57       & 64.80/13.51  \\
                                        &                       & 16       & 65.39/13.41        & 62.79/15.89       & 60.78/17.36       & 59.59/18.75  \\
%\midrule
%\textbf{ABC(7$\times$7)}\cite{lin2017}  & 8.0                   & 6        & 79.76/5.67         & 78.79/6.72        & 78.06/7.45        & 77.54/7.86   \\
%(91.34 \%)                              &                       & 16       & 77.48/7.69         & 75.99/8.92        & 74.87/9.62        & 74.35/10.21  \\
%                                        & 12.0                  & 6        & 71.62/7.17         & 69.42/8.82        & 68.19/10.03       & 66.73/10.75  \\
%                                        &                       & 16       & 67.39/10.62        & 64.92/12.72       & 63.20/14.29       & 61.97/15.32  \\
\bottomrule
\multicolumn{7}{r}{OA/TA = Accuracy to original label / Accuracy to target label.}\\ 
\end{tabular}}
\vspace{2ex}
\caption{Accuracy (Top1) [\%] of ResNet20 like CNNs after GenAttack adversarial attacks for CIFAR-10.}
\label{tab:GenAttack_ResNet20_eval}
\end{table*}
    
% ------------------------------------------ %
%              Gen Attack                    %
% ------------------------------------------ %

% ------------------------------------------ %
%              ResNet 20                     %
% ------------------------------------------ %

% ------------------------------------------ %
%              DS: Cifar-10                  %
% ------------------------------------------ %

% ------------------------------------------------------------------------------------ %
%  Without Adaptive parameter scaling for  alpha(mutaion_range) and rho(mutation_rate) %
% ------------------------------------------------------------------------------------ %

% -------------------------------------------------- TRAIN MODE ---------------------------------------------------------- %

\begin{table}[!h]
\centering
\resizebox{0.7\textwidth}{!}{
\begin{tabular}{l|c|l|cccc}
\toprule
\textbf{GenAttack}                      &$\epsilon$             & Pop      & $i=50$             & $i=100$           & $i=150$           & $i=200$     \\ 
                                        &                       & size     & \textbf{OA/TA}     & \textbf{OA/TA}    & \textbf{OA/TA}    & \textbf{OA/TA}\\
\midrule
\midrule
\textbf{Vanilla}~\cite{He2015DeepRL}    & 8.0                   & 6        & 57.12/11.96        & 46.91/17.14       & 43.02/19.28       & 40.49/20.51\\
(92.46 \%)                              &                       & 16       & 47.24/16.87        & 39.91/21.19       & 36.91/22.56       & 35.33/23.47\\
                                        & 12.0                  & 6        & 64.37/12.33        & 54.87/18.91       & 49.83/22.54       & 47.69/24.32\\     
                                        &                       & 16       & 46.77/17.52        & 41.20/21.08       & 38.68/22.16       & 36.17/22.76\\ 
\midrule                                        
\textbf{KD-CE}~\cite{contrastive_kd}   & 8.0                   & 6        & 59.81/10.70        & 48.14/16.05       & 42.91/19.05       & 40.09/20.95\\
(92.70 \%)                              &                       & 16       & 48.12/16.27        & 38.94/21.69       & 34.98/24.13       & 33.51/25.11\\
                                        & 12.0                  & 6        & 56.03/11.38        & 47.14/16.07       & 43.73/18.81       & 40.45/20.36\\
                                        &                       & 16       & 47.48/16.82        & 40.11/21.57       & 36.68/23.02       & 35.33/23.46\\ 
\midrule                                        
\textbf{KD-KL}~\cite{contrastive_kd}   & 8.0                   & 6        & 66.04/11.05        & 54.80/18.47       & 48.94/22.37       & 45.56/24.80\\
(93.25 \%)                              &                       & 16       & 53.73/18.59        & 44.62/25.15       & 41.90/26.86       & 40.68/26.36\\
                                        & 12.0                  & 6        & 64.28/12.09        & 54.18/19.41       & 49.97/22.42       & 48.33/24.20\\
                                        &                       & 16       & 53.73/18.59        & 46.20/24.95       & 43.78/25.69       & 42.53/25.55\\ 
\midrule                                        
\textbf{Ch.Prune}~\cite{AMC2019}   & 8.0                   & 6        & 54.81/12.69        & 45.85/17.76       & 41.44/20.22       & 39.40/21.33\\
(89.76 \%)                              &                       & 16       & 45.59/17.74        & 38.64/21.88       & 35.64/23.40       & 34.73/23.66\\
                                        & 12.0                  & 6        & 52.27/13.52        & 46.42/17.26       & 43.27/19.24       & 41.38/20.16\\
                                        &                       & 16       & 45.07/18.16        & 40.45/20.89       & 38.30/22.05       & 36.77/22.47\\ 
\midrule
\textbf{F.Prune}~\cite{AMC2019}    & 8.0                   & 6        & 56.38/12.22        & 47.82/17.12       & 45.17/19.09       & 42.32/20.20\\
(89.92 \%)                              &                       & 16       & 47.18/17.21        & 41.29/20.95       & 38.59/21.97       & 37.33/22.29\\
                                        & 12.0                  & 6        & 53.63/13.60        & 48.67/16.71       & 45.23/18.87       & 44.33/19.64\\
                                        &                       & 16       & 47.66/17.47        & 42.26/20.37       & 39.92/20.97       & 38.24/21.37\\
\midrule                                        
\textbf{K.Prune}~\cite{AMC2019}    & 8.0                   & 6        & 57.94/11.43        & 48.05/15.96       & 44.64/18.16       & 42.35/19.41\\
(90.73 \%)                              &                       & 16       & 48.15/16.50        & 41.30/20.19       & 38.99/21.70       & 37.50/22.77\\
                                        & 12.0                  & 6        & 55.30/12.27        & 49.74/15.82       & 46.34/17.73       & 44.69/18.57\\
                                        &                       & 16       & 47.84/16.47        & 43.65/19.35       & 41.11/20.36       & 39.41/21.40\\ 
\midrule
\textbf{W.Prune}~\cite{AMC2019}    & 8.0                   & 6        & 57.71/11.55        & 47.57/16.25       & 43.32/18.76       & 41.37/20.02\\
(91.98 \%)                              &                       & 16       & 47.43/16.33        & 39.76/20.29       & 37.04/22.06       & 35.12/23.02\\
                                        & 12.0                  & 6        & 55.02/12.21        & 47.93/16.36       & 44.71/18.22       & 42.53/19.54\\
                                        &                       & 16       & 46.82/17.08        & 41.12/20.55       & 38.76/22.10       & 37.03/22.84\\
\midrule                                        
\textbf{XNOR}~\cite{rastegari2016}      & 8.0                   & 6        & 80.65/2.94         & 80.02/3.25        & 79.85/3.33        & 80.36/3.10 \\
(82.71 \%)                              &                       & 16       & 79.93/3.07         & 79.48/3.16        & 80.01/3.16        & 79.79/3.24 \\
                                        & 12.0                  & 6        & 78.30/3.43         & 78.31/3.63        & 77.96/3.82        & 77.69/3.93 \\
                                        &                       & 16       & 77.90/3.88         & 77.57/4.20        & 77.43/4.34        & 77.10/4.17 \\ 
\midrule                                        
\textbf{ABC(1$\times$1)}~\cite{lin2017} & 8.0                   & 6        & 80.54/2.66         & 80.41/2.42        & 80.23/2.73        & 80.35/2.89\\
(83.42 \%)                              &                       & 16       & 80.00/2.87         & 80.28/2.92        & 79.91/2.79        & 79.68/2.95\\
                                        & 12.0                  & 6        & 77.63/2.97         & 77.86/3.25        & 77.59/3.18        & 77.14/2.97\\
                                        &                       & 16       & 76.45/3.28         & 77.10/\3.37       & 76.39/3.69        & 77.34/3.58\\ 
\midrule                                        
\textbf{ABC(3$\times$3)}~\cite{lin2017} & 8.0                   & 6        & 84.74/1.93         & 85.60/1.97        & 84.66/1.99        & 84.73/1.79\\
(88.94 \%)                              &                       & 16       & 84.46/2.14         & 84.44/2.19        & 84.07/2.33        & 84.03/2.50\\
                                        & 12.0                  & 6        & 82.41/2.19         & 82.28/2.33        & 81.95/2.37        & 81.91/2.37\\
                                        &                       & 16       & 81.70/2.53         & 81.22/2.69        & 81.58/2.77        & 80.70/2.87\\
\midrule        
\textbf{ABC(5$\times$5)}~\cite{lin2017} & 8.0                   & 6        & 86.40/1.81         & 86.46/1.85        & 86.35/1.83        & 86.35/1.97\\
(90.64 \%)                              &                       & 16       & 85.65/2.11         & 85.64/2.01        & 85.43/2.13        & 85.23/2.10\\
                                        & 12.0                  & 6        & 83.37/2.16         & 82.26/2.44        & 82.35/2.35        & 81.91/2.63\\
                                        &                       & 16       & 82.22/2.57         & 81.50/2.85        & 81.67/2.75        & 80.95/2.91\\ 
\midrule        
\textbf{ABC(7$\times$7)}~\cite{lin2017} & 8.0                   & 6        & 87.19/1.57         & 86.70/1.85        & 86.45/1.85        & 86.24/1.71\\
(91.34 \%)                              &                       & 16       & 86.02/1.88         & 85.71/1.87        & 85.71/2.02        & 85.29/2.10\\
                                        & 12.0                  & 6        & 83.70/2.05         & 83.10/2.36        & 82.35/2.42        & 82.63/2.37\\
                                        &                       & 16       & 81.86/2.71         & 81.33/2.98        & 80.70/3.03        & 80.44/3.28\\ 
\bottomrule
\multicolumn{7}{r}{OA/TA = Accuracy to original label / Accuracy to target label.}\\ 
\end{tabular}}
\vspace{2ex}
\caption{Accuracy (Top1) [\%] of ResNet20 like CNNs after GenAttack adversarial attacks for CIFAR-10. BatchNorm~\cite{bn} layers are in $\mathtt{mini}$-$\mathtt{batch}$ mode.}
\label{tab:GenAttack_ResNet20_train}
\end{table}

% ----------------------------------------- %
%              Gen Attack                   %
% ----------------------------------------- %

% ----------------------------------------- %
%              ResNet 56                    %
% ----------------------------------------- %

% ----------------------------------------- %
%              DS: Cifar-10                 %
% ----------------------------------------- %

% -------------------------------------------------- EVAL MODE ---------------------------------------------------------- %

\begin{table*}[ht]
\centering
\resizebox{0.7\textwidth}{!}{
\renewcommand{\arraystretch}{0.9}
\begin{tabular}{l|c|l|cccc}
\toprule
\textbf{GenAttack}                      &\textbf{$\epsilon$ }   & Pop      & $i=50$             & $i=100$           & $i=150$           & $i=200$        \\ 
                                        &                       & size     & \textbf{OA/TA}     & \textbf{OA/TA}    & \textbf{OA/TA}    & \textbf{OA/TA} \\
\midrule
\midrule
\textbf{Vanilla}\cite{He2015DeepRL}     & 8.0                   & 6        & 50.35/23.13        & 28.34/49.83       & 16.86/68.07       & 9.97/79.82     \\
(93.88 \%)                              &                       & 16       & 31.97/47.30        & 11.12/78.93       & 4.00/91.26        & 1.69/96.10     \\
                                        & 12.0                  & 6        & 34.21/32.73        & 16.15/62.43       & 8.47/78.65        & 4.52/87.94     \\      
                                        &                       & 16       & 18.01/61.75        &  4.54/88.72       & 1.48/95.99        & 0.49/98.51     \\
\midrule                                        
%\textbf{KD-CE}~\cite{contrastive_kd}   & 8.0                   & 6        & 56.09/20.83        & 34.33/46.20       & 21.31/64.21       & 13.80/76.09    \\
%(93.68 \%)                              &                       & 16       & 37.75/42.76        & 16.04/74.45       &  7.25/88.36       &  3.62/94.26    \\
%                                        & 12.0                  & 6        & 39.08/32.28        & 19.04/61.49       & 10.20/78.03       &  5.41/87.82    \\
%                                        &                       & 16       & 21.25/60.10        &  6.66/87.72       &  2.27/95.66       &  1.02/98.08    \\ 
%\midrule                                        
\textbf{KD-KL}~\cite{contrastive_kd}   & 8.0                   & 6        & 59.05/15.53        & 41.20/33.36       & 29.28/47.78       & 21.56/59.01    \\
(94.24 \%)                              &                       & 16       & 44.16/31.20        & 24.07/57.20       & 13.83/73.81       &  8.54/83.45    \\
                                        & 12.0                  & 6        & 42.77/23.83        & 25.40/45.96       & 16.71/61.70       & 11.57/72.07    \\
                                        &                       & 16       & 28.95/45.02        & 12.58/72.30       &  6.20/85.14       &  3.43/91.65    \\
\midrule                                        
\textbf{Ch.Prune} \cite{AMC2019}   & 8.0                   & 6        & 50.18/25.29        & 29.40/50.68       & 16.76/69.19       & 10.06/80.33    \\
(92.93 \%)                              &                       & 16       & 30.86/49.12        & 11.09/79.41       & 4.21/91.90        &  1.84/96.07    \\
                                        & 12.0                  & 6        & 34.63/35.33        & 16.11/63.98       & 8.58/79.50        &  4.61/88.61    \\
                                        &                       & 16       & 17.99/63.94        & 4.53/89.48        & 1.52/96.36        &  0.53/98.56    \\
\midrule                                        
%\textbf{F.Prune} \cite{AMC2019}    & 8.0                   & 6        & 49.70/22.99        & 29.89/47.15       & 19.18/63.53       & 12.00/74.55    \\
%(93.08 \%)                              &                       & 16       & 32.21/45.96        & 12.88/75.07       &  5.63/87.85       &  2.83/93.67    \\
%                                        & 12.0                  & 6        & 35.18/31.38        & 18.18/57.77       & 10.67/73.34       &  6.77/82.65    \\
%                                        &                       & 16       & 19.20/58.33        &  6.74/83.89       &  2.58/92.95       &  0.88/96.83    \\
%\midrule                                        
\textbf{K.Prune} \cite{AMC2019}    & 8.0                   & 6        & 47.00/25.97        & 23.83/54.98       & 12.60/73.48       &  7.31/83.68    \\
(93.03 \%)                              &                       & 16       & 26.85/52.01        &  7.72/82.75       &  2.65/93.71       &  1.07/97.44    \\
                                        & 12.0                  & 6        & 30.43/37.30        & 12.23/68.49       &  5.27/84.25       &  2.81/91.82    \\
                                        &                       & 16       & 13.88/67.42        &  2.76/92.12       &  0.85/97.67       &  0.35/99.16    \\
\midrule                                        
\textbf{W.Prune} \cite{AMC2019}    & 8.0                   & 6        & 49.48/23.25        & 27.61/49.16       & 16.42/67.39       &  9.85/78.75    \\
(93.54 \%)                              &                       & 16       & 31.46/46.66        & 11.03/77.82       &  4.40/90.48       &  1.91/95.60    \\
                                        & 12.0                  & 6        & 33.37/32.88        & 16.09/61.55       &  8.53/77.75       &  4.88/86.84    \\
                                        &                       & 16       & 17.50/61.47        &  4.61/87.89       &  1.47/95.70       &  0.40/98.44    \\
\midrule                                        
\textbf{XNOR}\cite{rastegari2016}       & 8.0                   & 6        & 69.76/13.61        & 67.81/15.67       & 66.81/16.91       & 66.40/17.68    \\
(83.24 \%)                              &                       & 16       & 66.19/17.86        & 64.48/19.92       & 63.34/21.23       & 62.12/22.27    \\
                                        & 12.0                  & 6        & 64.85/14.50        & 62.84/16.85       & 61.54/18.30       & 60.51/19.56    \\
                                        &                       & 16       & 60.56/19.67        & 58.18/22.63       & 56.74/24.33       & 55.47/25.60    \\
\midrule                                        
\textbf{ABC(1$\times$1)}\cite{lin2017}  & 8.0                   & 6        & 72.23/9.43         & 70.69/10.88       & 69.46/11.88       & 68.79/12.65    \\
(86.29 \%)                              &                       & 16       & 69.10/12.86        & 67.46/14.71       & 66.41/15.71       & 65.89/16.55    \\
                                        & 12.0                  & 6        & 64.10/10.41        & 62.28/12.40       & 60.97/13.44       & 60.14/14.22    \\
                                        &                       & 16       & 60.29/14.56        & 57.81/16.89       & 56.81/18.17       & 55.13/19.25    \\
\midrule                                        
\textbf{ABC(3$\times$3)}\cite{lin2017}  & 8.0                   & 6        & 80.96/5.58         & 79.37/6.82        & 78.00/7.52        & 77.53/8.04     \\
(92.48 \%)                              &                       & 16       & 77.82/8.16         & 76.15/9.53        & 75.07/10.51       & 74.45/11.18    \\
                                        & 12.0                  & 6        & 72.28/6.99         & 70.17/8.63        & 68.24/9.71        & 67.60/10.46    \\
                                        &                       & 16       & 69.23/10.23        & 66.55/12.48       & 64.55/13.89       & 63.48/15.01    \\
\midrule                                        
\textbf{ABC(5$\times$5)}\cite{lin2017}  & 8.0                   & 6        & 82.51/4.52         & 81.22/5.40        & 80.46/6.06        & 80.09/6.43     \\
(92.82 \%)                              &                       & 16       & 79.87/6.64         & 78.75/7.85        & 77.93/8.59        & 77.23/9.17     \\
                                        & 12.0                  & 6        & 74.32/6.42         & 71.83/8.04        & 70.28/9.20        & 69.59/9.95     \\
                                        &                       & 16       & 69.91/9.65         & 67.34/11.64       & 65.77/12.75       & 64.80/13.82    \\
\bottomrule
\multicolumn{7}{r}{OA/TA = Accuracy to original label / Accuracy to target label.}\\ 
\end{tabular}}
\vspace{2ex}
\caption{Accuracy (Top1) [\%] of ResNet56 like CNNs after GenAttack adversarial attacks for CIFAR-10. 
%BatchNorm~\cite{bn} layers are in $\mathtt{statistics}$ mode.
}
\label{tab:GenAttackResNet56_eval}
\end{table*}
    
% ----------------------------------------- %
%              Gen Attack                   %
% ----------------------------------------- %

% ----------------------------------------- %
%              ResNet 56                    %
% ----------------------------------------- %

% ----------------------------------------- %
%              DS: Cifar-10                 %
% ----------------------------------------- %

% -------------------------------------------------- TRAIN MODE ---------------------------------------------------------- %

\begin{table}[!h]
\centering
\resizebox{0.7\textwidth}{!}{
\begin{tabular}{l|c|l|cccc}
\toprule
\textbf{GenAttack}                      &\textbf{$\epsilon$}    & Pop      & $i=50$             & $i=100$           & $i=150$           & $i=200$        \\ 
                                        &                       & size     & \textbf{OA/TA}     & \textbf{OA/TA}    & \textbf{OA/TA}    & \textbf{OA/TA} \\ 
\midrule
\midrule
\textbf{Vanilla}\cite{He2015DeepRL}     & 8.0                   & 6        & 62.62/12.00        & 50.19/18.25       & 44.38/21.50       & 41.45/22.88    \\
(93.88 \%)                              &                       & 16       & 50.49/18.42        & 41.02/23.51       & 36.50/23.77       & 34.13/24.06    \\
                                        & 12.0                  & 6        & 58.25/13.45        & 48.84/19.32       & 44.19/21.46       & 42.11/22.36    \\      
                                        &                       & 16       & 49.07/18.82        & 40.42/22.35       & 37.04/22.93       & 36.18/22.75    \\ 
\midrule                                        
\textbf{KD-CE}~\cite{contrastive_kd}   & 8.0                   & 6        & 66.11/12.73        & 52.78/21.22       & 46.25/24.65       & 42.88/25.70   \\
(93.68 \%)                              &                       & 16       & 53.70/20.66        & 42.01/26.95       & 38.20/25.76       & 36.16/24.89   \\
                                        & 12.0                  & 6        & 60.90/15.31        & 49.75/22.47       & 44.28/24.68       & 41.50/24.56   \\
                                        &                       & 16       & 48.61/22.88        & 40.55/25.11       & 37.74/24.78       & 36.82/23.90   \\ 
\midrule                                        
\textbf{KD-KL}~\cite{contrastive_kd}   & 8.0                   & 6        & 67.06/11.06        & 55.10/18.72       & 48.86/22.86       & 44.87/24.86   \\
(94.24 \%)                              &                       & 16       & 54.59/18.60        & 43.13/25.92       & 39.20/26.69       & 37.12/25.86   \\
                                        & 12.0                  & 6        & 63.07/13.39        & 52.77/19.99       & 47.78/22.50       & 45.61/23.27   \\
                                        &                       & 16       & 51.60/20.73        & 42.93/24.27       & 40.21/24.04       & 38.46/23.43   \\ 
\midrule                                        
\textbf{Ch.Prune} \cite{AMC2019}   & 8.0                   & 6        & 61.65/12.12        & 49.16/18.17       & 43.51/20.98       & 40.39/22.23    \\
(92.93 \%)                              &                       & 16       & 49.03/18.06        & 39.15/22.36       & 35.29/23.01       & 33.67/22.87    \\
                                        & 12.0                  & 6        & 57.23/13.67        & 47.96/18.09       & 43.19/19.89       & 40.35/20.76    \\
                                        &                       & 16       & 46.54/18.98        & 39.35/21.75       & 36.60/22.05       & 34.26/22.01    \\
\midrule                                        
\textbf{F.Prune} \cite{AMC2019}    & 8.0                   & 6        & 60.86/11.64        & 48.45/18.19       & 43.29/21.24       & 40.23/22.39    \\
(93.08 \%)                              &                       & 16       & 48.80/18.13        & 39.40/23.03       & 35.72/23.70       & 33.30/23.66    \\
                                        & 12.0                  & 6        & 57.28/13.21        & 47.83/18.76       & 44.20/20.57       & 41.37/21.69    \\
                                        &                       & 16       & 46.24/19.30        & 39.43/22.48       & 37.01/23.08       & 35.00/22.71    \\
\midrule                                        
\textbf{K.Prune} \cite{AMC2019}    & 8.0                   & 6        & 59.59/12.17        & 47.67/18.55       & 42.18/21.32       & 39.32/22.64    \\
(93.03 \%)                              &                       & 16       & 47.97/18.49        & 38.68/22.70       & 34.61/23.77       & 32.86/23.89    \\
                                        & 12.0                  & 6        & 56.01/13.46        & 46.71/18.90       & 42.05/21.05       & 39.96/21.61    \\
                                        &                       & 16       & 46.33/19.17        & 39.31/22.82       & 36.55/22.51       & 34.46/22.67    \\
\midrule                                        
\textbf{W.Prune} \cite{AMC2019}    & 8.0                   & 6        & 62.01/11.76        & 49.63/18.50       & 44.14/21.47       & 41.10/22.99    \\
(93.54 \%)                              &                       & 16       & 49.73/18.41        & 40.41/23.27       & 36.37/24.08       & 34.63/23.82    \\
                                        & 12.0                  & 6        & 58.03/13.29        & 48.44/19.26       & 43.94/20.83       & 41.29/21.66    \\
                                        &                       & 16       & 46.89/19.34        & 40.30/22.85       & 36.85/23.72       & 35.87/22.96    \\
\midrule                                        
\textbf{XNOR}\cite{rastegari2016}       & 8.0                   & 6        & 80.38/2.52         & 80.48/2.58        & 80.61/2.61        & 79.86/2.96     \\
(83.24 \%)                              &                       & 16       & 80.10/2.86         & 79.90/2.88        & 79.94/2.80        & 80.13/2.81     \\
                                        & 12.0                  & 6        & 78.90/3.09         & 78.47/3.21        & 78.34/3.20        & 78.34/3.22     \\
                                        &                       & 16       & 77.65/3.47         & 77.62/3.53        & 77.22/3.75        & 77.54/3.72     \\
\midrule                                        
\textbf{ABC(1$\times$1)}\cite{lin2017}  & 8.0                   & 6        & 82.87/2.65         & 82.27/2.69        & 82.31/2.78        & 82.09/2.85     \\
(86.29 \%)                              &                       & 16       & 82.06/2.96         & 82.10/2.98        & 81.36/3.02        & 81.62/2.97     \\
                                        & 12.0                  & 6        & 79.83/3.19         & 79.47/3.35        & 79.12/3.36        & 79.12/3.51     \\
                                        &                       & 16       & 79.08/3.75         & 78.90/3.88        & 78.98/3.84        & 78.30/4.03     \\ 
\midrule                                        
\textbf{ABC(3$\times$3)}\cite{lin2017}  & 8.0                   & 6        & 88.11/1.62         & 87.93/1.66        & 87.45/1.81        & 87.18/1.87     \\
(92.48 \%)                              &                       & 16       & 87.18/1.88         & 86.82/1.94        & 87.16/1.90        & 86.97/2.02     \\
                                        & 12.0                  & 6        & 84.85/2.06         & 84.39/2.23        & 83.68/2.42        & 83.93/2.37     \\
                                        &                       & 16       & 83.70/2.24         & 83.34/2.72        & 83.24/2.55        & 82.64/2.81     \\
\midrule                                        
\textbf{ABC(5$\times$5)}\cite{lin2017}  & 8.0                   & 6        & 88.41/1.58         & 88.18/1.67        & 88.22/1.64        & 87.67/1.65     \\
(92.82 \%)                              &                       & 16       & 87.43/1.80         & 87.30/1.85        & 87.01/1.85        & 87.19/1.81     \\
                                        & 12.0                  & 6        & 85.29/2.05         & 84.54/2.20        & 84.33/2.36        & 84.10/2.57     \\
                                        &                       & 16       & 84.44/2.47         & 83.60/2.61        & 82.99/2.99        & 82.31/3.15     \\
\bottomrule
\multicolumn{7}{r}{OA/TA = Accuracy to original label / Accuracy to target label.}\\ 
\end{tabular}}
\vspace{2ex}
\caption{Accuracy (Top1) [\%] of ResNet56 like CNNs after GenAttack adversarial attacks for CIFAR-10. BatchNorm~\cite{bn} layers are in $\mathtt{mini}$-$\mathtt{batch}$ mode.}
\label{tab:GenAttackResNet56_train}
\end{table}

    \begin{table*}[t]
\centering
\resizebox{0.9\textwidth}{!}{
\renewcommand{\arraystretch}{1.0}
\begin{tabular}{l|c|ccccc}
\toprule
\multirow{2}{*}{\textbf{GenAttack}}&\multirow{2}{*}{\textbf{$\epsilon$ }}             & $i=200$          & $i=400$         & $i=600$        & $i=800$       & $i=1000$     \\ 
                  &                                 & \textbf{OA/TA}   & \textbf{OA/TA}  & \textbf{OA/TA} & \textbf{OA/TA}& \textbf{OA/TA}\\ 
\midrule
\midrule
\textbf{ResNet50}\cite{He2015DeepRL}        &  8.0  & 21.29/12.80   & 11.64/34.46   &  6.87/51.94   &  4.67/64.08   &  3.06/72.82   \\         
(75.43 \%)                                          & 12.0  &  13.16/17.45  & 5.67/41.19    &  3.55/56.65   &  2.40/67.29   &  1.60/74.58      \\
\midrule                                            
\textbf{ResNet18}\cite{He2015DeepRL}        & 8.0   & 16.41/14.52   & 8.11/41.83    &  4.35/62.58   &  2.36/75.62   &  1.34/83.29 \\         
(69.00 \%)                                          & 12.0  & 10.24/22.44   & 5.13/50.74    &  2.70/68.85   &  1.58/80.21   &  1.04/86.62 \\
\midrule                                            
\textbf{ResNet18-Ch.Prune}\cite{AMC2019}  & 8.0   & 12.34/12.82   & 6.05/39.02    &  3.17/60.46   &  2.00/74.46   &  1.22/82.79 \\         
(67.62 \%)                                          & 12.0  & 7.33/20.25    & 3.29/49.44    &  1.84/68.97   &  1.08/80.11   &  0.88/86.80 \\
\midrule
\textbf{ResNet18-XNOR}\cite{rastegari2016}           & 8.0   & 13.06/0.64    & 12.86/0.72    & 12.64/0.84       & 12.68/0.86        &  12.68/0.94   \\         
(49.10 \%)                                          & 12.0  & 11.56/0.78         & 11.14/0.92         &  11.14/1.04        &  11.04/1.16        &  10.82/1.22  \\
\midrule                                       
\textbf{ResNet18-ABC(1$\times$1)}\cite{lin2017}&  8.0  & 17.59/1.48    & 17.67/1.62    & 17.37/1.76    &  17.23/1.88   &  16.89/1.98  \\         
(51.07 \%)                                          & 12.0  & 15.83/1.90    & 15.40/2.08    &  15.20/2.26   &  15.02/2.34   & 14.86/2.52  \\
\midrule
\textbf{ResNet18-ABC(3$\times$3)}\cite{lin2017}&  8.0  & 26.00/0.68    & 25.02/0.82    & 25.26/0.92    &  25.46/0.98   &  25.58/0.96  \\         
(59.83 \%)                                             & 12.0  & 22.50/0.74    & 22.04/0.94    &  22.36/1.02   &  21.75/1.08   & 21.90/1.14  \\
\bottomrule
\multicolumn{7}{r}{OA/TA = Accuracy to original label / Accuracy to target label.}\\ 
\end{tabular}}
\vspace{2ex}
\caption{Accuracy (Top1) [\%] of CNNs after GenAttack adversarial attacks for ImageNet. Pop size = 6.}
\label{tab:GenAttackImageNet_eval}
\end{table*}

    \begin{table}[ht]
\centering
\resizebox{0.7\textwidth}{!}{
\renewcommand{\arraystretch}{0.9}
\begin{tabular}{l|c|ccccc}
\toprule
\multirow{2}{*}{\textbf{GenAttack}}&\multirow{2}{*}{\textbf{$\epsilon$ }}             & $i=200$          & $i=400$         & $i=600$        & $i=800$       & $i=1000$     \\ 
                  &                                 & \textbf{OA/TA}   & \textbf{OA/TA}  & \textbf{OA/TA} & \textbf{OA/TA}& \textbf{OA/TA}\\ 
\midrule
\midrule
\textbf{ResNet50-Vanilla}\cite{He2015DeepRL}        &  8.0  & 32.37/1.94   & 26.02/5.01   &  23.36/9.17   &  20.47/12.82   &  18.59/15.93   \\         
(75.43 \%)                                          & 12.0  &  28.71/2.06  & 23.42/5.37    &  21.17/8.75   &  18.57/11.46   &  16.49/13.34      \\
\midrule                                            
\textbf{ResNet18-Vanilla}\cite{He2015DeepRL}        & 8.0  &  15.34/1.46  & 10.90/2.88    &  8.31/3.08   &  7.31/3.79   &  6.57/4.65      \\
(69.00 \%)                                          & 12.0  &  12.02/0.78  & 9.56/1.76    &  8.31/3.08   &  7.31/3.79   &  6.57/4.65      \\
\midrule                                            
\textbf{ResNet18-Ch.Prune}\cite{He2015DeepRL}  & 8.0   &  17.47/0.72  & 15.00/1.94    &  13.28/3.21   &  11.92/4.61   &  11.92/4.61      \\ 
(67.62 \%)                                          & 12.0 &  16.27/0.76  & 13.70/1.90    &  12.12/3.12   &  11.34/4.41   &  10.60/5.97      \\
\midrule
\textbf{ResNet18-XNOR}\cite{He2015DeepRL}           & 8.0   &  25.70/0.12  & 25.64/0.06    &  25.70/0.08   &  25.72/0.14   &  25.68/0.10      \\        
(49.10 \%)                                          & 12.0 &  24.36/0.02  & 23.80/0.08    &  24.02/0.04   &  24.34/0.06   &  24.12/0.06      \\
\midrule                                       
\textbf{ResNet18-ABC(1$\times$1)}\cite{He2015DeepRL}&  8.0 &  26.28/0.06  & 27.40/0.08    &  26.56/0.12   &  26.82/0.06   &  26.58/0.12      \\
(51.07 \%)                                          & 12.0  &  25.26/0.18  & 24.90/0.10    &  24.24/0.12   &  24.14/0.16   &  24.28/0.14      \\
\midrule
\textbf{ResNet18-ABC(3$\times$3)}\cite{He2015DeepRL}&  8.0  &  33.17/0.10  & 33.17/0.12    &  32.55/0.14   &  32.57/0.08   &  32.59/0.10      \\
(59.83 \%)                                             & 12.0  &  31.87/0.04  & 30.39/0.04    &  30.87/0.04   &  31.51/0.08   &  30.97/0.06      \\
\bottomrule
\end{tabular}}
\vspace{2ex}
\caption{Accuracy (Top1) [\%] of CNNs after GenAttack adversarial attacks for ImageNet. BatchNorm~\cite{bn} layers are in $\mathtt{mini}$-$\mathtt{batch}$ mode.}
\label{tab:GenAttackImageNet_train}
\end{table}

%  __   ___  ___  ___  __   ___       __   ___  __  
% |__) |__  |__  |__  |__) |__  |\ | /  ` |__  /__` 
% |  \ |___ |    |___ |  \ |___ | \| \__, |___ .__/
\clearpage
{\small
\bibliographystyle{ieee_fullname}
\bibliography{references}
}